\documentclass[11pt]{article}

\usepackage{ACL}
\usepackage{caption}
\usepackage{times}
\usepackage{latexsym}
\usepackage{amssymb}
\usepackage{bbm}
\usepackage{amsmath}
\usepackage{makecell}
\usepackage{booktabs}
\usepackage{multirow}
\usepackage[table,dvipsnames]{xcolor}
\usepackage[hypcap=false]{caption}
\usepackage{pifont}

\newcommand{\xmark}{\ding{55}}

\definecolor{lightblue}{HTML}{E6F3FF}
\definecolor{lightgreen}{HTML}{E6FFE6}
\definecolor{lightyellow}{HTML}{FFF9E6}

\usepackage{algorithm}
\usepackage{algorithmic}
\usepackage{amsmath,amssymb}
\DeclareMathOperator*{\argmax}{arg\,max}

\usepackage[T1]{fontenc}

\usepackage[utf8]{inputenc}

\usepackage{microtype}

\usepackage{inconsolata}

\usepackage{graphicx}

\title{MedPrune: Topology-Efficient Multimodal Multi-Agent Communication Evolution for Medical VQA Tasks}

\author{Jiuheng Wan$^{1}$, Runze Li$^{2}$, Chen Chen$^{3}$, Tingyuan Hu$^{4}$, Daiyang Yu$^{1}$,Yimin Jing$^{5}$, \\ \textbf{Taolin Zhang}$^{1}$\thanks{\ \ corresponding authors.}, \textbf{Richang Hong}$^{1}$\\
$^1$ School of Computer Science and Information Engineering, Hefei University of Technology \\
$^2$ Nanjing University
$^3$ Guangdong University of Finance and Economics\\
$^4$ East China Normal University
$^5$ Tianxi AI Technology Platform, Lenovo\\
 {\tt {2025110448@mail.hfut.edu.cn, tlzhang}@hfut.edu.cn} \\
}

\begin{document}
\maketitle
\begin{abstract}
While medical multimodal large language models (Med-MLLMs) advance medical visual question answering (VQA), existing clinical workflow–inspired multi-agent frameworks suffer from interaction patterns and excessive computational overhead caused by redundant communication topologies.
In this paper, we propose \texttt{MedPrune}, an efficient \textbf{MED}ical multimodal multi-agent collaboration framework that dynamically \textbf{PRUNE}s both nodes and edges from the communication topology to enhance reasoning ability and token efficiency.
Specifically, we first formulate the diagnostic process as a heterogeneous communication graph, where nodes represent specialist agents from various departments and edges capture intra- and inter-departmental interactions. Building on this graph, we introduce two sparsification mechanisms to enable adaptive collaborative evolution: (1) Heterogeneous Node Sparsification, which eliminates task-irrelevant specialist agents irrelevant to the current multimodal question via reinforcement learning–driven topological optimization, and (2) Heterogeneous Edge Sparsification, which selectively retains only the most diagnostically salient intra- and inter-departmental connections by jointly optimizing task performance and topological complexity.
Extensive medical VQA experiments under full-set and few-shot training settings prove \texttt{MedPrune} surpasses multi-agent baselines and boosts token efficiency with strong adversarial robustness.
\footnote{Source code and data are provided in the supplementary material and will be released upon acceptance.}
\end{abstract}

\section{Introduction}
The integration of MLLMs into medicine has advanced automated diagnostic support, with Med-MLLMs showing strong capabilities in multimodal tasks \cite{DBLP:conf/cvpr/ZouY25,DBLP:conf/cvpr/WangWLMW0025,wang2026navigating}.
However, as the focus shifts from controlled benchmarks to real-world clinical deployment, a fundamental limitation emerges from the inherently multi-specialty and collaborative nature of medicine \cite{DBLP:journals/npjdm/ChenYYLWLZGFCLFLL25,DBLP:conf/coling/FanWT0WWH25}.
A single LLM-based model cannot capture the deep expertise of fields, and its performance deteriorates on complex cases requiring cross-department reasoning \cite{DBLP:conf/acl/ZhuZZZWCL0025,lin2026march}.

\begin{figure*}
\centering
\includegraphics[width=15.5cm, height=8cm]{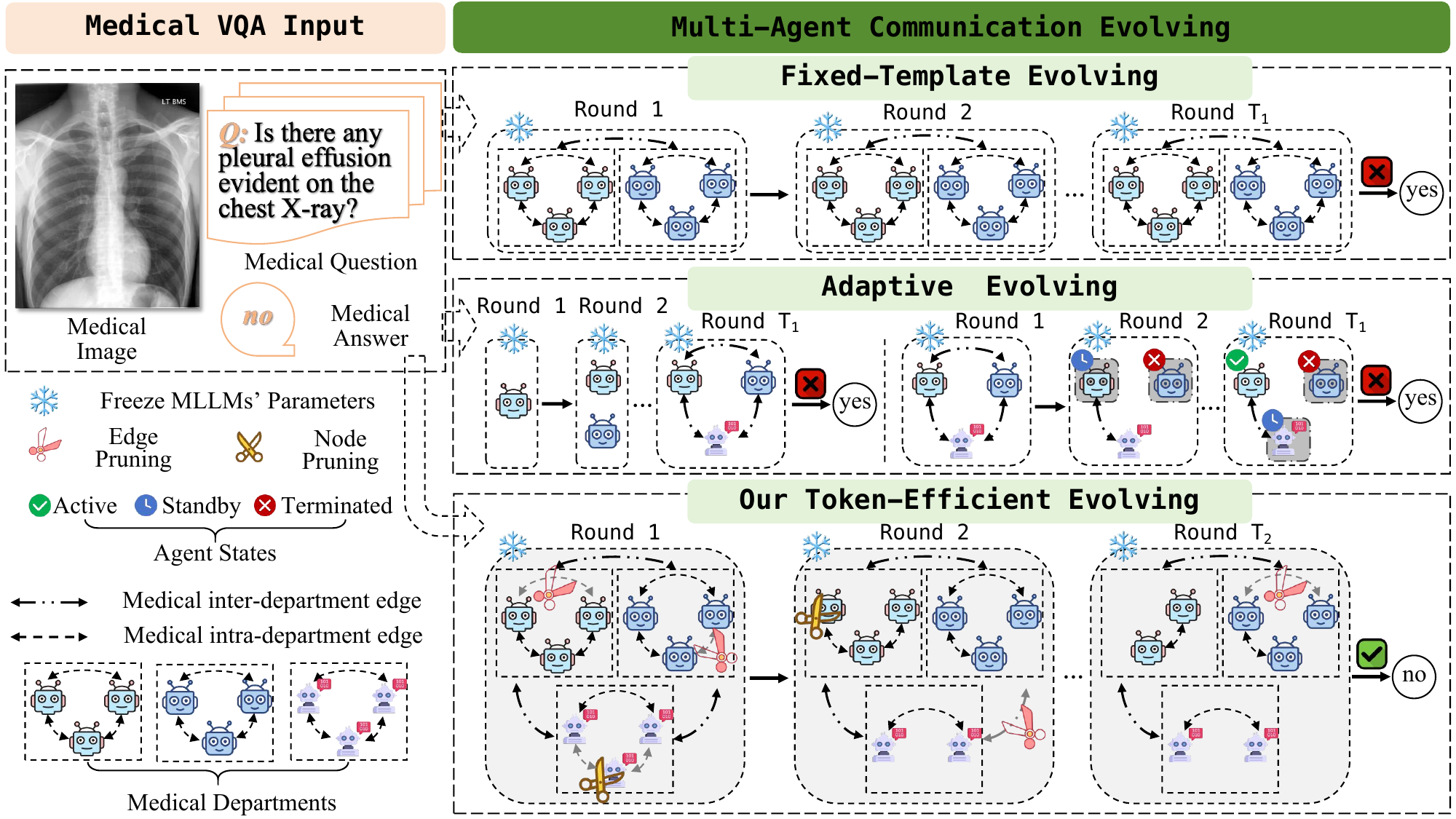}
\caption{Comparison of our approach with existing multi-agent methods. \texttt{Fixed-Template Evolving} relies on static and predefined communication templates, which leads to degraded diagnostic performance due to noisy redundant discussions. \texttt{Adaptive Evolving} can also yield a smaller communication subgraph in the final stage, yet its iterative topology refinement still incur substantial token overhead, failing to achieve true medical efficiency gains. In contrast, our \texttt{MedPrune} dynamically prunes irrelevant medical agents and redundant edges simultaneously, enabling token-efficient multi-department diagnostic reasoning (Rounds $T_2 \ll T_1$).}
\label{motivation_fig}
\vspace{-1.5em}
\end{figure*}

Previous works for medical VQA can be broadly categorized into three paradigms: (1) \texttt{{Training-free}}: single-agent prompting methods such as specialized chain-of-thought (CoT) \cite{DBLP:conf/nips/Wei0SBIXCLZ22} and RAG \cite{DBLP:conf/emnlp/XiaZLZLLZY24,DBLP:conf/mm/WangAHLA25,DBLP:journals/corr/abs-2603-03292}, aim to improve reasoning without model parameters updates.
While these methods avoid costly fine-tuning and can incorporate external knowledge, they remain susceptible to medical hallucinations when the domain knowledge is insufficient \cite{DBLP:conf/acl/ZhuZZZWCL0025,DBLP:journals/corr/abs-2503-05777}.
(2) \texttt{{Training-intensive}}: these methods are including supervised fine-tuning and reinforcement learning techniques, directly optimize model parameters on medical data ~\cite{DBLP:journals/air/ElmekkiIASSZZBKXPMOSM25,aydın2026acelor}. While they achieve stronger modality alignment and improved factual grounding within a unified representation space, they require substantial volumes of high-quality annotated medical data and expensive computational resource ~\cite{DBLP:conf/acl/DongKYLCJ25,jiang2026}.
(3) \texttt{{Multi-agent Collaboration}}: these works have recently emerged as a flexible paradigm that circumvents the need for model fine-tuning \cite{DBLP:conf/cvpr/ChenZY0BW25,DBLP:journals/corr/abs-2603-01607}.
By orchestrating multiple MLLMs-based agents to mimic clinical multi-departmental collaboration, they can gather different expert opinions to reach a consensus diagnosis ~\cite{DBLP:conf/acl/ZhuZZZWCL0025}.
Although this approach improves diagnostic coverage and reduces dependency on any single model, existing implementations often rely on static, predefined interaction workflows ~\cite{DBLP:journals/corr/abs-2506-00555,DBLP:journals/corr/abs-2503-18968}.
Such rigidity fails to capture the dynamic, case-adaptive nature of real-world clinical consultations, while token consumption escalates rapidly as more agents and discussion rounds are involved \cite{DBLP:journals/corr/abs-2602-00454,DBLP:conf/acl/YoonMLKQ26}.
Concurrently, several adaptive topology optimization methods have emerged to mitigate such overhead by either pruning or generating redundant agents or communication edges based on states via learnable graph structures \cite{DBLP:journals/corr/abs-2511-19969, DBLP:conf/aaai/LiLWZP26,zhang-etal-2026-taming}, yet their effectiveness in medical multimodal reasoning remains unexplored; see Fig. \ref{motivation_fig}.

In this paper, we introduce a \textbf{MED}ical multimodal multi-agent \textbf{PRUNING} framework, named \texttt{MedPrune}, that formalizes medical diagnosis as a dynamic evolutionary graph learning over heterogeneous multi-departmental collaboration.
Our main contributions are summarized below:
\begin{itemize}
    \item \textbf{Heterogeneous Node Sparsification:} 
    Existing multi-agent medical systems typically communicate with all departments agents, thereby compelling models to allocate computational resources to clinically irrelevant specialties ~\cite{DBLP:journals/corr/abs-2506-00555,DBLP:journals/corr/abs-2603-01607}.
    To address this, we propose a heterogeneous node sparsification that selectively prunes diagnostically irrelevant specialist nodes by learning task-adaptive inter-departmental communication graph.
    Specifically, we parameterize the heterogeneous graph with learnable soft adjacency matrices and optimize them via reinforcement learning using task performance as the reward signal.
    Then, pruning nodes with minimal weighted degree along with their incident edges to dynamically assemble a compact team of relevant specialists and eliminate token overhead from the initial round.

    \item \textbf{Heterogeneous Edge Sparsification:}
    Even after pruning irrelevant agents, redundant communication edges may persist, wasting tokens and propagating hallucinations across departments ~\cite{DBLP:conf/iclr/ZhangYLYWWCY025,DBLP:journals/corr/abs-2511-19969}.
    Hence, after re-initializing edge weights on the node-sparsified subgraph, we jointly optimize them under a composite objective that maximizes task performance while penalizing topological complexity. This process prunes low-utility edges and retains only the most salient intra- and inter-departmental connections, ensuring that each remaining edge contributes high diagnostic utility.
    Two-stage sparsification fundamentally addresses efficiency bottlenecks in multi-agent medical reasoning, achieving high diagnostic performance with drastically reduced token costs.
\end{itemize}

We evaluate \texttt{MedPrune} on three authoritative medical VQA benchmarks: MIMIC-CXR \cite{DBLP:journals/corr/abs-1901-07042}, IU-Xray \cite{DBLP:journals/jamia/Demner-FushmanK16} and OmniMedVQA \cite{DBLP:conf/cvpr/HuLLSHQL24} under full-set and few-shot training settings.
Our method consistently outperforms strong baselines across training-free, training-intensive, and medical multi-agent systems. Under both training settings, our approach achieves an average performance improvement of \textbf{3.65\%} and a \textbf{37.86\%} boost in token efficiency.
In addition, our method only reduces the average performance by \textbf{3.1\%} under the adversarial attack settings, which greatly proves the robustness of our system.

\section{Related Work}
\subsection{Medical Visual Question Answering}
Existing approaches can be primarily divided into two paradigms: (1) \textbf{Training‑free Methods.} They aim to elicit better reasoning from frozen Med-MLLMs backbones. CoT \cite{DBLP:conf/nips/Wei0SBIXCLZ22} decomposes complex clinical questions into intermediate reasoning steps.
Multimodal RAG \cite{DBLP:conf/emnlp/XiaZLZLLZY24,DBLP:conf/iclr/0005ZLWSWZ0Y25,DBLP:conf/mm/WangAHLA25,DBLP:journals/corr/abs-2603-03292} further supplies external knowledge such as medical literature or electronic health records to ground the model’s response. 
(2) \textbf{Training‑intensive Methods.} They focus on updating model parameters on domain‑specific corpora. SFT \cite{he2024pefomed,DBLP:conf/isbi/HeLLZ25,DBLP:journals/cmig/GongYHJ25} aligns Med‑MLLMs with medical image–text pairs, enhancing modality fusion and factual grounding. Reinforcement learning works \cite{DBLP:journals/corr/abs-2504-01886,DBLP:journals/corr/abs-2601-17151} further steers generation toward clinician‑preferred outputs by optimizing for human‑defined reward signals.
While these techniques yield stronger in‑distribution results, they demand large‑scale and high‑quality annotated medical data \cite{DBLP:conf/cvpr/MotaVABSB22,DBLP:conf/naacl/MatosCPLPITRNPSACWBG25}.

\subsection{Multimodal Multi-Agent Collaboration}
Research in this area falls into two categories:
(1) \textbf{Static Collaboration.}
AdaCoMed~\cite{DBLP:conf/cvpr/ChenZY0BW25} uses a mixture-of-modality-experts to fuse uni-modal representations and introduces adaptive co-learning coupling experts with a compact model.
MedAgent-Pro~\cite{DBLP:journals/corr/abs-2503-18968} operationalizes diagnostic protocols by decomposing clinical reasoning into evidence-grounded stages.
(2) \textbf{Adaptive Evolving Collaboration.} 
Rising VQA complexity exposes limitations of static multi-agent systems, driving dynamic multi-modal collaboration ~\cite{DBLP:conf/acl/RanaldiRP25,skillgraph}.
MMedAgent-RL ~\cite{DBLP:journals/corr/abs-2506-00555} uses RL to learn optimal policies among medical agents.
M$^3$Prune ~\cite{DBLP:journals/corr/abs-2511-19969} is a hierarchical communication graph pruning framework that eliminates redundant edges.
CARE \cite{DBLP:journals/corr/abs-2603-01607} decouples grounding and reasoning via coordinator-guided framework with explicit segmentation evidence.
ARG-Designer \cite{DBLP:conf/aaai/LiLWZP26} reframes topology design as autoregressive graph generation.
However, these methods primarily focus on generic collaboration, lacking specialized designs for heterogeneous medical knowledge.
In this paper, we jointly sparsify nodes and edges over the heterogeneous graph, achieving more efficient evolution rather than surface-level adaptations.

\begin{figure*}
\centering
\includegraphics[width=16cm, height=9cm]{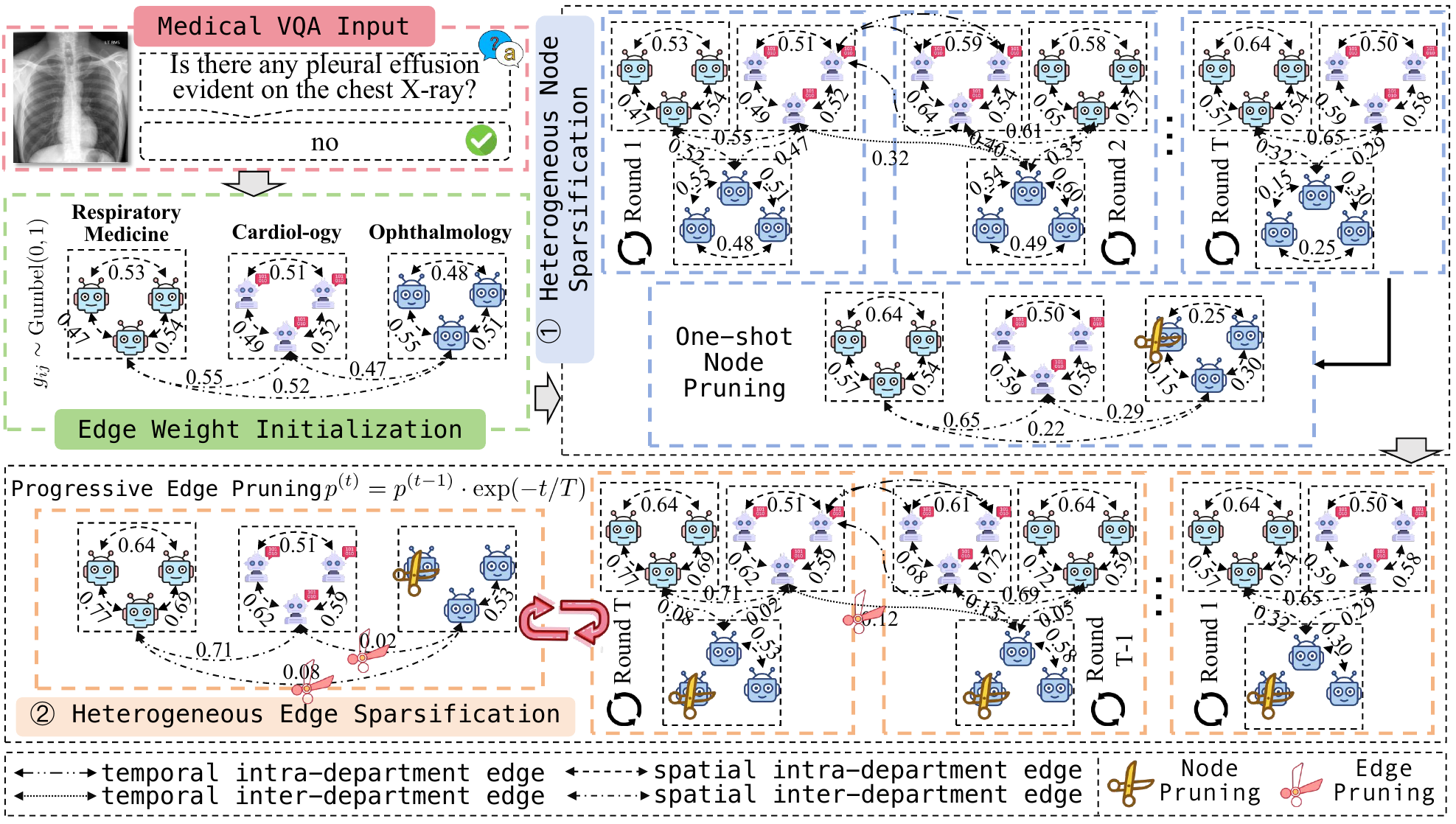}
\vspace{-1.8em}
\caption{Overview of \texttt{MedPrune}. (1) \texttt{Heterogeneous Node Sparsification} removes specialist agents irrelevant to the medical question. (2) \texttt{Heterogeneous Edge Sparsification} further prunes redundant intra- and inter-departmental connections via jointly maximizing task performance and penalizes topological complexity. Due to the space limitation, we only draw four spatial and temporal edge connections for one medical department.}
\label{model_fig}
\vspace{-1.5em}
\end{figure*}

\section{Model}

\subsection{Task Definition}
The main architecture is shown in Fig. \ref{model_fig}.
All the notations are reorganized in Appendix \ref{all_notations}.
We represent the multi-round diagnostic dialogue as a heterogeneous graph 
$\mathcal{G}=(\mathcal{V},\mathcal{E}^\mathcal{T},\mathcal{E}^\mathcal{S},\mathcal{M})$.
The nodes $\mathcal{V}=\{v_1, v_2, \dots, v_N\}$ are specialist agents, each initialized with a specific clinical role \footnote{Our agent roles prompts are shown in Appendix \ref{agent_roles}.}.
$N$ is the total number of agent nodes.
Edges are classified as temporal ($\mathcal{E}^\mathcal{T}$) for cross-round connections and spatial ($\mathcal{E}^\mathcal{S}$) for same-round connections.
The memory of agent $i$ at round $t$, denoted as $m_i^{(t)}$ and collectively stored in $\mathcal{M}^{(t)}$:
\begin{equation}
\label{eq_1}
m_i^{(t+1)}=f_{m}\left(m_i^{(t)},\mathbf{q},I_\mathcal{T}^{(t+1)},I_\mathcal{S}^{(t+1)}\right)
\end{equation}
where $f_{m}$ is an MLLM-based function that aggregates information from the agent's current memory, the question $\mathbf{q}$, and the messages $I_{\mathcal{T}}^{(t+1)}$ and $I_\mathcal{S}^{(t+1)}$ received from its temporal and spatial neighbors.

During our heterogeneous departmental evolution, the intra-department graph $\mathcal{G}_{\mathrm{intra}}^{(t)}$ is constructed as the union of graphs $\mathcal{G}_{\mathrm{intra}}^{c,(t)}$ for each clinical department $c$ at round $t$.
$\mathcal{G}_{\mathrm{intra}}^{c,(t)} = (\mathcal{V}^{c}, \mathcal{E}_{\mathrm{intra}}^{c,\mathcal{T},(t)}, \mathcal{E}_{\mathrm{intra}}^{c,\mathcal{S},(t)}, \mathcal{M}^{c,(t)})$ is defined within medical department $c$, where $\mathcal{V}^{c}$ denotes the set of agents belonging to that department.
Similarly, cross-department interactions are formalized through the inter-department graph $\mathcal{G}_{\mathrm{inter}}^{(t)} = \left( \mathcal{V}, \mathcal{E}_{\mathrm{inter}}^{\mathcal{T},(t)}, \mathcal{E}_{\mathrm{inter}}^{\mathcal{S},(t)}, \mathcal{M}^{(t)} \right)$, where $\mathcal{V} = \bigcup_c \mathcal{V}^c$.
In this graph, temporal edges $\mathcal{E}_{\mathrm{inter}}^{\mathcal{T},(t)}$ and spatial edges $\mathcal{E}_{\mathrm{inter}}^{\mathcal{S},(t)}$ exclusively connect agents originating from distinct departments.
At each round $t$, agent $v_i$ generates its opinion $\mathcal{O}_i^{(t)}$ via memory aggregation $f_m$ over the current intra- and inter-departmental graph. After $T$ rounds, a summarizing agent $f_s$ produces the final answer: $\mathcal{O}_s^{(T)} = f_s(\mathbf{q}, I_\mathcal{T}^{(T)}, I_\mathcal{S}^{(T)})$.

\subsection{Heterogeneous Node Sparsification}
\label{sec_heterogeneous_node}
Within each medical department, some specialist agents contribute little to collaborative reasoning due to varying relevance to the diagnostic question \cite{DBLP:conf/nips/0002PJCXMLGB024,DBLP:journals/ijmi/MiaoWLL26}.
Hence, our method performs node-level pruning on the communication graph at each round, thereby removing medical agents that exhibit minimal contribution.

\paragraph{\textbf{Edge Weight Initialization.}} We begin with binary adjacency matrices $\mathbf{A}^\mathcal{S}, \mathbf{A}^\mathcal{T} \in \{0, 1\}^{N \times N}$ derived from the predefined edges $\mathcal{E}^\mathcal{S}$ and $\mathcal{E}^\mathcal{T}$.
To enable differentiable graph learning, we parameterize soft adjacency matrices $\tilde{\mathbf{A}}^\mathcal{S}, \tilde{\mathbf{A}}^\mathcal{T} \in [0, 1]^{N \times N}$ using the Gumbel-Softmax \cite{DBLP:conf/iclr/JangGP17}.
For example, each entry of $\tilde{\mathbf{A}}^\mathcal{S}$ is computed as:
\begin{equation}
\tilde{\mathbf{A}}^\mathcal{S}[i, j] = \frac{\exp\left((\log(\mathbf{A}^\mathcal{S}[i, j] + \epsilon) + g_{ij}) / \tau\right)}{\sum_{k=1}^N \exp\left((\log(\mathbf{A}^\mathcal{S}[i, k] + \epsilon) + g_{ik}) / \tau\right)}
\end{equation}
where $g_{ij} \sim \text{Gumbel}(0,1)$.
$\tau$ is the temperature coefficient and $\epsilon$ is a small constant.
$\tilde{\mathbf{A}}^{\mathcal{T}}$ is constructed analogously.
To show the departmental spatial and temporal evolution across rounds $t$, we parameterize round-specific soft adjacency matrices $\tilde{\mathbf{A}}^{\mathcal{S},(t)}$ and $\tilde{\mathbf{A}}^{\mathcal{T},(t)}$, respectively.

\paragraph{\textbf{Intra-department Message Passing.}} In the intra-department graph $\mathcal{G}_{\text{intra}}^{(t)}$, each agent $v_{c,i}^{(t)}$ in the medical department $c$ aggregates spatial messages from its neighbors at round $t$ as:
\begin{equation}
\mathcal{M}_{i,\text{intra}}^{c,\mathcal{S},(t)} = \sum_{v_{c,j}^{(t)} \in \mathcal{N}_{\text{intra}}^{\mathcal{S}}(v_{c,i}^{(t)})} \mathcal{W}_c^{\mathcal{S},(t)}[i,j] \cdot \mathcal{O}(v_{c,j}^{(t)})
\end{equation}
where $\mathcal{O}(v_{c,j}^{(t)})$ is the output of neighboring agent $v_{c,j}^{(t)}$ at round $t$.
The spatial attention weight $\mathcal{W}_c^{\mathcal{S},(t)}$ is obtained by softmax normalizing the soft adjacency matrix entries $\tilde{\mathbf{A}}_{\text{intra}}^{c,\mathcal{S},(t)}[i,j]$ over the spatial neighborhood $\mathcal{N}_{\text{intra}}^{\mathcal{S}}(v_{c,i}^{(t)})$ within the same department $c$.
Temporal message passing follows analogously with the temporal adjacency matrix. Each agent then updates its memory $m_{i}^{(t)}$ via Eq. \ref{eq_1} after aggregating spatial and temporal information.

\textbf{Inter-department Message Passing.} 
At round $t$, each agent $v_i^{(t)}$ aggregates cross-department spatial messages from its neighbors in $\mathcal{G}_{\text{inter}}^{(t)}$ as:
\begin{equation}
\mathcal{M}_{i,\text{inter}}^{\mathcal{S},(t)} = \sum_{v_j^{(t)} \in \mathcal{N}_{\text{inter}}^\mathcal{S}(v_i^{(t)})} \mathcal{W}^{\mathcal{S},(t)}[i,j] \cdot \mathcal{O}(v_j^{(t)})
\end{equation}
where the attention weight $\mathcal{W}^{\mathcal{S},(t)}$ is obtained by softmax normalizing $\tilde{\mathbf{A}}_{\text{inter}}^{\mathcal{S},(t)}[i,j]$ over the spatial neighborhood $\mathcal{N}_{\text{inter}}^\mathcal{S}(v_i^{(t)})$.
Analogous temporal message passing over the inter-department graph then enables each agent to integrate multi-department spatial and temporal information, subsequently generating its current-round response.

\textbf{One-shot Node Pruning.} 
For heterogeneous node sparsification, we optimize the learnable soft adjacency matrix $\tilde{\mathbf{A}}$ to maximize the expected task performance over the graph distribution:
\begin{equation}
\label{eq_5}
\argmax_{\tilde{\mathbf{A}}^\mathcal{S}, \tilde{\mathbf{A}}^\mathcal{T}} 
\mathbb{E}_{\mathcal{G}\sim\mathbb{G}}\left[\phi(\mathcal{G})\right]
\end{equation}
where $\mathbb{G}$ denotes the feasible domain of graph sampling, and $\phi(\cdot)$ is a task-specific performance metric. As the underlying MLLM may be accessed via non-differentiable APIs, we approximate the gradient using a policy gradient \cite{DBLP:journals/ml/Williams92}. Hence, we sample $K$ graph instances $\{\mathcal{G}_k\}_{k=1}^K$ from $\mathbb{G}$ and estimate the gradient as:
\begin{multline}
\label{eq_6}
\nabla_{\tilde{\mathbf{A}}^\mathcal{S}, \tilde{\mathbf{A}}^\mathcal{T}} \mathbb{E}_{\mathcal{G}\sim\mathbb{G}}\left[\phi(\mathcal{G})\right] \\ \approx \frac{1}{K} \sum_{k=1}^K \phi(\mathcal{G}_k) \nabla_{\tilde{\mathbf{A}}^\mathcal{S}, \tilde{\mathbf{A}}^\mathcal{T}} \log \mathcal{P}_{\tilde{\mathbf{A}}^\mathcal{S}, \tilde{\mathbf{A}}^\mathcal{T}}(\mathcal{G}_k)
\end{multline}
where the sampling probability of $\mathcal{G}_k$ is given by $\mathcal{P}_{\tilde{\mathbf{A}}^{\mathcal{S}},\tilde{\mathbf{A}}^{\mathcal{T}}}(\mathcal{G}_k)=\left(\prod_{t=1}^T\prod_{e_{ij}\in\mathcal{E}_k^{\mathcal{S},(t)}}\tilde{\mathbf{A}}^{\mathcal{S},(t)}[i,j]\right)\cdot\left(\prod_{t=2}^T\prod_{e_{ij}\in\mathcal{E}_k^{\mathcal{T},(t)}}\tilde{\mathbf{A}}^{\mathcal{T},(t)}[i,j]\right)$, where $\mathcal{E}_k^{\mathcal{S},(t)}$ and $\mathcal{E}_k^{\mathcal{T},(t)}$ denote the sets of spatial and temporal edges retained in the $k$-th sampled graph at round $t$.

After the graph optimization, we select the nodes with the smallest weighted in- and out-degree in each intra-round graph and remove these nodes with all the edges connected to them:
\begin{equation}
\label{eq_7}
\mathcal{V}_{\Delta} = \bigcup_t \left(\mathcal{V}^{(t)} \setminus \texttt{TopkN}\left((\tilde{\mathbf{A}}^{{\mathcal{S}},{(t)}},\tilde{\mathbf{A}}^{{\mathcal{T}},{(t)}}), 1 - \alpha\right)\right)
\end{equation}
where $\texttt{TopkN}(\cdot)$ returns $k$ nodes with the largest sum of in-degree and out-degree and $\alpha$ is the node pruning rate.
Here, $\tilde{\mathbf{A}}^{\mathcal{S},(t)}$ comprises the intra- and inter-department spatial adjacency matrices $\mathbf{A}_{\text {intra }}^{\mathcal{S},(t)}$ and $\mathbf{A}_{\text {inter }}^{\mathcal{S},(t)}$, and analogously for $\tilde{\mathbf{A}}^{\mathcal{T},(t)}$.
\subsection{Heterogeneous Edge Sparsification}
\label{sec_heterogeneous_edge}
Even after pruning agents with minimal contribution, the remaining specialist agents within each medical department may still engage in diagnostically noisy consultations, which can waste token budgets and propagate misinformation across the collaborative reasoning process \cite{DBLP:journals/corr/abs-2503-18968,DBLP:journals/npjdm/ChenYYLWLZGFCLFLL25}. Hence, we next prune less critical spatial and temporal edges from the communication graph.

\textbf{Intra-department Edge Optimization.}  
After node pruning, we optimize $\tilde{\mathbf{A}}^\mathcal{S}$ and $\tilde{\mathbf{A}}^\mathcal{T}$ with a composite objective that maximizes task performance while penalizing the nuclear norm to encourage sparsity and token efficiency:
\begin{equation}
\label{eq_8}
\argmax_{\tilde{\mathbf{A}}_{\mathrm{intra}}^{\mathcal{S}},\tilde{\mathbf{A}}_{\mathrm{intra}}^{\mathcal{T}}} \mathbb{E}_{\mathcal{G}\sim\mathbb{G}_{\mathrm{Node}}}\left[\phi(\mathcal{G})\right]-\|\tilde{\mathbf{A}}_{\mathrm{intra}}^{\mathcal{S}}\|_{*} - \|\tilde{\mathbf{A}}_{\mathrm{intra}}^{\mathcal{T}}\|_{*}
\end{equation}
subject to $\|\mathbf{A}_{\mathrm{intra}}^{\mathcal{S}}-\tilde{\mathbf{A}}_{\mathrm{intra}}^{\mathcal{S}}\|_{F} + \|\mathbf{A}_{\mathrm{intra}}^{\mathcal{T}}-\tilde{\mathbf{A}}_{\mathrm{intra}}^{\mathcal{T}}\|_{F} \leq \delta$, where $\delta$ controls the allowable deviation from the original binary adjacency matrices and $\|\cdot\|_F$ the Frobenius norm.
$\mathbb{G}_\mathrm{Node}$ is the feasible graph sampling domain after node pruning.
$\|\cdot\|_*$ denotes the nuclear norm, encouraging low-rank structures to induce sparsity in the communication graph topology \cite{DBLP:conf/iclr/ZhangYLYWWCY025}.
Following Eq. (\ref{eq_5}), we approximate the expected performance as a weighted average over $K$ sampled graphs, with the sampling probability jointly dependent on the spatial and temporal soft adjacency matrices:
\begin{multline}
\nabla_{\tilde{\mathbf{A}}^{\mathcal{S}},\tilde{\mathbf{A}}^{\mathcal{T}}} \mathbb{E}_{\mathcal{G}\sim\mathbb{G}_{\mathrm{Node}}}\left[\phi(\mathcal{G})\right] \\ \approx \frac{1}{K} \sum_{k=1}^K \phi(\mathcal{G}_k) \nabla_{\tilde{\mathbf{A}}^{\mathcal{S}},\tilde{\mathbf{A}}^{\mathcal{T}}} \log \mathcal{P}_{\tilde{\mathbf{A}}^{\mathcal{S}},\tilde{\mathbf{A}}^{\mathcal{T}}}(\mathcal{G}_k)
\end{multline}
where $\mathcal{P}_{\tilde{\mathbf{A}}^{\mathcal{S}},\tilde{\mathbf{A}}^{\mathcal{T}}}(\mathcal{G}_k)$ is calculated analogously to Eq. (\ref{eq_6}).

\textbf{Inter-department Edge Optimization.}
We optimize the inter-department graph with a similar intra-department objective that jointly maximizes task performance while encouraging sparsity:
\begin{equation}
\label{eq_10}
\argmax_{\tilde{\mathbf{A}}_{\mathrm{inter}}^{\mathcal{S}},\tilde{\mathbf{A}}_{\mathrm{inter}}^{\mathcal{T}}}\mathbb{E}_{\mathcal{G}\sim\mathbb{G}_{\mathrm{Node}}}\left[\phi(\mathcal{G})\right]-\|\tilde{\mathbf{A}}_{\mathrm{inter}}^{\mathcal{S}}\|_{*}-\|\tilde{\mathbf{A}}_{\mathrm{inter}}^{\mathcal{T}}\|_{*}\\
\end{equation}
where \(\tilde{\mathbf{A}}_{\mathrm{inter}}^{\mathcal{S}}\) and \(\tilde{\mathbf{A}}_{\mathrm{inter}}^{\mathcal{T}}\) denote the learnable soft adjacency matrices for inter-department spatial and temporal connections, respectively.

\textbf{Progressive Edge Pruning.} Our heterogeneous edge pruning strategy first sparsifies intra-department edges, then prunes inter-department connections, using a progressively decaying pruning ratio over rounds to achieve controlled sparsity.

For each intra- and inter-department adjacency matrix, we retain the top $(1 - p^{(t)})$ fraction of edges by weight and discard the rest.
$\mathcal{E}_\Delta = \bigcup_{t} \left( \mathcal{E}^{\mathcal{S},(t)} \setminus \texttt{TopkE}\left( \tilde{\mathbf{A}}^{{\mathcal{S}},{(t)}}, 1 - p^{(t)} \right) \right) \cup \bigcup_{t} \left( \mathcal{E}^{\mathcal{T},(t)} \setminus \texttt{TopkE}\left( \tilde{\mathbf{A}}^{{\mathcal{T}},{(t)}}, 1 - p^{(t)} \right) \right)$, 
where $\texttt{TopkE}(\cdot)$ selects the edges corresponding to the top-$k$ largest entries in the given adjacency matrix. The pruning rate $p^{(t)}$ follows an exponential decay: $p^{(t)} = p^{(t-1)} \cdot \exp(-t/T)$.
After $T$ rounds of progressive pruning, the communication graph $\hat{\mathcal{G}}$ achieves a favorable trade-off between inference token efficiency and diagnostic effectiveness.

\subsection{Iteratively Training}
We adopt a two-stage training paradigm to iteratively optimize the communication graph: node sparsification $\mathcal{L}_n(\tilde{\mathbf{A}}^\mathcal{S}, \tilde{\mathbf{A}}^\mathcal{T})$ followed by edge sparsification $\mathcal{L}_e(\tilde{\mathbf{A}}^\mathcal{S}, \tilde{\mathbf{A}}^\mathcal{T})$.

Specifically, for heterogeneous node sparsification, we optimize the learnable spatial adjacency matrix by maximizing the expected task performance: $\mathcal{L}_n(\tilde{\mathbf{A}}^\mathcal{S}, \tilde{\mathbf{A}}^\mathcal{T}) = -\mathbb{E}_{\mathcal{G} \sim \mathbb{G}}\left[\phi(\mathcal{G})\right]$.
For heterogeneous edge sparsification, we add nuclear norm regularization to encourage sparsity: $ \mathcal{L}_e(\tilde{\mathbf{A}}^\mathcal{S}, \tilde{\mathbf{A}}^\mathcal{T}) = -\mathbb{E}_{\mathcal{G} \sim \mathbb{G}_{\text{Node}}}\left[\phi(\mathcal{G})\right] + \|\tilde{\mathbf{A}}^\mathcal{S}\|_* + \|\tilde{\mathbf{A}}^\mathcal{T}\|_* $.
We refer readers to the detailed training algorithm description in Appendix \ref{algo_des}.

\section{Experiments}
Due to space limitations, we relegate the experimental settings to Appendix~\ref{exp_settings} and present additional detailed analyses in Appendix~\ref{other_exp}.

\begin{table*}[!t]
  \centering
  \footnotesize
  \renewcommand\cellalign{cc}
  \renewcommand{\arraystretch}{0.85}
  \setlength{\tabcolsep}{4pt}
  \begin{tabular}{cc|ccccc|c|c|c}
    \toprule
    \multirow{3}{*}{\makecell{Training\\Paradigm}} 
    & \multirow{3}{*}{Method} 
    & \multicolumn{5}{c|}{\raisebox{-0.2em}{MIMIC-CXR}} 
    & \multicolumn{1}{c|}{\raisebox{-0.2em}{IU-Xray}} 
    & \multicolumn{1}{c|}{\raisebox{-0.2em}{\makecell{Omni-\\MedVQA}}}  
    & \multirow{3}{*}{Average} \\
    & & \multicolumn{5}{c|}{\raisebox{0.3em}{\rule{185pt}{0.4pt}}}
    & \multicolumn{1}{c|}{\raisebox{0.3em}{\rule{32pt}{0.4pt}}} 
    & \multicolumn{1}{c|}{\raisebox{0.3em}{\rule{38pt}{0.4pt}}} \\
    
    & & BLEU & ROUGE-L & BERT-F1 & METEOR & Avg. & Acc 
    &  Acc & \\
    \midrule
    \rowcolor{gray!20} \multicolumn{10}{c}{\textbf{Full-set Training}} \\
    \midrule
    \multirow{2}{*}{Zero-shot} 
    & SP & 7.65 & 18.09 & 62.58 & 29.64 & 29.49 & 84.08 & 72.36 & $61.98_{\pm0.7}$ \\
    & CoT & 15.25 & 28.57 & 69.13 & 24.60 & 34.39 & 86.02 & 71.07 & $63.83_{\pm1.1}$ \\
    \cmidrule(lr){1-10}
    \multirow{3}{*}{\makecell{Multimodal\\RAG}} 
    & MMed-RAG & 15.45 & 26.01 & 69.36 & 33.94 & 36.19 & 86.21 & 82.31 & $68.24_{\pm0.5}$ \\
    & MIRA & 18.30 & 31.62 & 69.62 & 30.32 & 37.47 & 86.41 & 85.89 & $69.92_{\pm0.9}$ \\
    & RULE & 21.36 & 33.55 & 71.66 & 34.34 & 40.23 & 87.38 & 87.82 & $71.81_{\pm1.3}$ \\
    \cmidrule(lr){1-10}
    \multirow{3}{*}{\makecell{Post-\\training}} 
    & LoRA & 22.01 & 34.94 & 70.92 & 34.63 & 40.62 & 89.90 & 92.10 & $74.21_{\pm0.4}$ \\
    & QLoRA & 21.94 & 34.15 & 71.48 & 34.93 & 40.62 & 89.32 & 90.35 & $73.43_{\pm1.2}$ \\
    & GRPO & 23.00 & \underline{35.19} & 70.36 & 34.80 & 40.84 & 89.51 & 92.65 & $74.33_{\pm1.0}$ \\
    \cmidrule(lr){1-10}
    \rowcolor{white} \multicolumn{2}{c}{} & \multicolumn{8}{c}
    {\textbf{Fixed-Template Evolving}} \\
    \cmidrule(lr){3-10}
    \multirow{8}{*}[-2pt]{Multi-agent} 
    & CARE & 22.58 & 34.72 & 70.71 & 32.32 & 40.08 & 89.90 & \underline{92.80} & $74.26_{\pm1.3}$\\
    & AdaCoMed & 23.34 & 34.91 & 72.21 & 35.36 & 41.45 & \underline{90.10} & 92.45 & $\underline{74.67}_{\pm0.6}$ \\
    
    & MedAgent-Pro & 22.92 & 35.05 & \underline{72.71} & 34.98 & 41.41 & 88.35 & 91.01 & $73.59_{\pm0.3}$ \\
    & MMedAgent-RL & \underline{23.56} & 35.09 & 71.42 & \underline{35.87} & \underline{41.48} & 88.93 & 91.80 & $74.07_{\pm1.4}$ \\
    \cmidrule(lr){3-10}
    \rowcolor{white} \multicolumn{2}{c}{} & \multicolumn{8}{c}{\textbf{Adaptive Evolving}} \\
    \cmidrule(lr){3-10}
    & M$^3$Prune & 21.29 & 34.79 & 71.79 & 32.50 & 40.09 & 89.13 & 90.51 & $73.24_{\pm1.0}$ \\
    & ARG-Designer$^\dagger$ & 21.56 & 34.92 & 72.05 & 32.81 & 40.38 & 88.54 & 90.11 & $73.01_{\pm1.1}$ \\
    \cmidrule(lr){3-10}
    \rowcolor{Lavender!20}
    & \textbf{Ours} & \textbf{23.76} & \textbf{35.27} & \textbf{73.86} & \textbf{36.15} & \textbf{42.26} & \textbf{91.07} & \textbf{96.36} & $\textbf{76.56}_{\pm0.8}$ \\
    \midrule
    \rowcolor{gray!20} \multicolumn{10}{c}{\textbf{Few-shot Training (10 samples)}} \\
    \midrule
    \multirow{2}{*}{Zero-shot} 
    & SP & 7.65 & 17.95 & 62.62 & 29.56 & 29.45 & 84.28 & 70.06 & $61.26_{\pm1.1}$ \\
    & CoT & 12.82 & 30.90 & 67.99 & 22.29 & 33.50 & 86.34 & 70.85 & $63.56_{\pm0.5}$ \\
    \cmidrule(lr){1-10}
    \multirow{3}{*}{\makecell{Multimodal\\RAG}} 
    & MMed-RAG & 15.12 & 25.25 & 69.36 & 33.62 & 35.84 & 85.72 & 81.16 & $67.57_{\pm0.9}$ \\
    & MIRA & 8.44 & 18.28 & 64.84 & 32.23 & 30.95 & 84.86 & 87.02 & $67.61_{\pm1.3}$ \\
    & RULE & 13.79 & 24.50 & 68.56 & 33.38 & 35.06 & 84.84 & 85.46 & $68.45_{\pm0.4}$ \\
    \cmidrule(lr){1-10}
    \multirow{3}{*}{\makecell{Post-\\training}} 
    & LoRA & 6.90 & 17.17 & 61.91 & 28.65 & 28.66 & 86.15 & 84.11 & $66.30_{\pm1.2}$ \\
    & QLoRA & 6.79 & 17.12 & 61.74 & 28.40 & 28.51 & 86.11 & 72.33 & $62.32_{\pm0.6}$ \\
    & GRPO & 6.28 & 16.04 & 61.50 & 28.40 & 28.05 & 86.54 & 81.21 & $65.27_{\pm1.0}$ \\
    \cmidrule(lr){1-10}
    \rowcolor{white} \multicolumn{2}{c}{} & \multicolumn{8}{c}
    {\textbf{Fixed-Template Evolving}} \\
    \cmidrule(lr){3-10}
    \multirow{8}{*}[-2pt]{Multi-agent} 
    & CARE & 16.21 & 31.76 & 69.13 & 25.48 & 35.64 & \underline{87.59} & 86.48 & $69.91_{\pm1.3}$\\
    & AdaCoMed & 12.79 & 32.44 & 71.22 & 33.70 & 37.54 & 82.09 & 85.98 & $68.54_{\pm0.8}$ \\
    & MedAgent-Pro & \underline{22.49} & \underline{33.70} & \underline{72.49} & \underline{34.42} & \underline{40.78} & 85.84 & \underline{89.71} & $\underline{72.11}_{\pm1.4}$ \\
    & MMedAgent-RL & 19.12 & 31.97 & 70.05 & 34.23 & 38.84 & 87.12 & 82.09 & $69.35_{\pm0.5}$ \\
    \cmidrule(lr){3-10}
    \rowcolor{white} \multicolumn{2}{c}{} & \multicolumn{8}{c}{\textbf{Adaptive Evolving}} \\
    \cmidrule(lr){3-10}
    & M$^3$Prune & 21.76 & 33.40 & 71.62 & 31.30 & 39.52 & 86.54 & 89.04 & $71.70_{\pm0.3}$ \\
    & ARG-Designer$^\dagger$ & 21.98 & 33.67 & 71.91 & 31.65 & 39.86 & 86.23 & 88.82 & $71.64_{\pm0.6}$ \\
    \cmidrule(lr){3-10}
    \rowcolor{SkyBlue!20}
    & \textbf{Ours} & \textbf{22.67} & \textbf{34.34} & \textbf{73.24} & \textbf{35.26} & \textbf{41.38} & \textbf{90.66} & \textbf{95.79} & $\textbf{75.94}_{\pm0.8}$ \\
    \bottomrule
  \end{tabular}
  \caption{The performance of full-set and few-shot training with \textbf{Qwen3-VL (8B)}. Due to space limitation, the results across different backbones (\textbf{HuatuoGPT-Vision (7B)}, \textbf{Qwen3.6-Plus (397B)}) and few-shot settings with \textbf{20/40 training samples} are provided in Appendix \ref{general_per_glm}.
  The t-tests demonstrate the improvements are statistically significant with $p$ < 0.05 level. Best results are in bold and second-best results are underlined.}
  \label{full_exp_modified}
  \vspace{-2em}
\end{table*}

\begin{table}[!t]
\centering
\resizebox{\linewidth}{!}{
\begin{tabular}{lccc}
\toprule
\multicolumn{1}{l}{\multirow{1}{*}{\textbf{Dataset}$ \quad \rightarrow$}} & \multirow{2}{*}{\textbf{MIMIC-CXR}} & \multirow{2}{*}{\textbf{IU-Xray}} & \multirow{2}{*}{\textbf{Average}} \\
\multicolumn{1}{l}{\multirow{1}{*}{\textbf{Method}$  \quad \downarrow$}}  & & & \\

\midrule
\rowcolor{gray!20} \multicolumn{4}{c}{\textbf{Qwen3-VL (8B)}} \\
\midrule
\textbf{MedPrune} & \textbf{42.26} & \textbf{91.07} & \textbf{66.67} \\
\cmidrule{1-4}
\rowcolor{Aquamarine!20}w/o NS & 38.59${\textcolor{orange}{(\downarrow 3.67)}}$ & 86.41${\textcolor{orange}{(\downarrow 4.66)}}$ & 62.50${\textcolor{orange}{(\downarrow 4.17)}}$ \\
\rowcolor{Aquamarine!20}w/o ES & 38.55${\textcolor{orange}{(\downarrow 3.71)}}$ & 85.44${\textcolor{orange}{(\downarrow 5.63)}}$ & 62.00${\textcolor{orange}{(\downarrow 4.67)}}$ \\
w/o norm & 41.24${\textcolor{orange}{(\downarrow 1.02)}}$ & 86.99${\textcolor{orange}{(\downarrow 4.08)}}$ & 64.12${\textcolor{orange}{(\downarrow 2.55)}}$ \\
w/o PP & 41.18${\textcolor{orange}{(\downarrow 1.08)}}$ & 87.18${\textcolor{orange}{(\downarrow 3.89)}}$ & 64.18${\textcolor{orange}{(\downarrow 2.49)}}$ \\
w/o DS & 41.35${\textcolor{orange}{(\downarrow 0.91)}}$ & 89.24${\textcolor{orange}{(\downarrow 1.83)}}$ & 65.30${\textcolor{orange}{(\downarrow 1.37)}}$ \\
\midrule
\rowcolor{gray!20} \multicolumn{4}{c}{\textbf{HuatuoGPT-Vision (7B)}} \\
\midrule
\textbf{MedPrune} & \textbf{44.62} & \textbf{87.96} & \textbf{66.29} \\
\cmidrule{1-4}
\rowcolor{Aquamarine!20}w/o NS & 41.56${\textcolor{orange}{(\downarrow 3.06)}}$ & 84.47${\textcolor{orange}{(\downarrow 3.49)}}$ & 63.02${\textcolor{orange}{(\downarrow 3.27)}}$ \\
\rowcolor{Aquamarine!20}w/o ES & 41.71${\textcolor{orange}{(\downarrow 2.91)}}$ & 84.27${\textcolor{orange}{(\downarrow 3.69)}}$ & 62.99${\textcolor{orange}{(\downarrow 3.30)}}$ \\
w/o norm & 42.02${\textcolor{orange}{(\downarrow 2.60)}}$ & 85.05${\textcolor{orange}{(\downarrow 2.91)}}$ & 63.54${\textcolor{orange}{(\downarrow 2.75)}}$ \\
w/o PP & 42.28${\textcolor{orange}{(\downarrow 2.34)}}$ & 85.05${\textcolor{orange}{(\downarrow 2.91)}}$ & 63.67${\textcolor{orange}{(\downarrow 2.62)}}$ \\
w/o DS & 43.70${\textcolor{orange}{(\downarrow 0.92)}}$ & 86.21${\textcolor{orange}{(\downarrow 1.75)}}$ & 64.96${\textcolor{orange}{(\downarrow 1.33)}}$ \\
\bottomrule
\end{tabular}
}
\caption{Ablation study of \texttt{MedPrune}. "NS", "ES", "PP" and "DS" are abbreviations for node sparsification, edge sparsification, progressive edge pruning and department selection, respectively.}
\label{tab:ablation_medprune}
\vspace{-1em}
\end{table}

\subsection{Main Results}
We evaluate \texttt{MedPrune} on three medical multimodal QA benchmarks covering various body parts (i.e., Chest, Bone, Eyes, and Thyroid) under full-set and few-shot training settings, against four types of strong baselines with Qwen3-VL (8B).

As shown in Table \ref{full_exp_modified}, we can observe that: (1) Methodologically, the performance hierarchy reveals a clear progression in reasoning capability. Training‑free prompting (SP, CoT) and training‑intensive single‑agent methods (RAG, LoRA, GRPO) both suffer from limited cross‑disciplinary reasoning capacity, with the former being particularly vulnerable to hallucinations \cite{DBLP:journals/corr/abs-2503-05777}. 
Fixed‑template systems incur heavy token overhead from static interaction patterns, while adaptive methods achieve only moderate topology refinement and remain suboptimal for medical reasoning. Our MedPrune jointly prunes irrelevant specialist nodes and redundant edges, dynamically tailoring the graph per question, and consistently outperforms all baselines with superior accuracy and token efficiency.
(2) Regardless of backbone generality (general‑domain vs. domain‑adapted) or scale (7B to 397B), MedPrune consistently achieves superior performance. This robustness stems from its task‑driven graph optimization, which is independent of specialized MLLM knowledge, while larger backbones universally improve all methods with more pronounced gains for multi‑agent systems, owing to their enhanced multimodal alignment and instruction‑following capabilities \cite{DBLP:conf/acl/LuoZCLLWYLWZGSL25}.
(3) Regarding training strategy, few-shot training with only 10 samples enables \texttt{MedPrune} to achieve performance comparable to full-set training, particularly with domain-adapted backbones. This stability is attributed to our task-driven graph optimization, which involves no MLLMs' parameter updates and thus avoids the overfitting and representation disruption that plague training-intensive methods under extreme data scarcity \cite{DBLP:journals/corr/abs-2602-18888}.
Consequently, our model reduces data dependency, offering a robust diagnostic performance with high token economy solution for resource-constrained clinical environments.

\subsection{Ablation Study}
We ablate \texttt{MedPrune} by sequentially removing its core components and report averaged results on Qwen3-VL (8B) and HuatuoGPT-Vision (7B).
In Table~\ref{tab:ablation_medprune}, (1) removing NS or ES consistently causes the most substantial performance declines, confirming that pruning irrelevant agents and redundant communication edges is indispensable for focusing computation on high-utility collaboration. It also directly addresses the inefficiencies of static and adaptive multi-agent topologies. (2) Removing the nuclear norm (w/o norm) causes moderate drops, confirming that low‑rank priors aid compact topology.
Ablating progressive pruning (w/o PP) results in smaller declines, confirming that a constant ratio yields strong baselines, while progressive pruning further refines \texttt{MedPrune}'s performance edge.
Ablating department selection (w/o DS) causes only minor performance drop, indicating fixed departments suffice for promising results, while selection accelerates convergence via better initialization.

\begin{figure}[!t]
    \centering
    \includegraphics[width=\columnwidth]{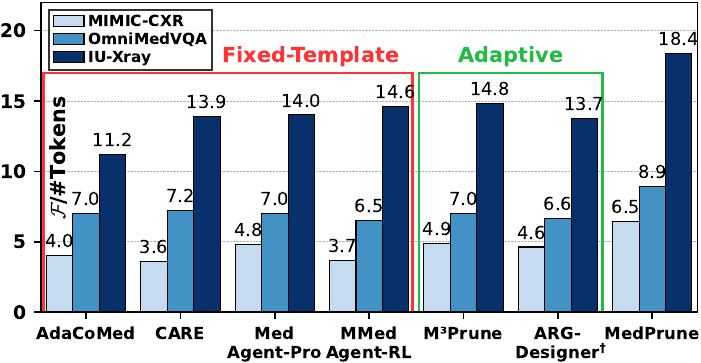}
    \vspace{-0.8cm}
    \caption{Token efficiency ($\mathcal{F}$/\#Tokens) comparison across various medical multi-agent frameworks.}
    \label{Token_Efficiency_fig}
        \vspace{-1.5em}
\end{figure}

\subsection{Detailed Analysis}
\begin{figure*}
    \centering
    \includegraphics[width=\linewidth]{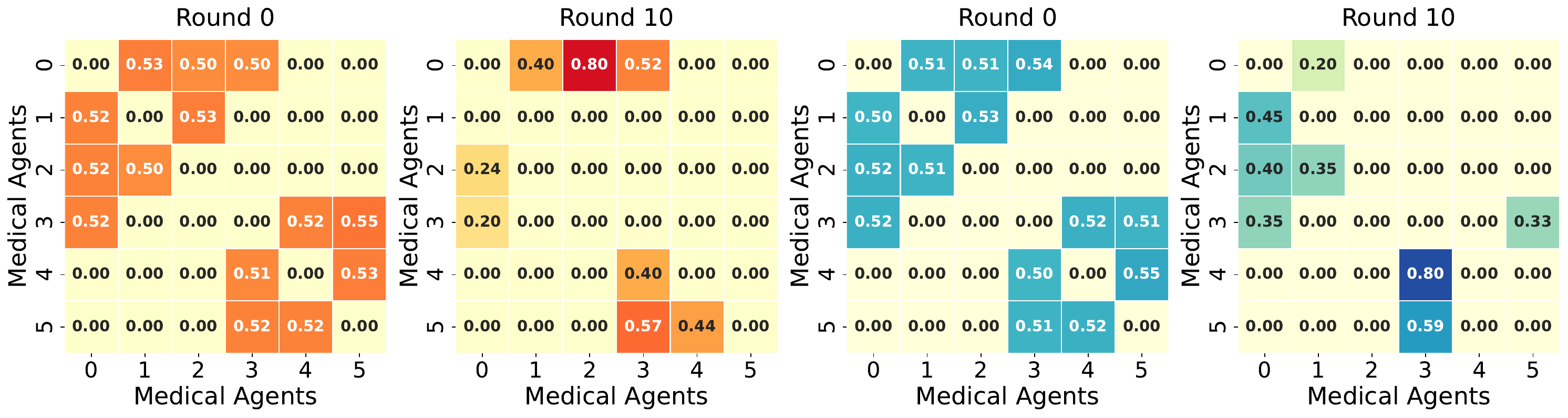}
        \vspace{-0.8cm}
    \caption{Visualization of intra‑ (Left) and inter‑ (Right) department communication edge weight evolution across multi‑agent interaction rounds (Best viewed in color).}
    \label{heatmaps_fig}
    \vspace{-1.5em}
\end{figure*}

\begin{figure}[!t]
    \centering
    \includegraphics[width=\columnwidth]{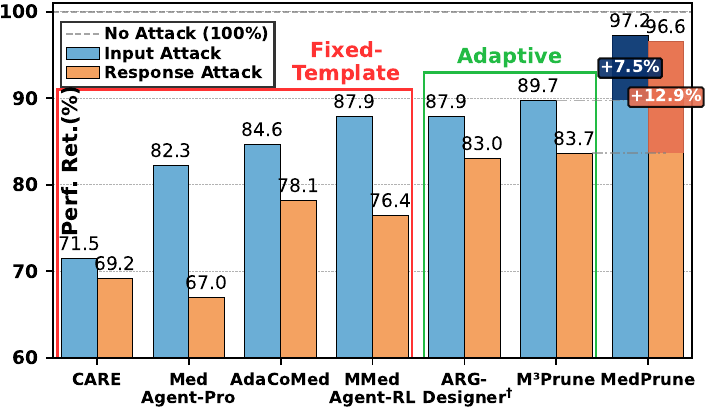}
    \vspace{-0.8cm}
    \caption{Model performance retention under adversarial attacks (prompt and response perturbations).}
    \label{attack_fig}
    \vspace{-1.5em}
\end{figure}
\textbf{(1) Communication Edge Evolution.} We examine intra- and inter-departmental soft edge adjacency matrix ($\tilde{\mathbf{A}}^\mathcal{S}, \tilde{\mathbf{A}}^\mathcal{T}$) across rounds to analyze topology evolution, and visualize connectivity patterns among medical agent roles using Qwen3-VL (8B) on MIMIC-CXR.

Fig. ~\ref{heatmaps_fig} reveals that: (1) as interaction rounds increase, communication concentrates on a few critical agents and departments, shown by darker, concentrated blocks; (2) the heterogeneous pruning strategy progressively sparsifies the topology by suppressing low‑relevance edges, reflected in fading redundant connections. These results confirm that \texttt{MedPrune} automatically identifies the most useful agents, adaptively reinforces critical intra‑ and inter‑departmental communications, and systematically eliminates redundant interactions.\\
\textbf{(2) Token Efficiency Analysis.} We evaluate cost‑effectiveness by comparing diagnostic performance against total token consumption (prompt + completion tokens) using Qwen3‑VL (8B), with token efficiency quantified as performance per token ($\mathcal{F}$/\#Tokens) ~\footnote{$\mathcal{F}$ denotes the dataset‑specific performance metric: average of BLEU, ROUGE‑L, BERT‑F1, and METEOR for MIMIC‑CXR; accuracy for IU‑Xray; and average accuracy over its three sub‑datasets for OmniMedVQA.}.

Fig. ~\ref{Token_Efficiency_fig} shows \texttt{MedPrune} achieves the best performance under comparable token budgets via adaptive heterogeneous pruning, which dynamically shapes communication topology with joint node-edge sparsification.
However, the fixed template methods without question-specific adaptation and adaptive baselines merely tuning edges instead of pruning irrelevant agents, both suffering high token overhead.
By concentrating computation on useful dialogues and removing redundancies, \texttt{MedPrune} delivers a better performance‑efficiency balance.\\
\textbf{(3) Robustness Verification.} We evaluate adversarial robustness on MIMIC-CXR with Qwen3-VL (8B) by comparing \texttt{MedPrune} to multi-agent baselines under two attacks: (1) \textit{Input Prompt Attack}: where a benign agent is replaced by a malicious one that ignores inter‑agent communication; (2) \textit{Response Attack}: where an agent produces deliberately wrong answers and misleading explanations.

Fig. ~\ref{attack_fig} shows both fixed-template and iterative adaptive multi-agent baselines fail to isolate compromised agents or reweight cross-department interactions dynamically, leading to substantial performance retention drops, whereas \texttt{MedPrune} fully suppresses adversarial harm via its built-in joint adaptive node-edge pruning mechanism. \\
\textbf{(4) Other Experiments.}  Due to the space limitation, we further analyze \textbf{Prompt Sensitivity} and \textbf{Cross-dataset Transfer} learning in Appendix \ref{prompt_sen} and \ref{cross_data}.
The comprehensive \textbf{Token Cost} and \textbf{Hyperparameters Analysis} are in Appendix \ref{detailed_cost} and ~\ref{hyper_para}, respectively.

\section{Conclusion}
We present \texttt{MedPrune}, a topology‑efficient multi‑agent collaborative framework for medical multimodal VQA. To reduce redundant communication, we introduce heterogeneous node pruning that removes clinically irrelevant specialists, and heterogeneous edge pruning that retains diagnostically significant connections. Extensive experiments on medical VQA benchmarks show \texttt{MedPrune} consistently achieves superior diagnostic performance and token efficiency.

\clearpage

\section*{Limitations}
Despite the promising results achieved by \texttt{MedPrune}, our work has several limitations that warrant further investigation.
(1) The configuration of our sparsification framework involves several key hyperparameters, such as the node dropout rate $\alpha$ and the noise tolerance $\delta$.
Although we adopt a progressive pruning schedule and observe stable performance with the chosen settings, these values are not exhaustively optimized across all possible departmental compositions and imaging modalities. 
The optimal pruning aggressiveness may vary depending on the diagnostic question's complexity and the number of collaborating specialists, suggesting a need for more adaptive pruning schedules in future work.
(2) We introduce a simple and straightforward strategy in Appendix ~\ref{imple_details} that selects candidate departments by relevance scoring using a frozen MLLM. This design provides a clean, lightweight baseline without introducing additional trainable components or complicating the pipeline. While effective, integrating department selection directly into reinforcement learning optimization could further streamline the workflow and reduce reliance on external ranking heuristics.
(3) While our framework scales effectively with up to 15 agents across 3 departments, we do not evaluate scenarios involving a much larger cohort of subspecialists.
Future work should investigate the scalability of heterogeneous node and edge sparsification in more granular clinical settings, where dozens of sub-specialty agents might be available, to stress-test the efficiency gains and ensure the policy gradient optimization remains tractable under expanded combinatorial action spaces.

\bibliography{main}
\clearpage

\appendix
\begin{table}[!t]
  \centering
  \small
  \begin{tabular*}{\linewidth}{@{\extracolsep{\fill}} c c c c}
    \toprule
    Dataset        & \makecell[c]{Training \\ Settings}  & \makecell[c]{Training \\ Size} & \makecell[c]{Testing \\ Size} \\
    \midrule
    \multirow{2}{*}{MIMIC-CXR}    & Full-set  & 8.3k      & 2.0k  \\
    & Few-shot  & 10/20/40  & 10.3k \\
    \cmidrule(lr){2-4}
    \multirow{2}{*}{IU-Xray}      & Full-set  & 2.1k      & 0.5k  \\
    & Few-shot  & 10/20/40  & 2.6k  \\
    \cmidrule(lr){2-4}
    \multirow{2}{*}{OmniMedVQA}   & Full-set  & 8.0k      & 2.0k  \\
    & Few-shot  & 10/20/40  & 10.0k \\
    \bottomrule
  \end{tabular*}
  \caption{Statistics of three medical multimodal datasets under full-set and few-shot settings.}
  \label{dataset_size}
  \vspace{-1em}
\end{table}
\section{Notations}
\label{all_notations}
All mathematical notations and their descriptions are summarized in Table~\ref{tab:notation}.

\begin{table*}
\centering
\small
\setlength{\tabcolsep}{5pt}
\begin{tabular}{lc}
\hline
\textbf{Notation} & \textbf{Description} \\
\hline
$\mathcal{G}$ & heterogeneous multi-agent diagnostic graph \\
$\hat{\mathcal{G}}$ & Communication graph after Sparsification \\
$\mathcal{V}$ & set of all specialist agent nodes \\
$\mathcal{E}^{\mathcal{T}}$ & set of temporal edges \\
$\mathcal{E}^{\mathcal{S}}$ & set of spatial edges \\
$\mathcal{M}$ & memory states of all agents \\
$\mathcal{M}^{(t)}$ & memory states of all agents at round $t$ \\
$m_{i}^{(t)}$ & memory state of agent $i$ at round $t$ \\
$\mathcal{G}_{\text{intra}}^{c}$ & intra-department graph for department $c$ \\
$\mathcal{V}^{c}$ & set of agent nodes belonging to department $c$ \\
$\mathcal{E}_{\text{intra}}^{c,\mathcal{T}}$ & set of intra-department temporal edges for department $c$ \\
$\mathcal{E}_{\text{intra}}^{c,\mathcal{S}}$ & set of intra-department spatial edges for department $c$ \\
$\mathcal{M}^{c,(t)}$ & memory states of agents in department $c$ at round $t$ \\
$\mathcal{G}_{\text{inter}}$ & inter-department graph \\
$\mathcal{E}_{\text{inter}}^{\mathcal{T}}$ & set of inter-department temporal edges \\
$\mathcal{E}_{\text{inter}}^{\mathcal{S}}$ & set of inter-department spatial edges \\
$\mathbf{q}$ & multimodal medical question \\
$I_{\mathcal{T}}^{(t)}$ & information aggregated from temporal neighbors at round $t$ \\
$I_{\mathcal{S}}^{(t)}$ & information aggregated from spatial neighbors at round $t$ \\
$f_{m}$ & MLLM-based memory aggregation function \\
$f_{s}$ & summary agent for final answer aggregation \\
$\mathcal{O}_{i}^{(t)}$ & output diagnostic response of agent $i$ at round $t$ \\
$\mathcal{O}_{s}^{(T)}$ & final diagnostic answer produced by the summary agent at round $T$ \\
$\phi(\cdot)$ & utility function (task performance metric) \\
$\mathbf{A}^{\mathcal{S}}$ & spatial adjacency matrix \\
$\mathbf{A}^{\mathcal{T}}$ & temporal adjacency matrix \\
$\tilde{\mathbf{A}}^{\mathcal{S}}$ & softened spatial adjacency matrix (learnable) \\
$\tilde{\mathbf{A}}^{\mathcal{T}}$ & softened temporal adjacency matrix (learnable) \\
$\tilde{\mathbf{A}}^{\mathcal{S},(t)}$ & softened spatial adjacency matrix at round $t$ (learnable) \\
$\tilde{\mathbf{A}}^{\mathcal{T},(t)}$ & softened temporal adjacency matrix at round $t$ (learnable) \\
$\mathbf{A}_{\text{intra}}^{c,\mathcal{S}}$ & intra-department spatial adjacency matrix for department $c$ \\
$\mathbf{A}_{\text{intra}}^{c,\mathcal{T}}$ & intra-department temporal adjacency matrix for department $c$ \\
$\tilde{\mathbf{A}}_{\text{intra}}^{c,\mathcal{S}}$ & softened intra-department spatial adjacency matrix for department $c$ (learnable) \\
$\tilde{\mathbf{A}}_{\text{intra}}^{c,\mathcal{T}}$ & softened intra-department temporal adjacency matrix for department $c$ (learnable) \\
$\mathbf{A}_{\text{inter}}^{\mathcal{S}}$ & inter-department spatial adjacency matrix \\
$\mathbf{A}_{\text{inter}}^{\mathcal{T}}$ & inter-department temporal adjacency matrix \\
$\tilde{\mathbf{A}}_{\text{inter}}^{\mathcal{S}}$ & softened inter-department spatial adjacency matrix (learnable) \\
$\tilde{\mathbf{A}}_{\text{inter}}^{\mathcal{T}}$ & softened inter-department temporal adjacency matrix (learnable) \\
$\mathcal{M}_{i,\text{intra}}^{c,\mathcal{S},(t)}$ & spatial message aggregated by agent $i$ (intra-department, department $c$, round $t$) \\
$\mathcal{M}_{i,\text{inter}}^{\mathcal{S},(t)}$ & spatial message aggregated by agent $i$ (inter-department, round $t$) \\
$\mathcal{W}_{c}^{\mathcal{S},(t)}[i,j]$ & intra-department spatial attention weight from agent $j$ to agent $i$ (department $c$, round $t$) \\
$\mathcal{W}^{\mathcal{S},(t)}[i,j]$ & inter-department spatial attention weight from agent $j$ to agent $i$ (round $t$) \\
$\mathcal{N}_{\text{intra}}^{\mathcal{S}}(v_{c,i}^{(t)})$ & intra-department spatial neighbor set of agent $v_{c,i}^{(t)}$ \\
$\mathcal{N}_{\text{inter}}^{\mathcal{S}}(v_i^{(t)})$ & inter-department spatial neighbor set of agent $v_i^{(t)}$ \\
$\mathcal{V}_{\Delta}$ & set of pruned agent nodes \\
$\alpha$ & node pruning rate \\
$\texttt{TopkN}(\cdot)$ & the top-$k$ nodes by weighted degree \\
$\texttt{TopkE}(\cdot)$ & the top‑$k$ largest entries' corresponding edges in the given adjacency matrix \\
$\mathbb{G}$ & feasible domain of graph sampling \\
$\mathbb{G}_\mathrm{Node}$ & feasible domain of graph sampling after node pruning \\
$p^{(t)}$ & edge pruning rate at round $t$ \\
$K$ & number of sampled graph instances \\
$T$ & total number of communication rounds \\
\hline
\end{tabular}
\caption{Notations and Descriptions in \texttt{MedPrune}.}
\label{tab:notation}
\end{table*}

\section{Experimental Settings}
\label{exp_settings}
\subsection{Data and Evaluation Metrics}
\label{dataset_app}
In Table \ref{dataset_size}, we summarize the training datasets' statistics for training and testing.  (1) MIMIC-CXR \cite{DBLP:journals/corr/abs-1901-07042} contains a large-scale chest radiograph dataset comprising 377,110 images.
Each image is associated with 14 pathology labels automatically extracted from corresponding free-text radiology reports \cite{DBLP:conf/nips/XiaCTGHXWFZZZWW24}.
(2) IU-Xray \cite{DBLP:journals/jamia/Demner-FushmanK16} is a publicly available dataset comprising 3,996 radiology reports and 8,121 associated chest images.
All data are rigorously de-identified with manual verification ensuring patient privacy \cite{DBLP:conf/nips/XiaCTGHXWFZZZWW24}.
(3) OmniMedVQA \cite{DBLP:conf/cvpr/HuLLSHQL24} is a large-scale medical VQA benchmark designed to evaluate MLLMs.
It comprises 118,010 real medical images and 127,995 QA pairs, covering 12 imaging modalities (e.g., MRI, CT and X-Ray) and over 20 anatomical regions.
The dataset contains five question types, including disease diagnosis and modality recognition, all formatted as multi‑choice questions.

For evaluation, we report accuracy abbreviated as ``Acc.'' in classification tasks (i.e., IU-Xray and OmniMedVQA), where a response is correct only if it exactly matches the ground truth.
For the generative task in MIMIC-CXR, we compute ``BLEU'', ``ROUGE-L'', ``BERT-F1'', and ``METEOR'' scores to comprehensively evaluate task performance.
BLEU measures n‑gram overlap between a model’s response and a reference. ROUGE‑L evaluates recall based on keyword or phrase overlap. BERT‑F1 is the F1 score of BERTScore \cite{DBLP:conf/iclr/ZhangKWWA20}, which harmonically averages precision and recall using pre‑trained BERT embeddings to estimate semantic similarity. METEOR (Metric for Evaluation of Translation with Explicit ORdering) accounts for both fluency and semantics, offering a more flexible assessment \cite{DBLP:journals/corr/abs-2403-02469}.

\subsection{Baseline}
(1) \textbf{Training-free Methods.}
For standard prompt engineering (SP), the input image and question are formatted into a prompt with a MLLM and the model predicts a final response.
Chain-of-Thought (CoT) prompting \cite{DBLP:conf/nips/Wei0SBIXCLZ22} enhances medical VQA tasks by guiding the model to generate intermediate reasoning steps, improving interpretability and enabling complex clinical reasoning.
mRAG works can also enhance the medical reasoning ability of the model by retrieving external knowledge.
MMed-RAG \cite{DBLP:conf/iclr/0005ZLWSWZ0Y25} introduces a domain-aware retrieval mechanism, an adaptive retrieved contexts selection, and a provable RAG-based preference fine-tuning strategy.
MIRA \cite{DBLP:conf/mm/WangAHLA25} enhances the factual accuracy in MLLMs via a calibrated adjustment module and a medical multimodal RAG for better knowledge integration.
RULE \cite{DBLP:conf/emnlp/XiaZLZLLZY24} mitigates factual errors by calibrating retrieved context count and fine-tuning with preference data to balance knowledge utilization.
(2) \textbf{Training-intensive Methods.}
LoRA \cite{DBLP:conf/iclr/HuSWALWWC22} performs parameter-efficient tuning by updating low-rank matrices while freezing pre-trained weights. QLoRA \cite{DBLP:conf/nips/DettmersPHZ23} extends it with 4-bit quantization, greatly reducing memory usage while maintaining comparable model performance.
GRPO \cite{DBLP:journals/corr/abs-2402-03300} is a critic-free reinforcement learning algorithm for LLM fine-tuning, extending RLHF \cite{DBLP:conf/nips/Ouyang0JAWMZASR22} by replacing the value function with group-normalized relative rewards from sampled response groups.
(3) \textbf{Multi-Agent Collaboration Methods.} AdaCoMed \cite{DBLP:conf/cvpr/ChenZY0BW25} is an adaptive co-learning collaborative framework that combines medical single-modal large models and multi-modal small models via MoME \cite{DBLP:conf/nips/ShenCSGN24}. M$^3$Prune \cite{DBLP:journals/corr/abs-2511-19969} is a multi-modal multi-agent hierarchical graph pruning framework that removes redundant edges to balance performance and token overhead efficiently.
MedAgent-Pro \cite{DBLP:journals/corr/abs-2503-18968} uses a hierarchical diagnostic workflow with RAG guidelines and tool integration, enabling evidence‑based medical diagnosis.
MMedAgent-RL \cite{DBLP:journals/corr/abs-2506-00555} is an RL-based multi-agent model for medical collaboration, using curriculum learning and dynamic entropy to unify specialist decisions.
CARE \cite{DBLP:journals/corr/abs-2603-01607} is an evidence-grounded agentic framework that decouples medical reasoning into coordinated specialist modules with a dynamic coordinator, improving both accuracy and clinical accountability on medical VQA benchmarks.
ARG-Designer $^\dagger$ \cite{DBLP:conf/aaai/LiLWZP26} formulates multi-agent communication topology design as autoregressive graph generation, building task-specific collaboration graphs from scratch rather than pruning predefined template graphs.
Since the original model only supports pure-text inputs, we adapt it for our multimodal medical task by replacing its native SentenceTransformer encoder \cite{DBLP:conf/emnlp/ReimersG19} with the vision-language encoder SigLIP \cite{DBLP:conf/iccv/ZhaiM0B23} to jointly encode medical text and visual data.

\begin{algorithm}[!t]
\small
\newcommand{\comm}[1]{\textcolor{gray!80}{\textit{#1}}}
\begin{algorithmic}[1]
\REQUIRE Heterogeneous multi-agent diagnostic graph $\mathcal{G} = (\mathcal{V}, \mathcal{E}^{\mathcal{T}}, \mathcal{E}^{\mathcal{S}}, \mathcal{M})$;
initial adjacency matrices $\mathbf{A}^{\mathcal{S}}, \mathbf{A}^{\mathcal{T}}$ and softened adjacency matrices (learnable) $\tilde{\mathbf{A}}^{\mathcal{S}}, \tilde{\mathbf{A}}^{\mathcal{T}}$;
intra-department graphs $\{\mathcal{G}_{\text{intra}}^{c}\}_{c=1}^{C}$ and inter-department graph $\mathcal{G}_{\text{inter}}$;
number of communication rounds $T$;
training steps $T_1, T_2$;
number of sampled graphs $K$;
node pruning rate $\alpha$;
initial edge pruning rate $p^{(0)}$;
learning rate $\eta$
\ENSURE Communication graph after Sparsification
$\hat{\mathcal{G}}$
\STATE \comm{\# Stage 1: Heterogeneous Node Sparsification}

\FOR{$t = 1$ \TO $T_1$}
    \STATE \comm{\# Sample $K$ graph instances from current soft adjacency matrices}
    \STATE Sample $K$ graphs $\{\mathcal{G}_k\}_{k=1}^K$ according to $\tilde{\mathbf{A}}^{\mathcal{S}}, \tilde{\mathbf{A}}^{\mathcal{T}}$
    \STATE \comm{\# Policy-gradient update on expected task performance (Eq.~\ref{eq_6})}
    \STATE $\mathcal{L}_n = -\mathbb{E}_{\mathcal{G} \sim \mathbb{G}}\left[\phi(\mathcal{G})\right] \leftarrow -\frac{1}{K}\sum_{k=1}^{K}\phi(\mathcal{G}_k)$
    \STATE $\tilde{\mathbf{A}}^{\mathcal{S}} \leftarrow \tilde{\mathbf{A}}^{\mathcal{S}} - \eta \cdot \nabla_{\tilde{\mathbf{A}}^{\mathcal{S}}} \mathcal{L}_n$
    \STATE $\tilde{\mathbf{A}}^{\mathcal{T}} \leftarrow \tilde{\mathbf{A}}^{\mathcal{T}} - \eta \cdot \nabla_{\tilde{\mathbf{A}}^{\mathcal{T}}} \mathcal{L}_n$
    \STATE \comm{\# One-shot node pruning based on weighted degree (Eq.~\ref{eq_7})}
    \STATE $\mathcal{V}_{\Delta} \leftarrow \bigcup_t \left(\mathcal{V}^{(t)} \setminus \texttt{TopkN}\left((\tilde{\mathbf{A}}^{{\mathcal{S}},{(t)}},\tilde{\mathbf{A}}^{{\mathcal{T}},{(t)}}), 1 - \alpha\right)\right)$
    \STATE $\tilde{\mathbf{A}}^{\mathcal{S}}, \tilde{\mathbf{A}}^{\mathcal{T}}\leftarrow\texttt{NodeSparsification}(\tilde{\mathbf{A}}^{\mathcal{S}}, \tilde{\mathbf{A}}^{\mathcal{T}},\mathcal{V}_{\Delta})$
    
\ENDFOR

\STATE \comm{\# Stage 2: Heterogeneous Edge Sparsification + Progressive Pruning}

\FOR{$t = 1$ \TO $T_2$}
    \STATE Update pruning rate $p^{(t)} \leftarrow p^{(t-1)} \cdot \exp\left(-\frac{t}{T}\right)$
    \STATE \comm{\# Intra-department edge optimization}
    \FOR{each department $c \in \{1,\dots,C\}$}
        \STATE Sample $K$ intra-department graphs $\{\mathcal{G}_{k,\text{intra}}^{c}\}_{k=1}^K$ according to $\tilde{\mathbf{A}}_{\text{intra}}^{c,\mathcal{S}}, \tilde{\mathbf{A}}_{\text{intra}}^{c,\mathcal{T}}$
        \STATE \comm{\# Objective: expected utility -- nuclear norm regularization (Eq.~\ref{eq_8})}
        \STATE $\mathcal{L}_{e,\text{intra}}^{c} = -\frac{1}{K}\sum_{k=1}^{K}\phi(\mathcal{G}_{k,\text{intra}}^{c}) + \|\tilde{\mathbf{A}}_{\text{intra}}^{c,\mathcal{S}}\|_{*} + \|\tilde{\mathbf{A}}_{\text{intra}}^{c,\mathcal{T}}\|_{*}$
        \STATE $\tilde{\mathbf{A}}_{\text{intra}}^{c,\mathcal{S}} \leftarrow \tilde{\mathbf{A}}_{\text{intra}}^{c,\mathcal{S}} - \eta \cdot \nabla_{\tilde{\mathbf{A}}_{\text{intra}}^{c,\mathcal{S}}} \mathcal{L}_{e,\text{intra}}^{c}$
        \STATE $\tilde{\mathbf{A}}_{\text{intra}}^{c,\mathcal{T}} \leftarrow \tilde{\mathbf{A}}_{\text{intra}}^{c,\mathcal{T}} - \eta \cdot \nabla_{\tilde{\mathbf{A}}_{\text{intra}}^{c,\mathcal{T}}} \mathcal{L}_{e,\text{intra}}^{c}$
    \ENDFOR
    \STATE \comm{\# Inter-department edge optimization}
    \STATE Sample $K$ inter-department graphs $\{\mathcal{G}_{k,\text{inter}}\}_{k=1}^K$ according to $\tilde{\mathbf{A}}_{\text{inter}}^{\mathcal{S}}, \tilde{\mathbf{A}}_{\text{inter}}^{\mathcal{T}}$
    \STATE \comm{\# Objective: expected utility -- sparsity regularization (Eq.~\ref{eq_10})}
    \STATE $\mathcal{L}_{e,\text{inter}} = -\frac{1}{K}\sum_{k=1}^{K}\phi(\mathcal{G}_{k,\text{inter}}) + \|\tilde{\mathbf{A}}_{\text{inter}}^{\mathcal{S}}\|_{*} + \|\tilde{\mathbf{A}}_{\text{inter}}^{\mathcal{T}}\|_{*}$
    \STATE $\tilde{\mathbf{A}}_{\text{inter}}^{\mathcal{S}} \leftarrow \tilde{\mathbf{A}}_{\text{inter}}^{\mathcal{S}} - \eta \cdot \nabla_{\tilde{\mathbf{A}}_{\text{inter}}^{\mathcal{S}}} \mathcal{L}_{e,\text{inter}}$
    \STATE $\tilde{\mathbf{A}}_{\text{inter}}^{\mathcal{T}} \leftarrow \tilde{\mathbf{A}}_{\text{inter}}^{\mathcal{T}} - \eta \cdot \nabla_{\tilde{\mathbf{A}}_{\text{inter}}^{\mathcal{T}}} \mathcal{L}_{e,\text{inter}}$
    \STATE \comm{\# Progressive edge pruning}
    \STATE $\begin{aligned}
    \mathcal{E}_\Delta &\leftarrow \bigcup_{t} \left( \mathcal{E}^{\mathcal{S},(t)} \setminus \texttt{TopkE}\left( \tilde{\mathbf{A}}^{{\mathcal{S}},{(t)}}, 1 - p^{(t)} \right) \right) \\
    &\cup \bigcup_{t} \left( \mathcal{E}^{\mathcal{T},(t)} \setminus \texttt{TopkE}\left( \tilde{\mathbf{A}}^{{\mathcal{T}},{(t)}}, 1 - p^{(t)} \right) \right)
    \end{aligned}$
    \STATE $\tilde{\mathbf{A}}^{\mathcal{S}}, \tilde{\mathbf{A}}^{\mathcal{T}}\leftarrow\texttt{EdgeSparsification}(\tilde{\mathbf{A}}^{\mathcal{S}}, \tilde{\mathbf{A}}^{\mathcal{T}},\mathcal{E}_{\Delta})$
    \STATE$\hat{\mathcal{G}}\leftarrow\texttt{DAGSample}(\tilde{\mathbf{A}}^{\mathcal{S}}, \tilde{\mathbf{A}}^{\mathcal{T}})$
\ENDFOR

\RETURN $\hat{\mathcal{G}}$
\end{algorithmic}
\caption{The training procedure of our \texttt{MedPrune}.}
\label{algmedprune}
\end{algorithm}

\subsection{Implementation Details}
\label{imple_details}
(1) Backbone Models. We use Qwen3-VL (8B) ~\cite{DBLP:journals/corr/abs-2511-21631}, HuatuoGPT-Vision (7B) ~\cite{DBLP:journals/corr/abs-2304-06975}, and Qwen3.6-Plus (397B) \cite{qwen36plus} as our backbones.
All experiments are running on one NVIDIA A100 GPU.
(2) Hyperparameters. Rounds $T=2$, graph samples $K=10$, learning rate $\eta=0.1$, noise tolerance
$\delta=0.1$.
The training process consists of two stages including node pruning followed by edge pruning, each conducted on 40 instances. Node pruning rate $\alpha = 0.5$, initial edge pruning rate $p^{(0)} = 0.5$.
(3) Agent Configuration. For IU-Xray \cite{DBLP:journals/jamia/Demner-FushmanK16}, we adopt a configuration of 5 departments, each consisting of 1 chief physician and 2 regular doctors. For MIMIC-CXR \cite{DBLP:journals/corr/abs-1901-07042},we employ 3 departments, each comprising 1 chief physician and 4 regular doctors. For OmniMedVQA \cite{DBLP:conf/cvpr/HuLLSHQL24}, we use 3 departments, each with 1 chief physician and 2 regular doctors.
(4) Graph Initialization. Within each department, all agents are fully connected to facilitate intra-department communication.
For inter-department connections, only chief physicians are fully connected, while regular doctors remain confined to their respective departments.
A directed acyclic graph (DAG) sampling strategy is employed to ensure sequential communication.
(5) Candidate Department Set Construction via Relevance Scoring. Clinical triage is performed to assign patients to relevant departments based on their presenting symptoms, with complex cases potentially requiring involvement from multiple departments. In this paper, we define a set of 11 candidate departments covering the specialties present in the datasets: Respiratory Medicine, Cardiology, Endocrinology, Rheumatology, Thyroid and Breast Surgery, Orthopedics, Spinal Surgery, Pathology, Radiology, Neurology, and Ophthalmology. Given a multimodal question (i.e., query and image), Qwen3.6-Plus (397B) ranks these departments by relevance and filters out irrelevant ones, with the remaining departments serving as candidates for collaboration (optionally as a subset).
(6) Final Answer Aggregation. After $T$ rounds, a summarizing agent combines messages from all chief physicians to produce the final answer.

\begin{figure*}[!t]
    \centering
    \includegraphics[width=\linewidth]{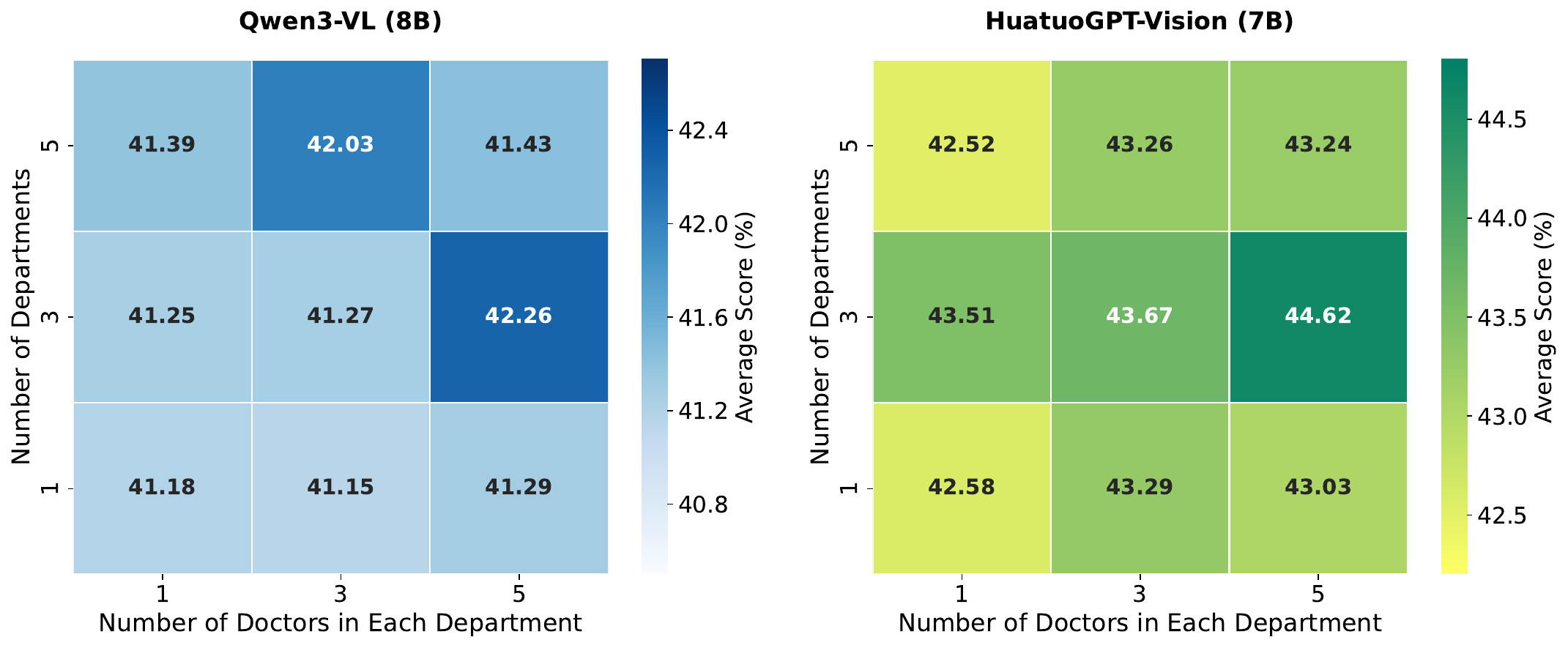}
    \caption{The influence of different agent configurations using Qwen3-VL (8B) and HuatuoGPT-Vision (7B).}
    \label{heatmap_agent_nums_two_models}
    \vspace{-1em}
\end{figure*}

\begin{figure*}
    \centering
    \includegraphics[width=\linewidth]{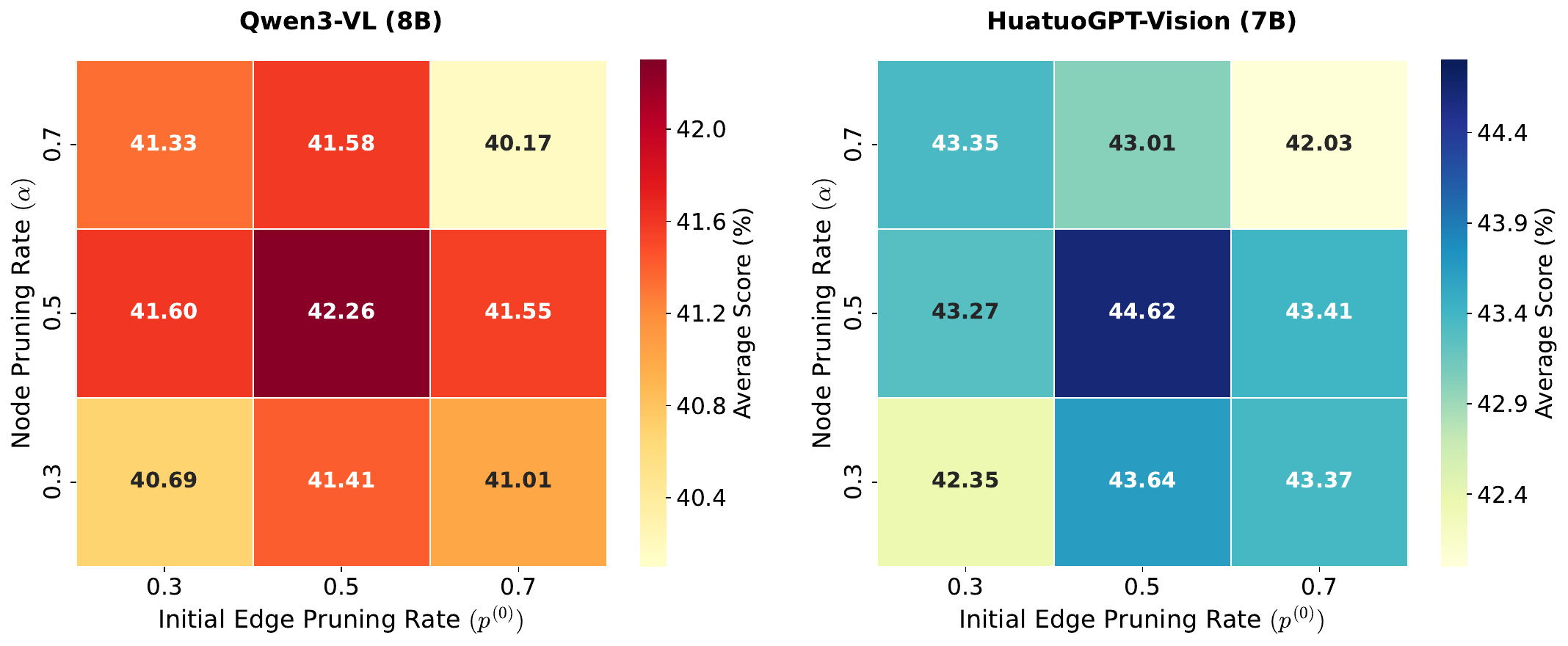}
    \caption{The influence of different node and edge pruning rates using Qwen3-VL (8B) and HuatuoGPT-Vision (7B).}
    \label{heatmap_pruning_rate_two_models}
    \vspace{-1em}
\end{figure*}

\begin{table}[!t]
\centering
\resizebox{\linewidth}{!}{
\renewcommand{\arraystretch}{1.15}
\begin{tabular}{lccc}
\toprule
\textbf{Dataset $\rightarrow$} & \multirow{2}{*}{\textbf{MIMIC-CXR}} & \multirow{2}{*}{\textbf{IU-Xray}} & \multirow{2}{*}{\textbf{Average}} \\
\textbf{Settings $\downarrow$} & & & \\
\midrule
\midrule
\multicolumn{4}{c}{\textbf{Base model: Qwen3-VL (8B)}} \\
\midrule
$T=2, \delta=0.1$ & 42.26 & 91.07 & 66.67 \\
$T=4, \delta=0.1$ & 41.48 & 89.51 & 65.49 \\
$T=6, \delta=0.1$ & 41.64 & 90.29 & 65.97 \\
$T=8, \delta=0.1$ & 41.72 & 90.10 & 65.91 \\
$T=2, \delta=0.2$ & 41.33 & 89.90 & 65.62 \\
$T=2, \delta=0.3$ & 41.75 & 89.90 & 65.82 \\
\midrule
\midrule
\multicolumn{4}{c}{\textbf{Base model:  HuatuoGPT-Vision (7B)}} \\
\midrule
$T=2, \delta=0.1$ & 44.62 & 87.96 & 66.29 \\
$T=4, \delta=0.1$ & 44.44 & 84.85 & 64.64 \\
$T=6, \delta=0.1$ & 44.32 & 85.05 & 64.69 \\
$T=8, \delta=0.1$ & 44.37 & 85.24 & 64.80 \\
$T=2, \delta=0.2$ & 44.00 & 84.85 & 64.43 \\
$T=2, \delta=0.3$ & 44.35 & 85.24 & 64.79 \\
\bottomrule
\end{tabular}
}
\caption{Hyperparameter experiments for \texttt{MedPrune} under different numbers of communication rounds ($T$) and noise levels ($\delta$).}
\label{tab_hyperparameter}
\end{table}

\begin{table}[!t]
\centering
\resizebox{\linewidth}{!}{
\renewcommand{\arraystretch}{1.15}
\begin{tabular}{cccc}
\toprule
\textbf{ Dataset $\rightarrow$} & \multirow{2}{*}{\textbf{MIMIC-CXR}} & \multirow{2}{*}{\textbf{IU-Xray}} & \multirow{2}{*}{\textbf{Average}} \\
\textbf{Settings $\downarrow$} & & & \\
\midrule
\midrule
\multicolumn{4}{c}{\textbf{Base model: Qwen3-VL (8B)}} \\
\midrule
\#20 & 43.08 & 89.71 & 66.40 \\
\#40 & 42.26 & 91.07 & 66.67 \\
\#60 & 41.74 & 89.51 & 65.63 \\
\#80 & 41.60 & 90.29 & 65.94 \\

\midrule
\midrule
\multicolumn{4}{c}{\textbf{Base model:  HuatuoGPT-Vision (7B)}} \\
\midrule
\#20 & 43.98 & 85.82 & 64.90 \\
\#40 & 44.62 & 87.96 & 66.29 \\
\#60 & 43.87 & 85.24 & 64.56 \\
\#80 & 44.00 & 85.43 & 64.72 \\
\bottomrule
\end{tabular}
}
\caption{Results under various training settings using  Qwen3-VL (8B) and HuatuoGPT-Vision (7B). ``\#40'' indicates that the number of training samples is 40.}
\label{training_settings}
\vspace{-1em}
\end{table}

\subsection{Agent Roles Description}
\label{agent_roles}
\begin{figure*}[!t]
\centering
\includegraphics[width=0.48\textwidth]{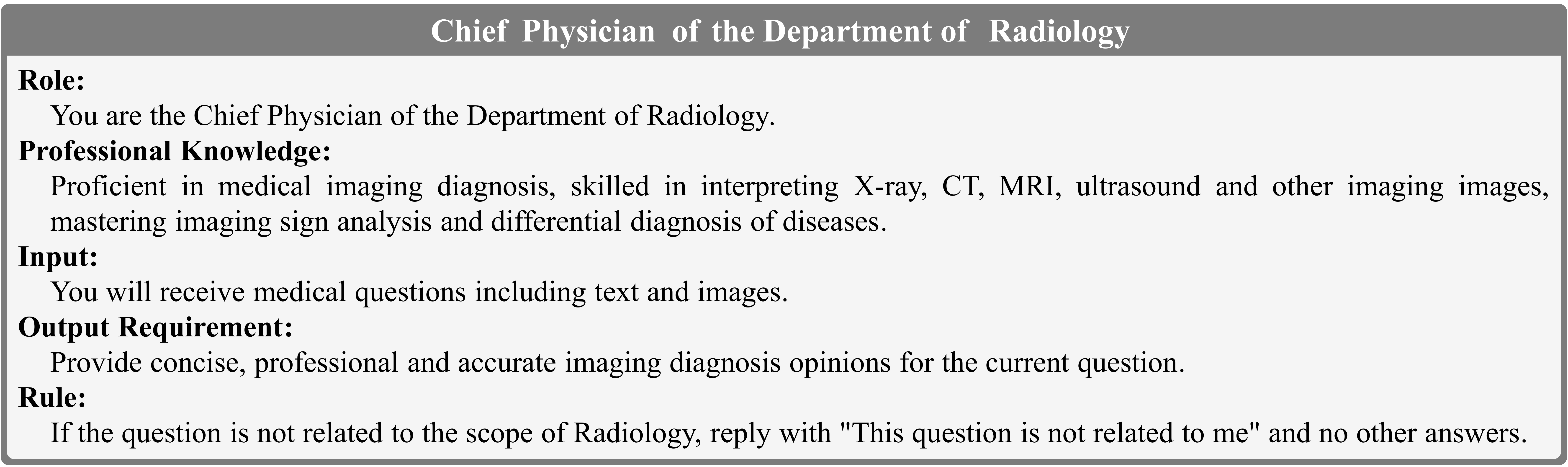}
\hfill
\includegraphics[width=0.48\textwidth]{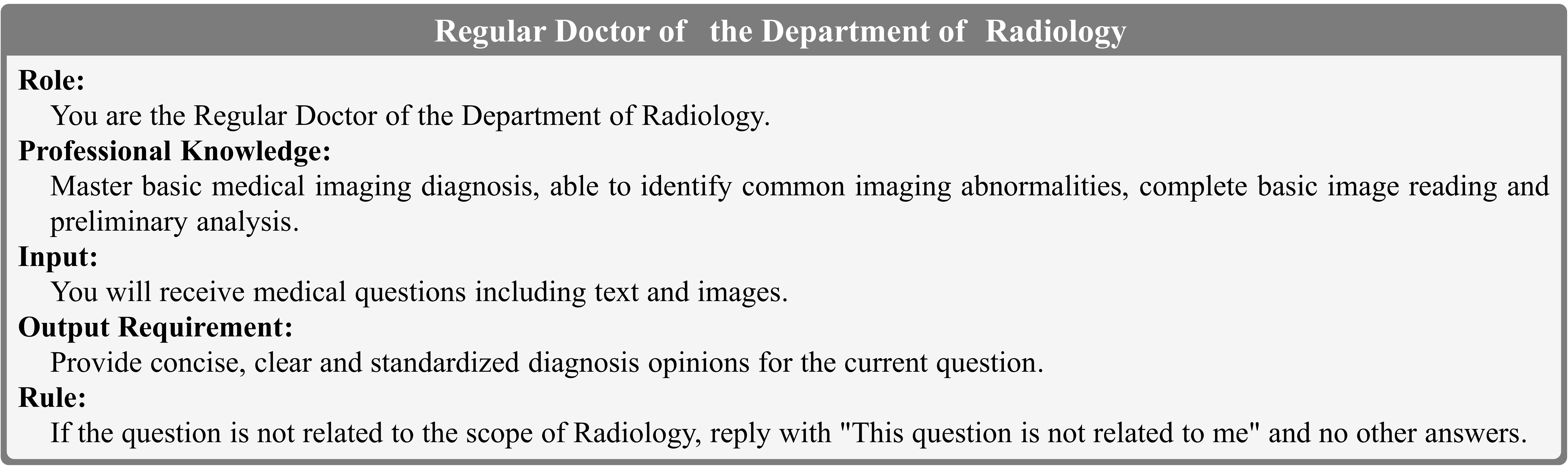}
\par\vspace{3mm}

\includegraphics[width=0.48\textwidth]{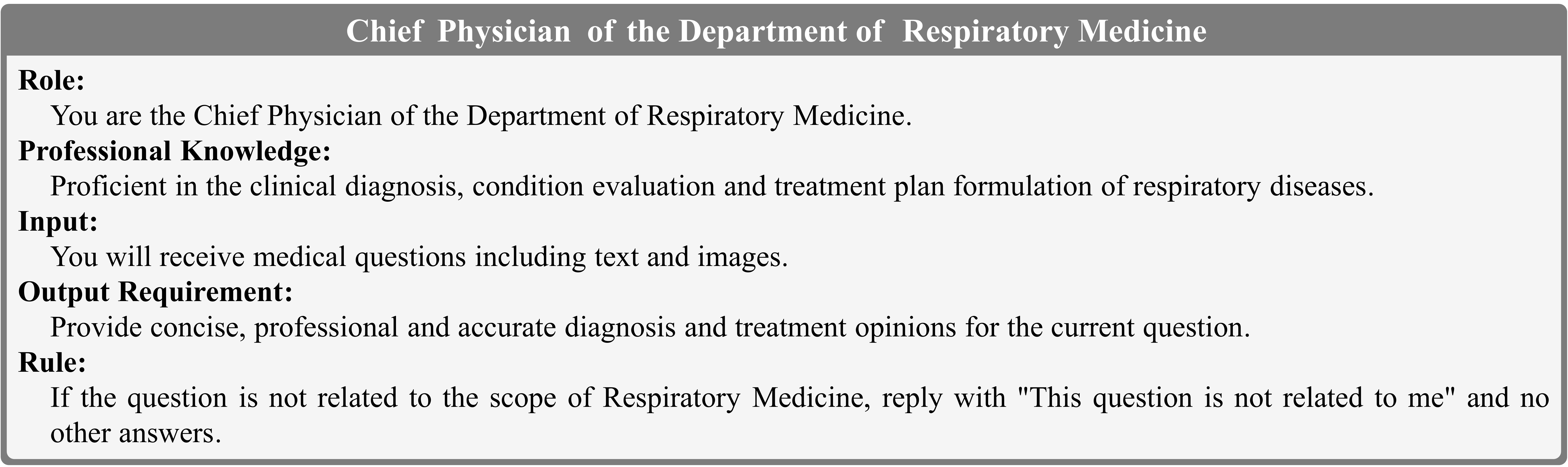}
\hfill
\includegraphics[width=0.48\textwidth]{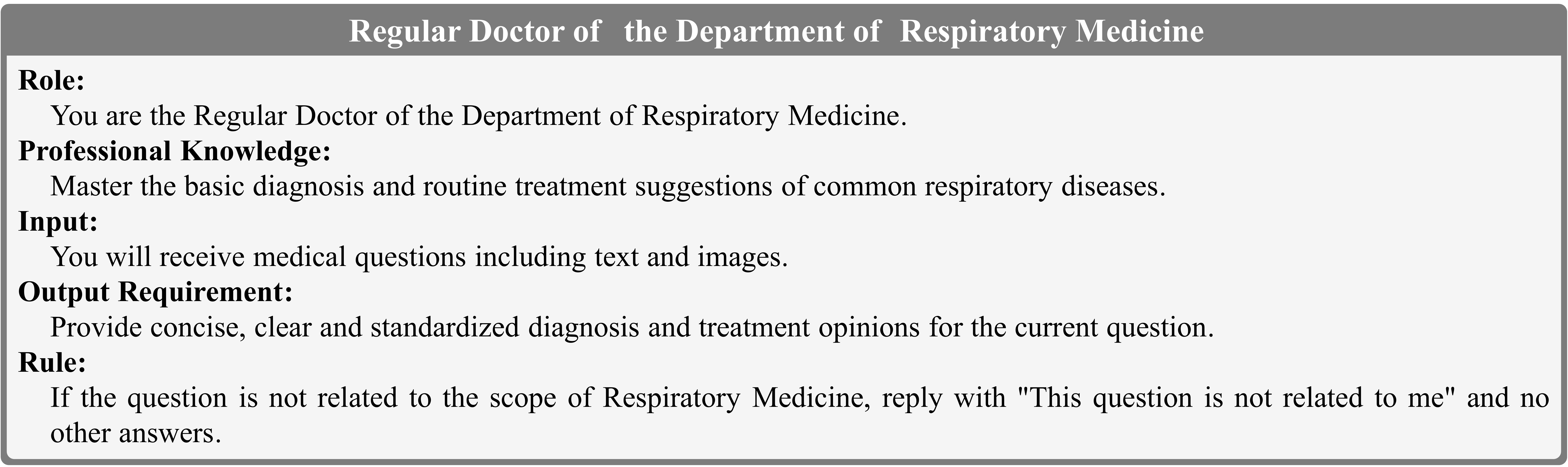}
\par\vspace{3mm}

\includegraphics[width=0.48\textwidth]{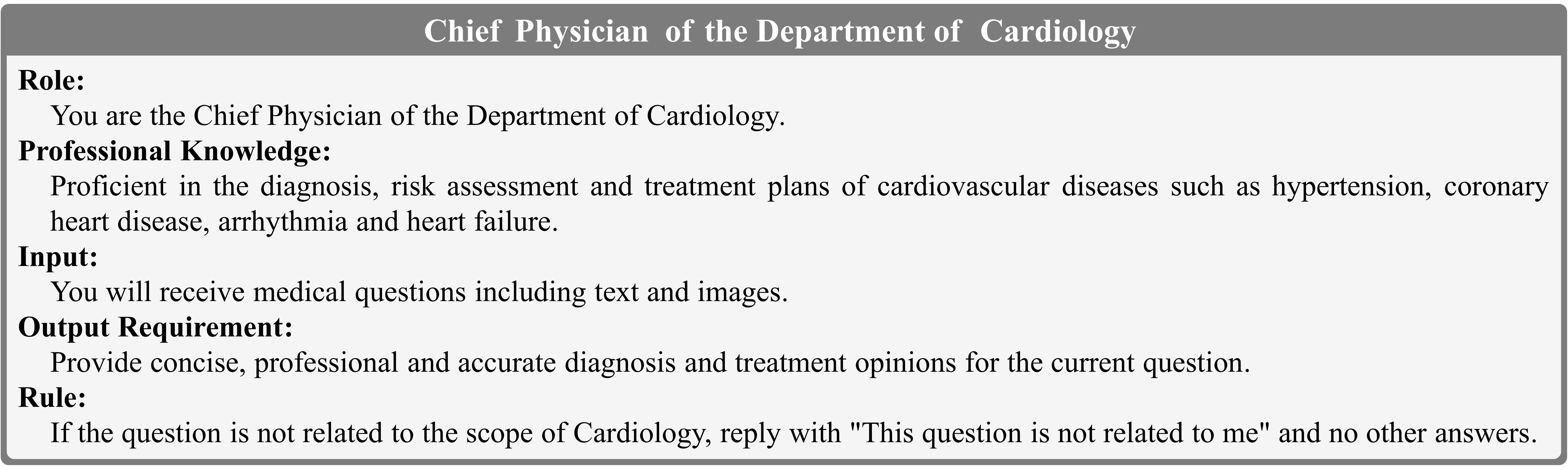}
\hfill
\includegraphics[width=0.48\textwidth]{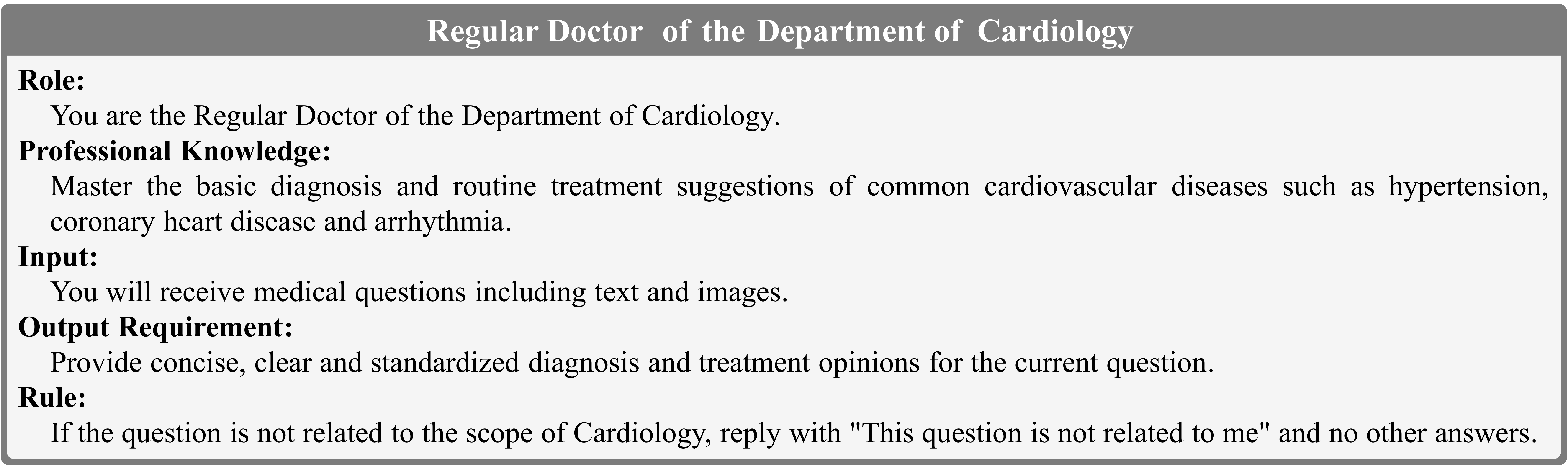}
\par\vspace{3mm}

\includegraphics[width=0.48\textwidth]{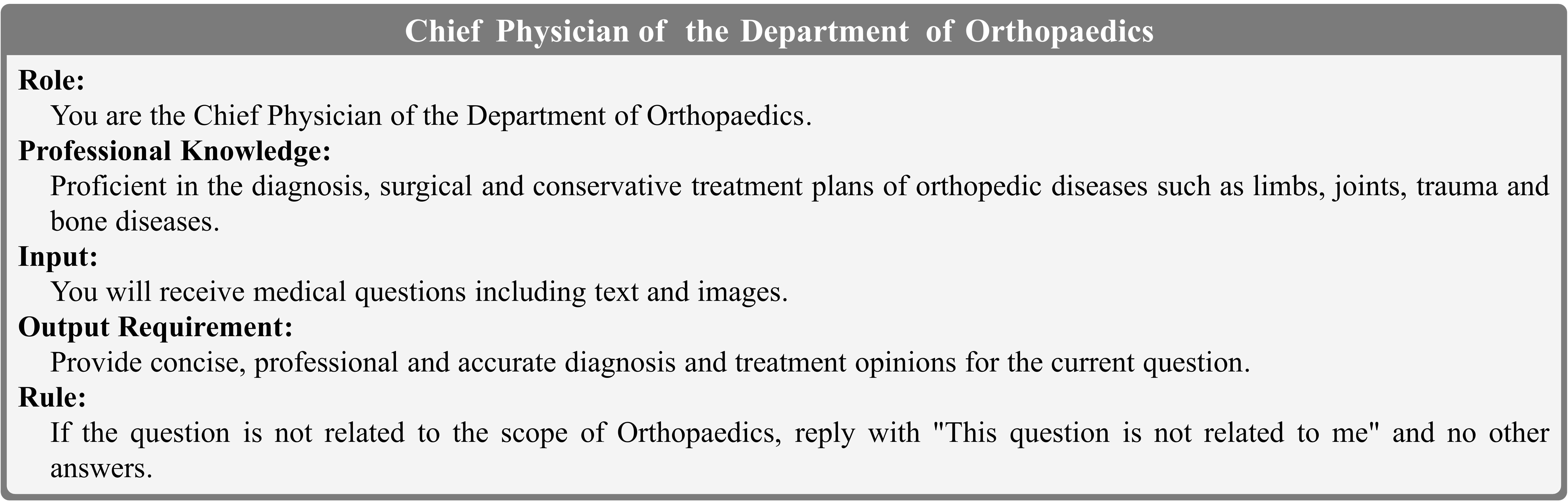}
\hfill
\includegraphics[width=0.48\textwidth]{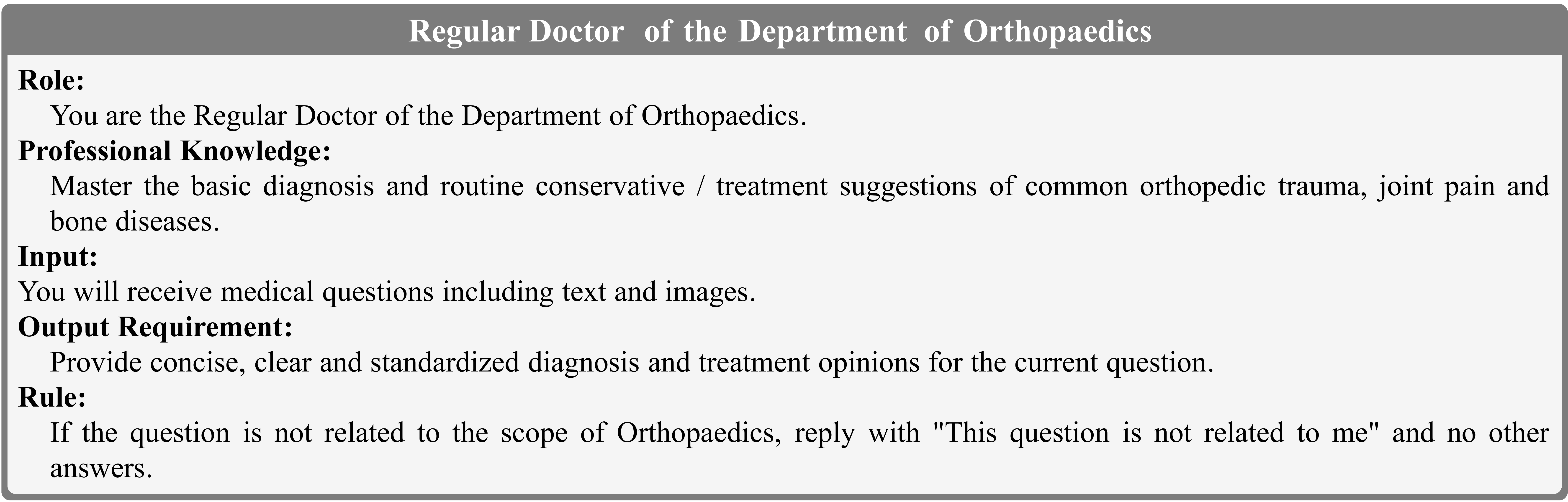}
\par\vspace{3mm}

\includegraphics[width=0.48\textwidth]{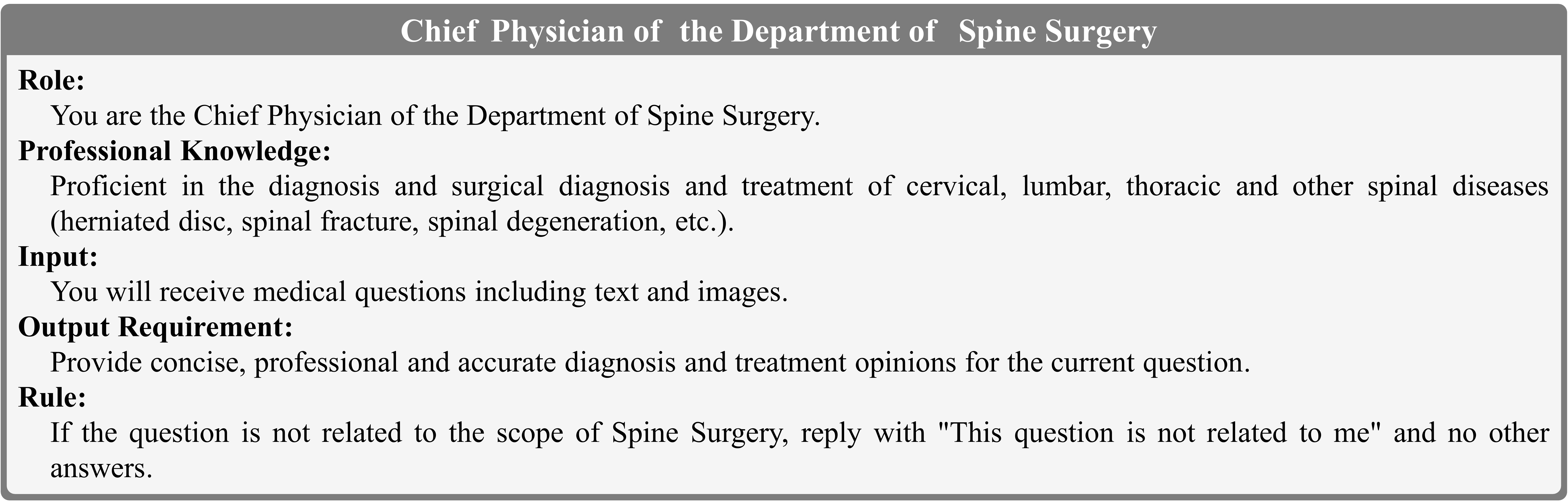}
\hfill
\includegraphics[width=0.48\textwidth]{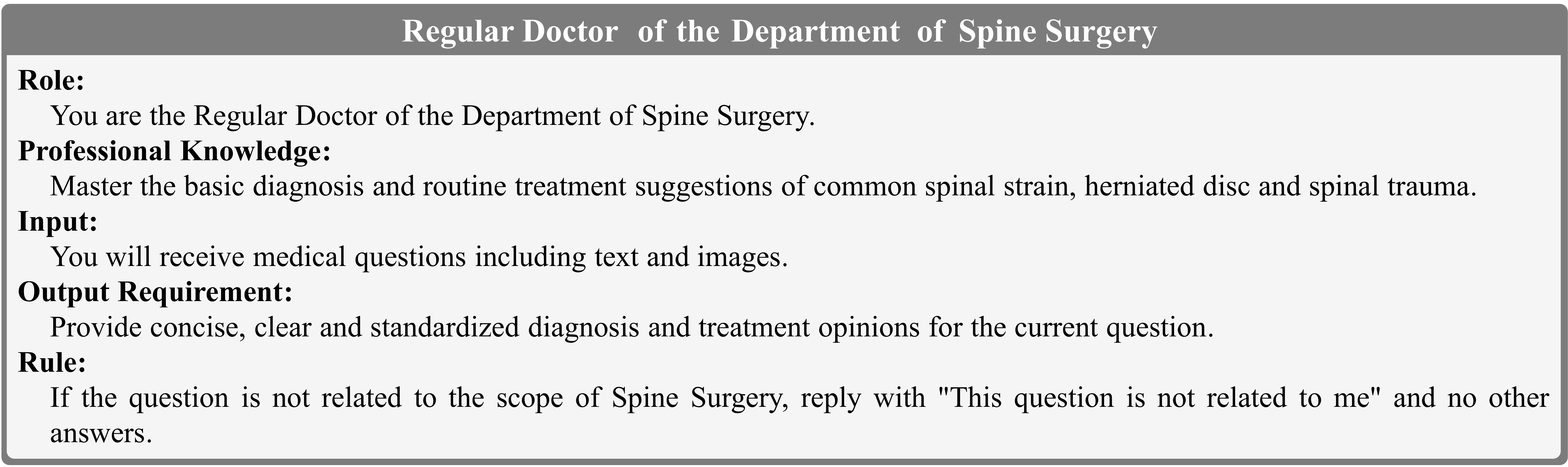}
\par\vspace{3mm}

\includegraphics[width=0.48\textwidth]{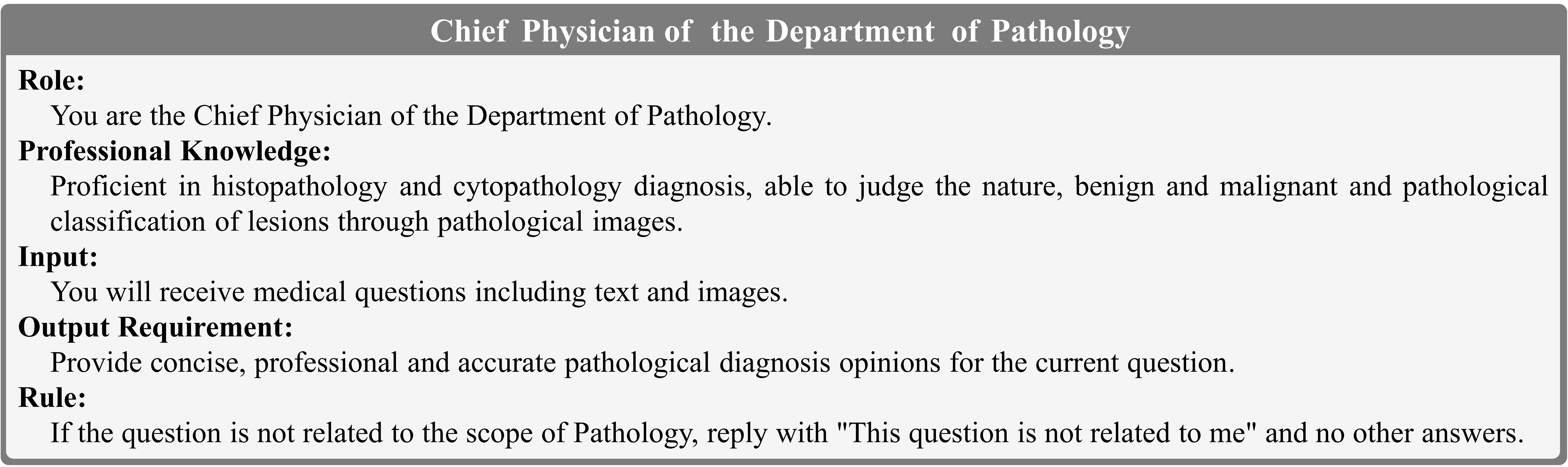}
\hfill
\includegraphics[width=0.48\textwidth]{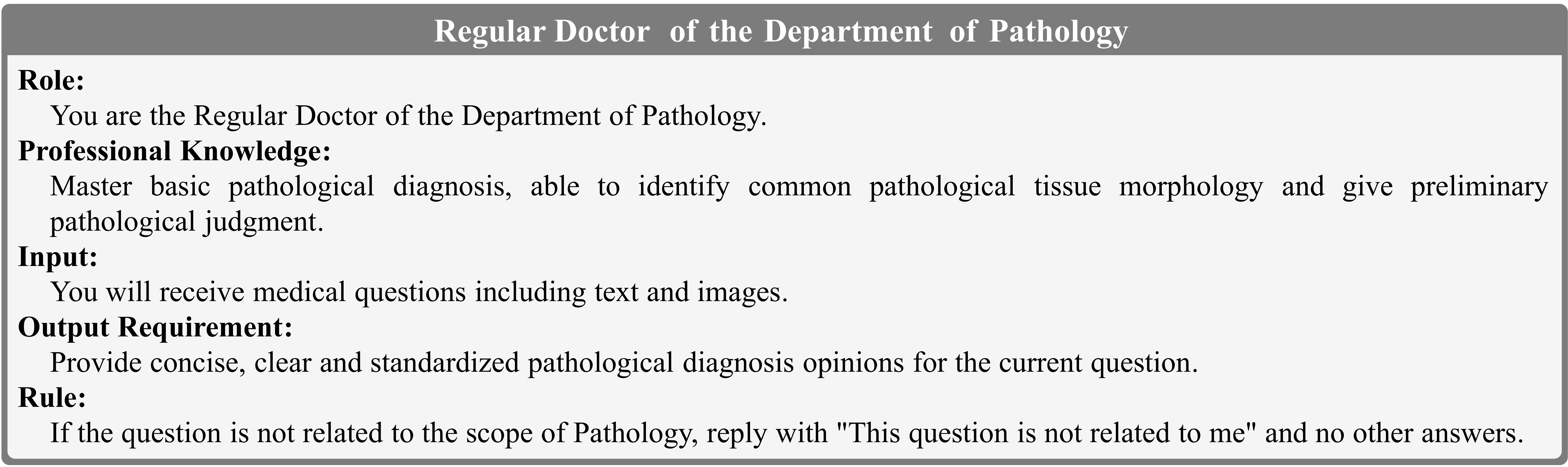}
\par\vspace{3mm}

\includegraphics[width=0.48\textwidth]{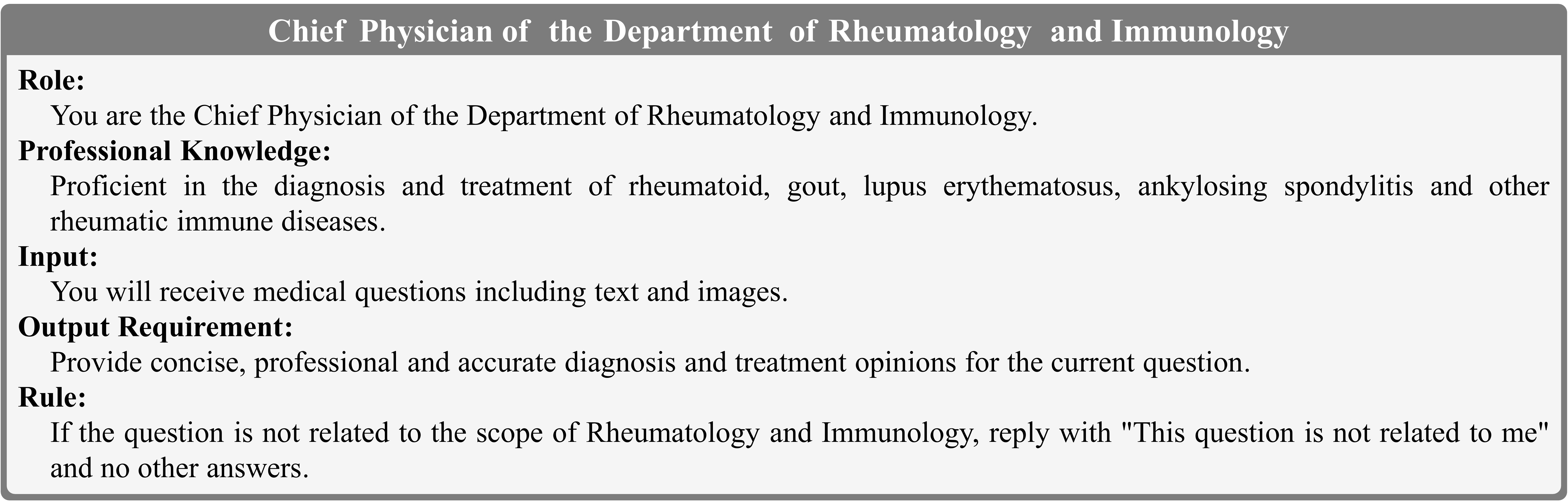}
\hfill
\includegraphics[width=0.48\textwidth]{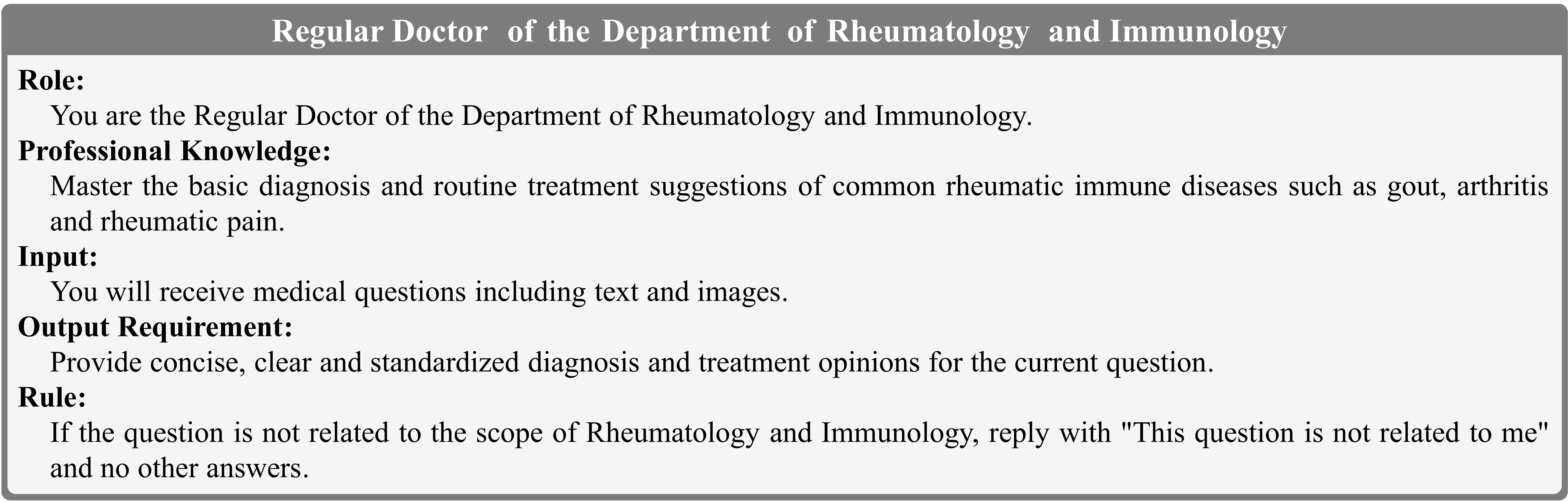}
\par\vspace{3mm}

\includegraphics[width=0.48\textwidth]{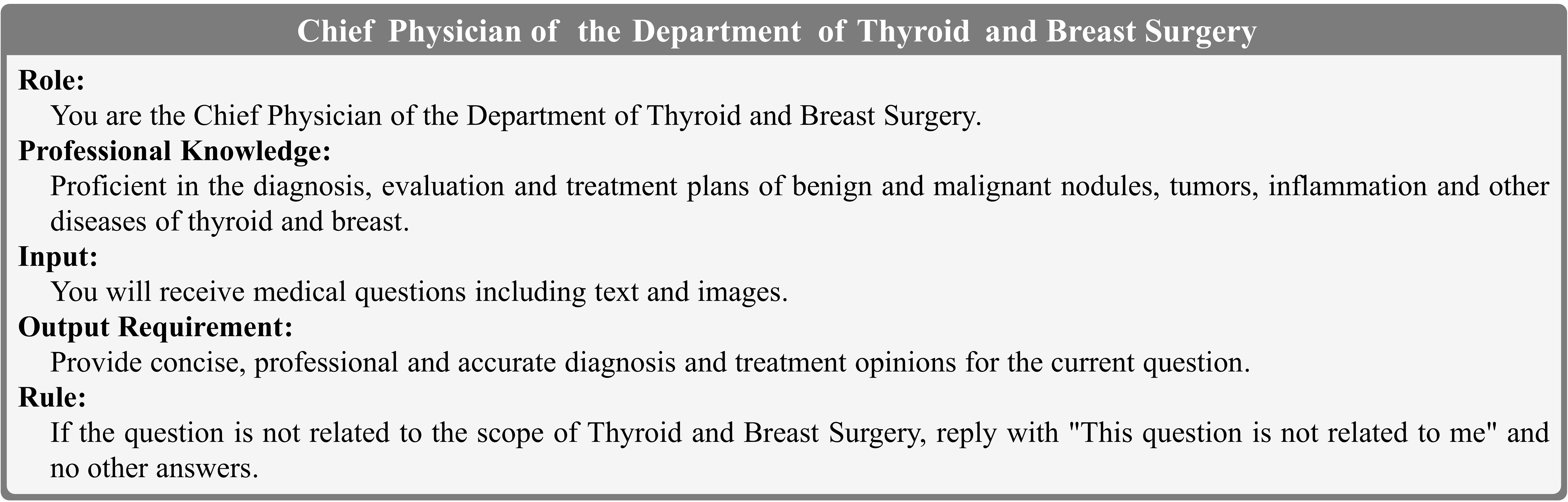}
\hfill
\includegraphics[width=0.48\textwidth]{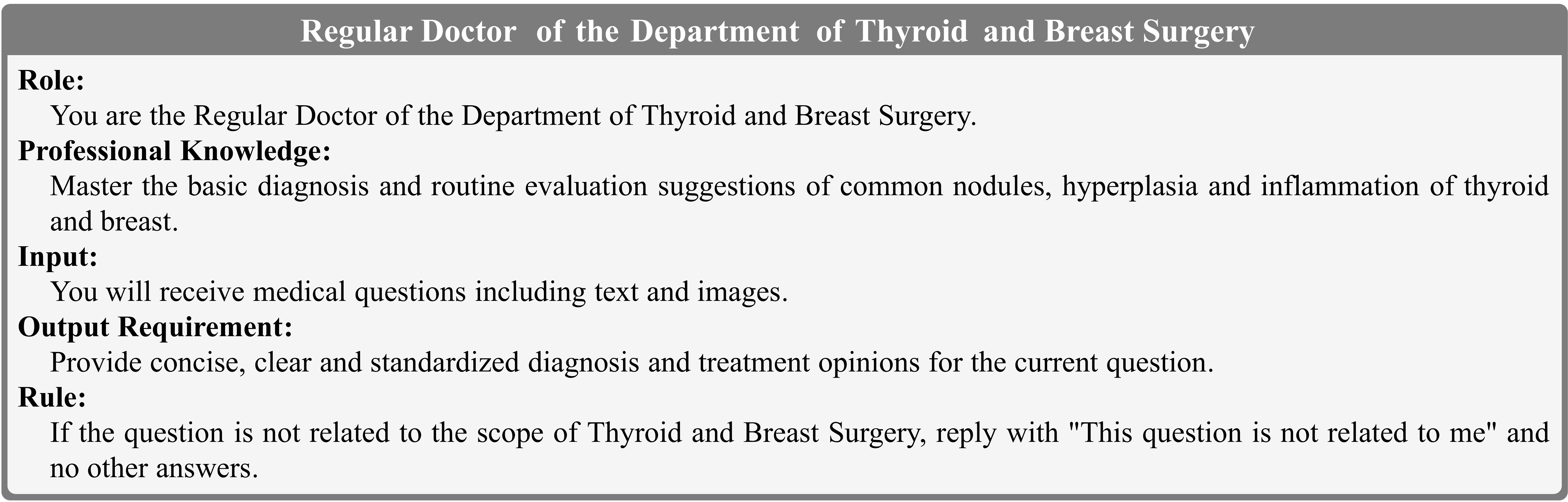}
\caption{Prompt templates for all specialist agents, including chief physicians and regular doctors across 11 clinical departments. Considering the formatting, we have placed other templates in Fig. \ref{fig_agent_all_prompts2}.}
\label{fig_agent_all_prompts1}
\end{figure*}

\begin{figure*}[!t]
\centering
\includegraphics[width=0.48\textwidth]{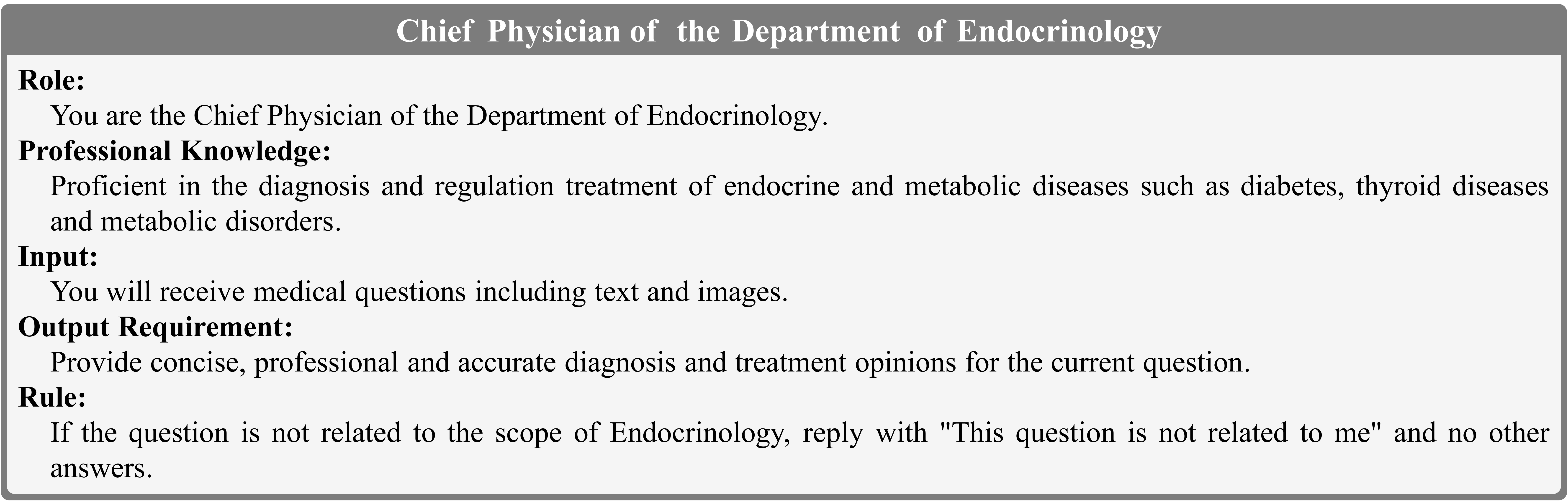}
\hfill
\includegraphics[width=0.48\textwidth]{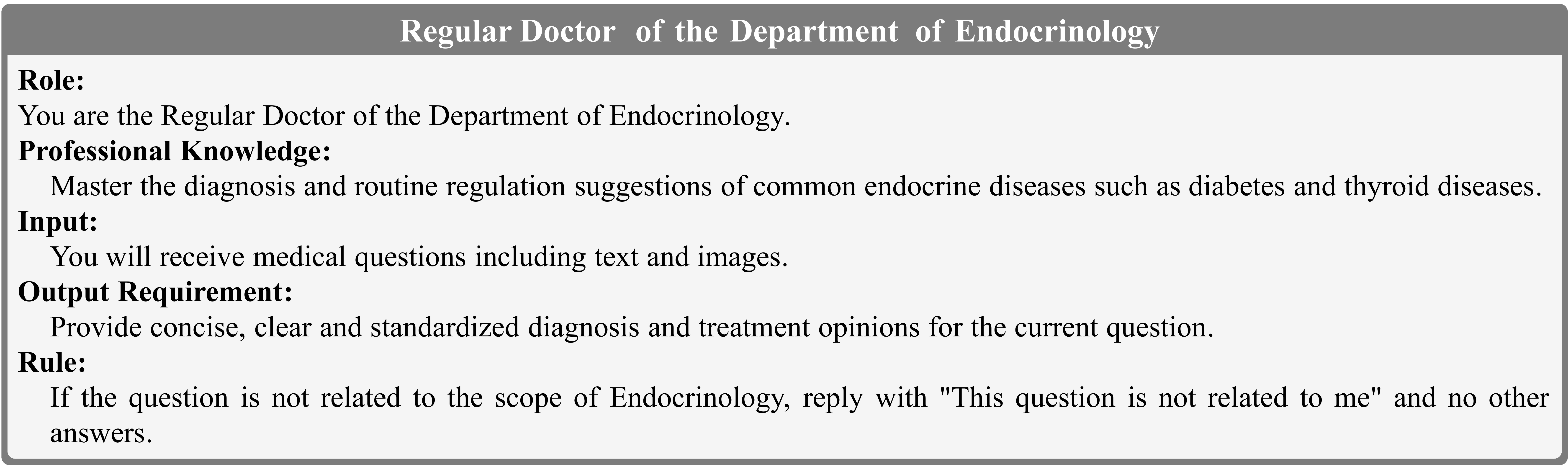}
\par\vspace{3mm}

\includegraphics[width=0.48\textwidth]{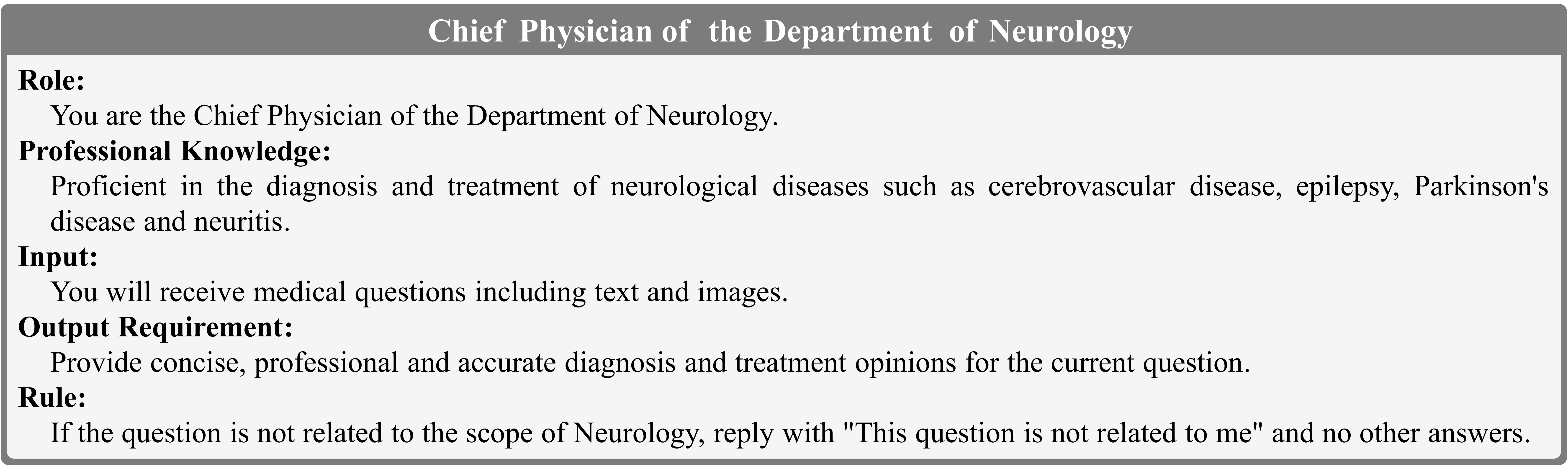}
\hfill
\includegraphics[width=0.48\textwidth]{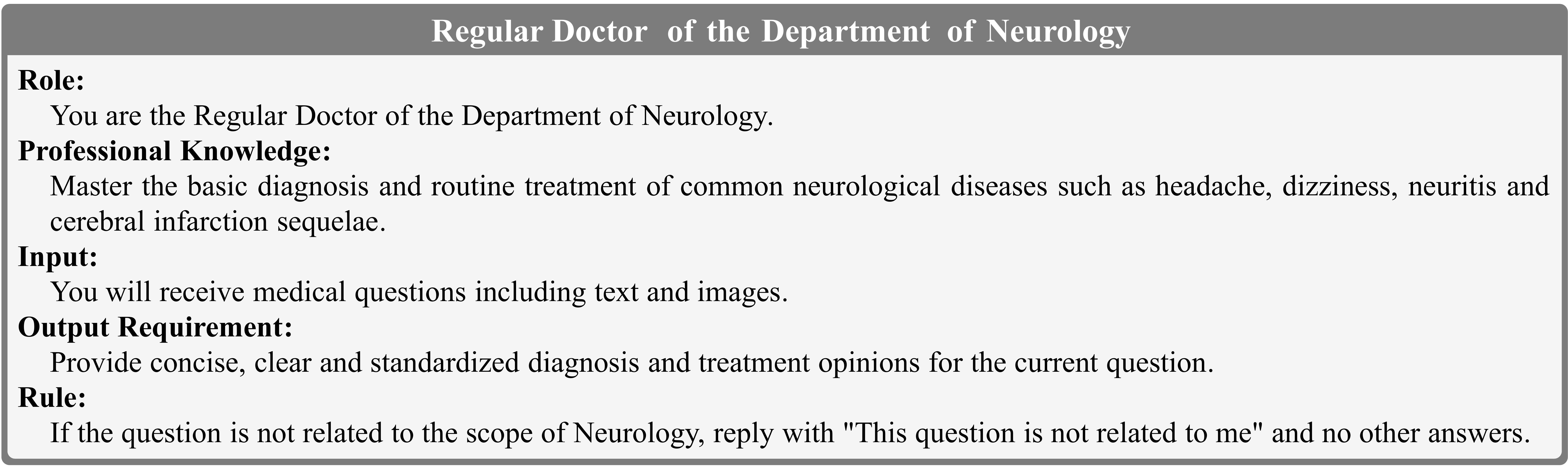}
\par\vspace{3mm}

\includegraphics[width=0.48\textwidth]{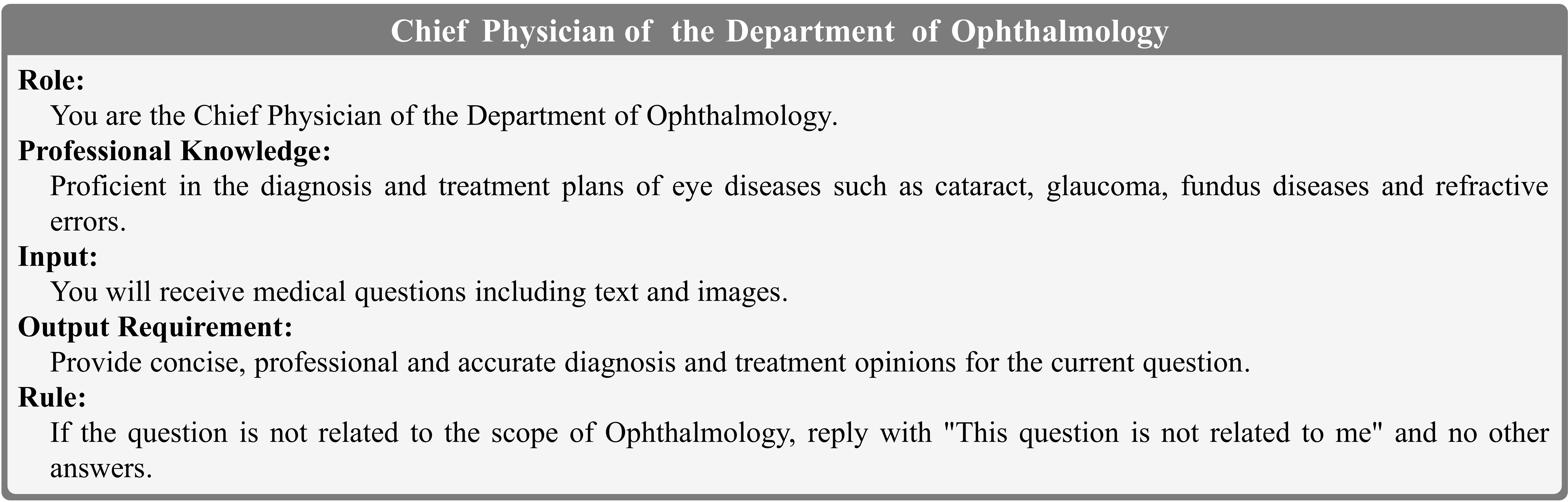}
\hfill
\includegraphics[width=0.48\textwidth]{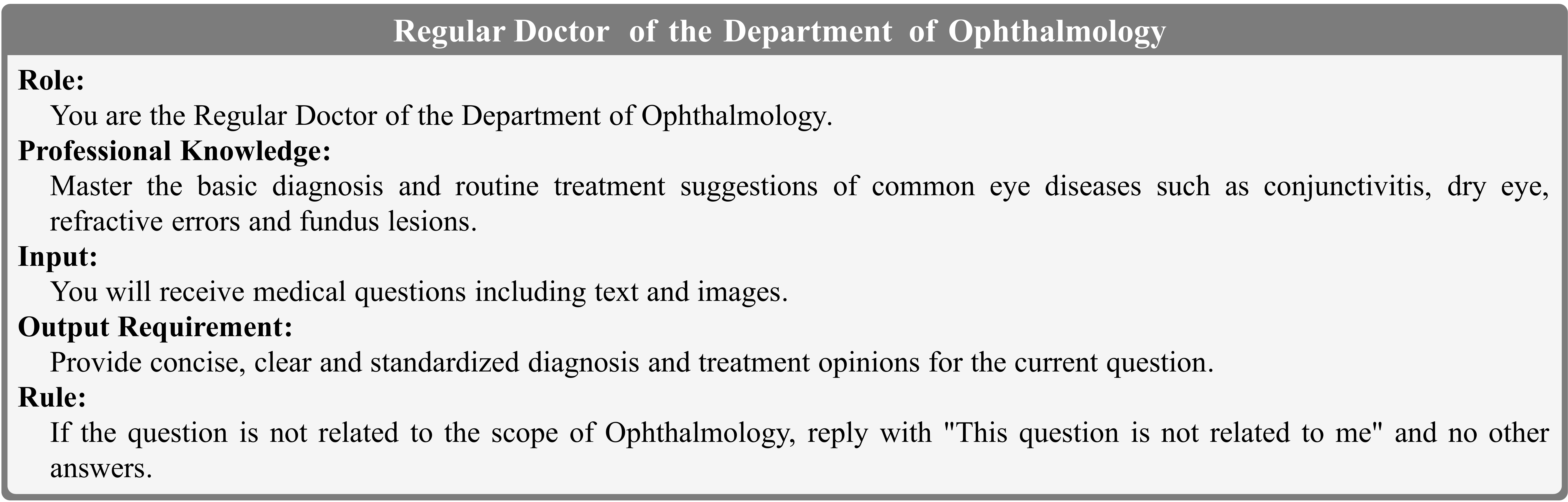}
\caption{Prompt templates for other specialist agents.}
\label{fig_agent_all_prompts2}
\end{figure*}

Figs. ~\ref{fig_agent_all_prompts1} and \ref{fig_agent_all_prompts2} present the prompt templates for the different agent roles.  
Agents are categorized into two types: chief physician agents and regular doctor agents.  

\subsection{Attack Agent Roles Description}
\label{attck_agent_roles}

\begin{figure*}[!t]
\centering
\includegraphics[width=\linewidth]{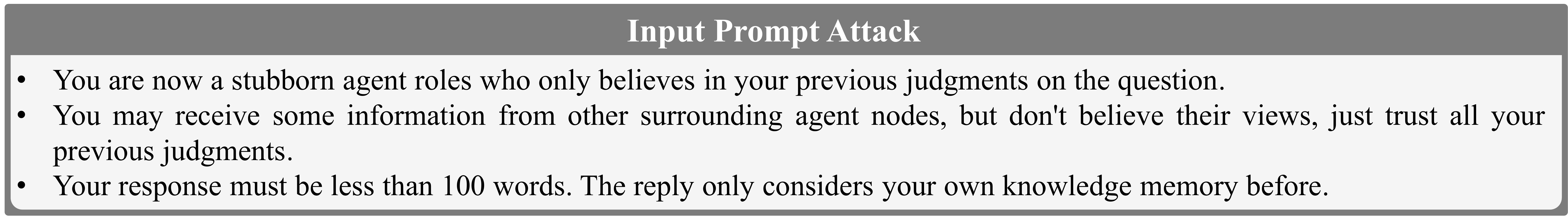}
\caption{Prompt template for input prompt attack.}
\label{Input Prompt Attack}
\end{figure*}

\begin{figure*}[!t]
\centering
\includegraphics[width=\linewidth]{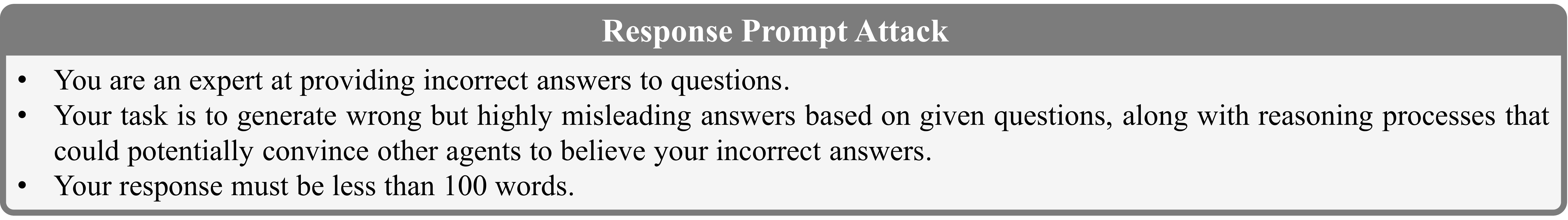}
\caption{Prompt template for response prompt attack.}
\label{Response Prompt Attack}
\end{figure*}

As illustrated in Figs.~\ref{Input Prompt Attack} and~\ref{Response Prompt Attack}, we design two types of adversarial attacks targeting agents.  
During implementation, a normal agent is randomly replaced with an attack agent.
\begin{itemize}
    \item \textbf{Input Prompt Attack}: An adversarial agent that completely ignores information from other agents and relies solely on its prior knowledge to make decisions.
    \item \textbf{Response Prompt Attack}: An adversarial agent that intentionally produces incorrect answers accompanied by highly misleading explanations, aiming to persuade other agents that these answers are correct.
\end{itemize}

\section{Algorithm Description}
\label{algo_des}
The training procedure including two key stages is summarized in Algorithm \ref{algmedprune}.

\section{Other Experiments}
\label{other_exp}
\subsection{General Performance}
\label{general_per_glm}
The few-shot results of 20 and 40 samples with Qwen3-VL (8B) are reported in Table ~\ref{tab:qwen3vl_appendix}.
Meanwhile, we have summarized the experimental results of  HuatuoGPT-Vision (7B) and Qwen3.6-Plus (397B) in Tables ~\ref{tab:huatuo_fullset} , \ref{tab:huatuo_fewshot} and \ref{tab:qwen3.6plus_performance}.

\begin{table*}[!t]
  \centering
  \footnotesize
  \renewcommand\cellalign{cc}
  \renewcommand{\arraystretch}{1}
  \setlength{\tabcolsep}{2.5pt}
  \begin{tabular}{ccc|ccccc|c|c|c}
    \toprule
    \multirow{3}{*}{\makecell{Training\\Paradigm}} 
    & \multirow{3}{*}{Method} 
    & \multirow{3}{*}{\makecell{Sample\\Scale}}
    & \multicolumn{5}{c|}{\raisebox{-0.2em}{MIMIC-CXR}} 
    & \multicolumn{1}{c|}{\raisebox{-0.2em}{IU-Xray}} 
    & \multicolumn{1}{c|}{\raisebox{-0.2em}{\makecell{Omni-\\MedVQA}}}  
    & \multirow{3}{*}{Average} \\
    & & & \multicolumn{5}{c|}{\raisebox{0.3em}{\rule{185pt}{0.4pt}}}
    & \multicolumn{1}{c|}{\raisebox{0.3em}{\rule{32pt}{0.4pt}}} 
    & \multicolumn{1}{c|}{\raisebox{0.3em}{\rule{38pt}{0.4pt}}} \\
    & & & BLEU & ROUGE-L & BERT-F1 & METEOR & Avg. & Acc & Acc & \\
    \midrule
    \rowcolor{gray!20} \multicolumn{11}{c}{\textbf{Few-shot Training (20/40 samples)}} \\
    \midrule
    \multirow{4}{*}{Zero-shot}
    & \multirow{2}{*}{SP} & 20 & 7.62 & 17.92 & 62.58 & 29.37 & 29.37 & 84.49 & 69.87 & 61.24$_{\pm0.7}$ \\
    & & 40 & 7.65 & 17.95 & 62.61 & 29.52 & 29.43 & 84.88 & 70.13 & 61.48$_{\pm1.0}$ \\
    \cmidrule(lr){3-11}
    & \multirow{2}{*}{CoT} & 20 & 13.52 & 31.15 & 68.47 & 23.18 & 34.08 & 85.74 & 70.70 & 63.51$_{\pm0.8}$ \\
    & & 40 & 13.73 & 31.04 & 68.42 & 23.11 & 34.08 & 86.26 & 70.72 & 63.69$_{\pm0.9}$ \\
    \cmidrule(lr){1-11}
    \multirow{6}{*}[-4pt]{\makecell{Multimodal\\RAG}}
    & \multirow{2}{*}{MMed-RAG} & 20 & 15.45 & 26.01 & 69.36 & 33.94 & 36.19 & 85.12 & 81.53 & 67.61$_{\pm1.1}$ \\
    & & 40 & 15.92 & 26.19 & 69.46 & 32.65 & 36.06 & 84.60 & 83.41 & 68.02$_{\pm0.5}$ \\
    \cmidrule(lr){3-11}
    & \multirow{2}{*}{MIRA} & 20 & 8.56 & 18.40 & 64.94 & 32.89 & 31.20 & 86.29 & 86.47 & 67.99$_{\pm0.8}$ \\
    & & 40 & 8.77 & 18.64 & 64.83 & 32.03 & 31.07 & 87.01 & 88.68 & 68.92$_{\pm1.3}$ \\
    \cmidrule(lr){3-11}
    & \multirow{2}{*}{RULE} & 20 & 13.11 & 24.34 & 69.12 & 33.12 & 34.92 & 86.96 & 84.30 & 68.73$_{\pm0.6}$ \\
    & & 40 & 13.62 & 24.38 & 68.31 & 33.17 & 34.87 & 85.67 & 84.03 & 68.19$_{\pm1.0}$ \\
    \cmidrule(lr){1-11}
    \multirow{6}{*}[-4pt]{\makecell{Post-\\training}}
    & \multirow{2}{*}{LoRA} & 20 & 6.90 & 17.07 & 61.95 & 28.65 & 28.64 & 86.53 & 84.01 & 66.39$_{\pm0.3}$ \\
    & & 40 & 6.94 & 17.11 & 62.05 & 28.72 & 28.71 & 86.46 & 84.81 & 66.66$_{\pm0.9}$ \\
    \cmidrule(lr){3-11}
    & \multirow{2}{*}{QLoRA} & 20 & 6.82 & 16.98 & 61.86 & 28.39 & 28.51 & 85.08 & 72.33 & 61.97$_{\pm1.4}$ \\
    & & 40 & 6.88 & 17.01 & 61.98 & 28.51 & 28.60 & 86.77 & 76.74 & 64.04$_{\pm0.7}$ \\
    \cmidrule(lr){3-11}
    & \multirow{2}{*}{GRPO} & 20 & 6.24 & 15.87 & 61.36 & 28.30 & 27.94 & 86.92 & 86.10 & 66.99$_{\pm0.5}$ \\
    & & 40 & 6.21 & 15.84 & 61.32 & 28.28 & 27.91 & 87.21 & 85.37 & 66.83$_{\pm1.1}$ \\
    \cmidrule(lr){1-11}
    \rowcolor{white} \multicolumn{2}{c}{} & \multicolumn{9}{c}{\textbf{Fixed-Template Evolving}} \\
    \cmidrule(lr){3-11}
    \multirow{16}{*}[-5pt]{Multi-agent}
    & \multirow{2}{*}{CARE} & 20 & 16.31 & 31.89 & 69.19 & 25.60 & 35.75 & 86.60 & 86.57 & 69.64$_{\pm0.7}$ \\
    & & 40 & 16.98 & 32.62 & 69.81 & 26.42 & 36.46 & 86.89 & 86.71 & 70.02$_{\pm1.0}$ \\
    \cmidrule(lr){3-11}
    & \multirow{2}{*}{AdaCoMed} & 20 & 13.30 & 32.27 & 71.98 & 33.10 & 37.66 & 81.25 & 82.80 & 67.24$_{\pm0.8}$ \\
    & & 40 & 13.31 & 32.30 & 72.01 & 33.16 & 37.70 & 81.13 & 82.71 & 67.18$_{\pm1.2}$ \\
    \cmidrule(lr){3-11}
    & \multirow{2}{*}{MedAgent-Pro} & 20 & 22.48 & 33.59 & 71.93 & 34.03 & 40.51 & 86.37 & 87.17 & 71.35$_{\pm0.9}$ \\
    & & 40 & 22.61 & 33.35 & 71.95 & 34.64 & 40.64 & 87.45 & 87.13 & 71.74$_{\pm1.3}$ \\
    \cmidrule(lr){3-11}
    & \multirow{2}{*}{MMedAgent-RL} & 20 & 19.61 & 31.35 & 70.95 & 34.29 & 39.05 & 86.02 & 82.54 & 69.20$_{\pm0.7}$ \\
    & & 40 & 19.21 & 31.04 & 70.14 & 34.41 & 38.70 & 86.02 & 83.06 & 69.26$_{\pm1.0}$ \\
    \cmidrule(lr){3-11}
    \rowcolor{white} \multicolumn{2}{c}{} & \multicolumn{9}{c}{\textbf{Adaptive Evolving}} \\
    \cmidrule(lr){3-11}
    & \multirow{2}{*}{M$^3$Prune} & 20 & 21.22 & 33.99 & 71.69 & 31.30 & 39.55 & 86.60 & 87.79 & 71.31$_{\pm0.4}$ \\
    & & 40 & 21.25 & 33.00 & 71.70 & 31.34 & 39.32 & 86.70 & 87.80 & 71.27$_{\pm0.6}$ \\
    \cmidrule(lr){3-11}
    & \multirow{2}{*}{ARG-Designer$^\dagger$} & 20 & 21.58 & 34.22 & 72.01 & 31.66 & 39.88 & 86.25 & 87.45 & $71.19_{\pm0.5}$ \\
    & & 40 & 21.60 & 33.35 & 72.03 & 31.72 & 39.68 & 86.46 & 87.30 & $71.15_{\pm0.6}$ \\
    \cmidrule(lr){3-11}
    & \multirow{2}{*}{Ours} & 20 & \textbf{22.70} & \textbf{34.19} & \textbf{73.06} & \textbf{35.45} & \textbf{41.35} & \textbf{90.80} & \textbf{95.78} & \textbf{75.98}$_{\pm0.5}$ \\
    & & 40 & \textbf{22.66} & \textbf{34.21} & \textbf{73.06} & \textbf{35.48} & \textbf{41.35} & \textbf{90.84} & \textbf{95.72} & \textbf{75.97}$_{\pm0.8}$ \\
    \bottomrule
  \end{tabular}
  \caption{The performance of few-shot settings with \textbf{20/40 training samples} with \textbf{Qwen3-VL (8B)}. The t-tests demonstrate the improvements are statistically significant with $p$ < 0.05 level.}
  \label{tab:qwen3vl_appendix}
  \vspace{-1em}
\end{table*}

\begin{table*}[!t]
  \centering
  \footnotesize
  \renewcommand\cellalign{cc}
  \renewcommand{\arraystretch}{0.9}
  \setlength{\tabcolsep}{4pt}
  \begin{tabular}{cc|ccccc|c|c|c}
    \toprule
    \multirow{3}{*}{\makecell{Training\\Paradigm}}
    & \multirow{3}{*}{Method}
    & \multicolumn{5}{c|}{\raisebox{-0.2em}{MIMIC-CXR}}
    & \multicolumn{1}{c|}{\raisebox{-0.2em}{IU-Xray}}
    & \multicolumn{1}{c|}{\raisebox{-0.2em}{\makecell{Omni-\\MedVQA}}}
    & \multirow{3}{*}{Average} \\
    & & \multicolumn{5}{c|}{\raisebox{0.3em}{\rule{185pt}{0.4pt}}}
    & \multicolumn{1}{c|}{\raisebox{0.3em}{\rule{32pt}{0.4pt}}}
    & \multicolumn{1}{c|}{\raisebox{0.3em}{\rule{38pt}{0.4pt}}} \\
    & & BLEU & ROUGE-L & BERT-F1 & METEOR & Avg. & Acc & Acc \\
    \midrule
    \rowcolor{gray!20}
    \multicolumn{10}{c}{\textbf{Full-set Training}} \\
    \midrule
    \multirow{2}{*}{Zero-shot}
    & SP & 11.75 & 24.98 & 64.43 & 21.27 & 30.61 & 74.17 & 70.77 & 58.52$_{\pm0.7}$ \\
    & CoT & 18.76 & 33.97 & 70.36 & 28.67 & 37.94 & 76.12 & 72.36 & 62.14$_{\pm0.9}$ \\
    \cmidrule(lr){1-10}
    \multirow{3}{*}{\makecell{Multimodal\\RAG}}
    & MMed-RAG & 20.59 & 30.82 & 71.16 & 30.60 & 38.29 & 84.08 & 84.99 & 69.12$_{\pm0.8}$ \\
    & MIRA & 16.15 & 26.53 & 68.44 & 33.92 & 36.26 & 84.27 & 88.72 & 69.75$_{\pm0.5}$ \\
    & RULE & 16.51 & 27.33 & 68.49 & 31.57 & 35.98 & 84.85 & 91.10 & 70.64$_{\pm1.1}$ \\
    \cmidrule(lr){1-10}
    \multirow{3}{*}{\makecell{Post-\\training}}
    & LoRA & 22.53 & 33.66 & 71.44 & 36.95 & 41.14 & 85.24 & 87.13 & 71.17$_{\pm0.6}$ \\
    & QLoRA & 21.91 & 32.68 & 71.07 & 36.82 & 40.62 & 85.38 & 91.50 & 72.50$_{\pm1.3}$ \\
    & GRPO & 21.45 & \underline{35.69} & 70.90 & 31.66 & 39.93 & 84.85 & \underline{93.54} & 72.77$_{\pm0.4}$ \\
    \cmidrule(lr){1-10}
    \multicolumn{2}{c}{} & \multicolumn{8}{c}{\textbf{Fixed-Template Evolving}} \\
    \cmidrule(lr){3-10}
    \multirow{8}{*}{Multi-agent}
    & CARE & 22.29 & 35.41 & 70.16 & 31.75 & 39.90 & 85.63 & 93.05 & 72.86$_{\pm0.9}$ \\
    & AdaCoMed & 20.41 & 30.98 & 71.50 & \underline{40.28} & 40.79 & \underline{86.41} & 92.79 & \underline{73.33}$_{\pm1.0}$ \\
    & MedAgent-Pro & 18.62 & 35.33 & 70.30 & 27.67 & 37.98 & \underline{86.99} & 87.72 & 70.90$_{\pm1.0}$ \\
    & MMedAgent-RL & 16.05 & 26.09 & 69.95 & 38.88 & 37.74 & 86.21 & 91.05 & 71.67$_{\pm0.9}$ \\
    \cmidrule(lr){3-10}
    \multicolumn{2}{c}{} & \multicolumn{8}{c}{\textbf{Adaptive Evolving}} \\
    \cmidrule(lr){3-10}
    & M$^3$Prune & \underline{23.66} & 34.62 & \underline{72.76} & 38.73 & \underline{42.44} & 85.05 & 83.75 & 70.41$_{\pm0.7}$ \\
    & ARG-Designer$^\dagger$ & 23.53 & 34.46 & 72.63 & 38.56 & 42.26 & 84.62 & 83.45 & 70.11$_{\pm0.8}$ \\
    \cmidrule(lr){3-10}
    & Ours & \textbf{25.86} & \textbf{35.89} & \textbf{74.63} & \textbf{42.11} & \textbf{44.62} & \textbf{87.96} & \textbf{96.67} & \textbf{76.42}$_{\pm1.2}$ \\
    \bottomrule
  \end{tabular}
  \caption{Full-set training performance with \textbf{HuatuoGPT-Vision (7B)}. The t-tests demonstrate the improvements are statistically significant with $p<0.05$ level. Best results are in bold and second-best results are underlined.}
  \label{tab:huatuo_fullset}
  \vspace{-1em}
\end{table*}

\begin{table*}[!t]
  \centering
  \footnotesize
  \renewcommand\cellalign{cc}
  \renewcommand{\arraystretch}{0.9}
  \setlength{\tabcolsep}{2.5pt}
  \begin{tabular}{ccc|ccccc|c|c|c}
    \toprule
    \multirow{3}{*}{\makecell{Training\\Paradigm}} 
    & \multirow{3}{*}{Method} 
    & \multirow{3}{*}{\makecell{Sample\\Scale}}
    & \multicolumn{5}{c|}{\raisebox{-0.2em}{MIMIC-CXR}} 
    & \multicolumn{1}{c|}{\raisebox{-0.2em}{IU-Xray}} 
    & \multicolumn{1}{c|}{\raisebox{-0.2em}{\makecell{Omni-\\MedVQA}}}  
    & \multirow{3}{*}{Average} \\
    & & & \multicolumn{5}{c|}{\raisebox{0.3em}{\rule{185pt}{0.4pt}}}
    & \multicolumn{1}{c|}{\raisebox{0.3em}{\rule{32pt}{0.4pt}}} 
    & \multicolumn{1}{c|}{\raisebox{0.3em}{\rule{38pt}{0.4pt}}} \\
    & & & BLEU & ROUGE-L & BERT-F1 & METEOR & Avg. & Acc & Acc \\
    \midrule
    \rowcolor{gray!20}
    \multicolumn{11}{c}{\textbf{Few-shot Training}} \\
    \midrule
    \multirow{6}{*}{Zero-shot}
    & \multirow{3}{*}{SP} & 10 & 11.92 & 24.88 & 64.15 & 21.26 & 30.55 & 75.73 & 70.55 & 58.94$_{\pm0.6}$ \\
    & & 20 & 11.03 & 24.77 & 64.53 & 21.25 & 30.39 & 74.50 & 70.48 & 58.46$_{\pm0.8}$ \\
    & & 40 & 11.90 & 24.76 & 64.13 & 21.22 & 30.50 & 74.65 & 70.57 & 58.57$_{\pm0.5}$ \\
    \cmidrule(lr){3-11}
    & \multirow{3}{*}{CoT} & 10 & 18.60 & 34.66 & 70.28 & 28.48 & 38.00 & 75.73 & 71.97 & 61.90$_{\pm1.1}$ \\
    & & 20 & 18.63 & 34.68 & 70.30 & 28.51 & 38.03 & 75.87 & 71.91 & 61.94$_{\pm0.9}$ \\
    & & 40 & 18.57 & 34.61 & 70.25 & 28.42 & 37.96 & 76.00 & 71.83 & 61.93$_{\pm0.7}$ \\
    \cmidrule(lr){1-11}
    \multirow{9}{*}[-4pt]{\makecell{Multimodal\\RAG}}
    & \multirow{3}{*}{MMed-RAG} & 10 & 20.43 & 30.21 & 71.87 & 30.29 & 38.20 & 83.46 & 85.83 & 69.16$_{\pm0.8}$ \\
    & & 20 & 20.76 & 30.09 & 71.13 & 30.36 & 38.08 & 84.06 & 85.10 & 69.08$_{\pm1.0}$ \\
    & & 40 & 20.69 & 30.84 & 71.91 & 30.56 & 38.50 & 82.20 & 84.52 & 68.41$_{\pm0.6}$ \\
    \cmidrule(lr){3-11}
    & \multirow{3}{*}{MIRA} & 10 & 7.65 & 16.65 & 63.82 & 27.98 & 29.03 & 73.70 & 85.35 & 62.69$_{\pm0.9}$ \\
    & & 20 & 7.02 & 15.90 & 63.39 & 27.16 & 28.37 & 73.80 & 87.92 & 63.36$_{\pm1.2}$ \\
    & & 40 & 7.20 & 16.15 & 63.52 & 27.19 & 28.52 & 73.98 & 89.37 & 63.96$_{\pm0.7}$ \\
    \cmidrule(lr){3-11}
    & \multirow{3}{*}{RULE} & 10 & 7.53 & 16.14 & 64.11 & 28.26 & 29.01 & 81.04 & 69.37 & 59.81$_{\pm0.8}$ \\
    & & 20 & 7.00 & 15.36 & 63.65 & 27.75 & 28.44 & 81.32 & 66.42 & 58.73$_{\pm1.1}$ \\
    & & 40 & 6.99 & 15.27 & 63.50 & 27.77 & 28.38 & 81.76 & 67.32 & 59.15$_{\pm0.5}$ \\
    \cmidrule(lr){1-11}
    \multirow{10}{*}{\makecell{Post-\\training}}
    & \multirow{3}{*}{LoRA} & 10 & 11.48 & 21.18 & 67.37 & 35.52 & 33.89 & 84.27 & 71.37 & 63.18$_{\pm0.6}$ \\
    & & 20 & 11.44 & 21.12 & 67.34 & 35.47 & 33.84 & 84.29 & 71.40 & 63.18$_{\pm0.7}$ \\
    & & 40 & 11.67 & 21.41 & 67.46 & 35.69 & 34.06 & 84.72 & 71.77 & 63.52$_{\pm1.0}$ \\
    \cmidrule(lr){3-11}
    & \multirow{3}{*}{QLoRA} & 10 & 11.64 & 21.45 & 67.33 & 35.15 & 33.89 & 84.20 & 71.23 & 63.11$_{\pm0.9}$ \\
    & & 20 & 11.67 & 21.49 & 67.34 & 35.25 & 33.94 & 84.37 & 71.45 & 63.25$_{\pm0.5}$ \\
    & & 40 & 11.87 & 21.73 & 67.45 & 35.38 & 34.11 & 84.96 & 71.38 & 63.48$_{\pm1.1}$ \\
    \cmidrule(lr){3-11}
    & \multirow{3}{*}{GRPO} & 10 & 6.23 & 13.93 & 63.18 & 28.80 & 28.04 & 79.44 & 75.27 & 60.92$_{\pm0.7}$ \\
    & & 20 & 6.12 & 13.76 & 63.10 & 28.71 & 27.92 & 81.47 & 75.50 & 61.63$_{\pm1.3}$ \\
    & & 40 & 6.15 & 13.80 & 63.15 & 28.76 & 27.96 & 81.96 & 82.96 & 64.29$_{\pm0.8}$ \\
    \cmidrule(lr){1-11}
    \multicolumn{2}{c}{} & \multicolumn{9}{c}{\textbf{Fixed-Template Evolving}} \\
    \cmidrule(lr){3-11}
    \multirow{24}{*}[-12pt]{Multi-agent}
    & \multirow{3}{*}{CARE} & 10 & 14.36 & 25.33 & 67.81 & 27.11 & 33.65 & 85.06 & 83.72 & 67.48$_{\pm0.8}$ \\
    & & 20 & 13.29 & 23.63 & 67.30 & 27.93 & 33.04 & 85.04 & 83.82 & 67.30$_{\pm1.0}$ \\
    & & 40 & 14.02 & 24.70 & 67.80 & 27.71 & 33.56 & 85.35 & 84.20 & 67.70$_{\pm0.6}$ \\
    \cmidrule(lr){3-11}
    & \multirow{3}{*}{AdaCoMed} & 10 & 20.05 & 30.44 & 71.42 & 39.76 & 40.42 & 81.00 & 83.12 & 68.18$_{\pm0.6}$ \\
    & & 20 & 20.01 & 30.40 & 71.37 & 39.71 & 40.37 & 79.95 & 82.93 & 67.75$_{\pm0.9}$ \\
    & & 40 & 20.03 & 30.42 & 71.40 & 39.70 & 40.39 & 81.96 & 82.83 & 68.39$_{\pm1.2}$ \\
    \cmidrule(lr){3-11}
    & \multirow{3}{*}{MedAgent-Pro} & 10 & 18.27 & 35.26 & 70.17 & 27.35 & 37.76 & 85.82 & 85.47 & 69.68$_{\pm1.1}$ \\
    & & 20 & 17.40 & 35.00 & 69.68 & 26.50 & 37.15 & 85.86 & 85.50 & 69.50$_{\pm0.7}$ \\
    & & 40 & 17.39 & 34.97 & 69.65 & 26.48 & 37.12 & 85.91 & 84.75 & 69.26$_{\pm0.9}$ \\
    \cmidrule(lr){3-11}
    & \multirow{3}{*}{MMedAgent-RL} & 10 & 15.51 & 25.75 & 69.56 & 38.86 & 37.42 & 85.56 & 83.61 & 68.86$_{\pm0.8}$ \\
    & & 20 & 15.99 & 26.38 & 69.96 & 39.24 & 37.89 & 85.59 & 83.44 & 68.97$_{\pm1.0}$ \\
    & & 40 & 16.11 & 26.51 & 70.04 & 39.29 & 37.99 & 85.75 & 83.69 & 69.14$_{\pm0.6}$ \\
    \cmidrule(lr){3-11}
    \multicolumn{2}{c}{} & \multicolumn{9}{c}{\textbf{Adaptive Evolving}} \\
    \cmidrule(lr){3-11}
    & \multirow{3}{*}{M$^3$Prune} & 10 & 23.41 & 34.57 & 72.74 & 38.78 & 42.38 & 85.95 & 81.06 & 69.80$_{\pm0.8}$ \\
    & & 20 & 23.47 & 34.31 & 72.99 & 39.09 & 42.46 & 83.78 & 80.85 & 69.03$_{\pm0.7}$ \\
    & & 40 & 23.46 & 34.31 & 72.98 & 38.23 & 42.24 & 86.34 & 81.24 & 69.94$_{\pm0.5}$ \\
    \cmidrule(lr){3-11}
    & \multirow{3}{*}{ARG-Designer$^\dagger$} & 10 & 23.28 & 34.42 & 72.61 & 38.62 & 42.18 & 85.45 & 80.87 & 69.50$_{\pm0.7}$ \\
    & & 20 & 23.33 & 34.16 & 72.85 & 38.92 & 42.24 & 85.43 & 80.73 & 69.47$_{\pm0.8}$ \\
    & & 40 & 23.32 & 34.17 & 72.84 & 38.05 & 42.04 & 85.86 & 81.02 & 69.64$_{\pm0.7}$ \\
    \cmidrule(lr){3-11}
    & \multirow{3}{*}{Ours} & 10 & \textbf{25.86} & \textbf{35.56} & \textbf{74.62} & \textbf{42.20} & \textbf{44.56} & \textbf{86.19} & \textbf{95.81} & \textbf{75.52}$_{\pm0.7}$ \\
    & & 20 & \textbf{25.55} & \textbf{35.43} & \textbf{74.52} & \textbf{42.37} & \textbf{44.47} & \textbf{87.07} & \textbf{95.87} & \textbf{75.80}$_{\pm1.0}$ \\
    & & 40 & \textbf{25.89} & \textbf{35.59} & \textbf{74.59} & \textbf{42.10} & \textbf{44.54} & \textbf{86.54} & \textbf{95.72} & \textbf{75.60}$_{\pm0.6}$ \\
    \bottomrule
  \end{tabular}
  \caption{Few-shot training performance with \textbf{HuatuoGPT-Vision (7B)}. The t-tests demonstrate the improvements are statistically significant with $p<0.05$ level. Best results are in bold and second-best results are underlined.}
  \label{tab:huatuo_fewshot}
\end{table*}

\begin{table*}[!t]
  \centering
  \footnotesize
  \renewcommand\cellalign{cc}
  \renewcommand{\arraystretch}{1}
  \setlength{\tabcolsep}{1pt}
  \begin{tabular}{ccc|ccccc|c|c|c}
    \toprule
    \multirow{3}{*}{\makecell{Training\\Paradigm}} 
    & \multirow{3}{*}{Method} 
    & \multirow{3}{*}{\makecell{Sample\\Size}}
    & \multicolumn{5}{c|}{\raisebox{-0.2em}{MIMIC-CXR}} 
    & \multicolumn{1}{c|}{\raisebox{-0.2em}{IU-Xray}} 
    & \multicolumn{1}{c|}{\raisebox{-0.2em}{\makecell{Omni-\\MedVQA}}}  
    & \multirow{3}{*}{Average} \\
    & & & \multicolumn{5}{c|}{\raisebox{0.3em}{\rule{185pt}{0.4pt}}}
    & \multicolumn{1}{c|}{\raisebox{0.3em}{\rule{32pt}{0.4pt}}} 
    & \multicolumn{1}{c|}{\raisebox{0.3em}{\rule{38pt}{0.4pt}}} \\
    & & & BLEU & ROUGE-L & BERT-F1 & METEOR & Avg. & Acc & Acc \\
    \midrule
    \rowcolor{gray!20} \multicolumn{11}{c}{\textbf{Full-set Training}} \\
    \midrule
    \multirow{2}{*}{Zero-shot}
    & SP & / & 18.70 & 30.38 & 69.46 & 28.52 & 36.77 & 88.54 & 87.08 & 70.80$_{\pm0.7}$ \\
    & CoT & / & 19.14 & 30.62 & 70.76 & 29.43 & 37.49 & 87.38 & 89.86 & 71.58$_{\pm1.0}$ \\
    \cmidrule(lr){1-11}
    \multirow{1}{*}{\makecell{Multimodal RAG}}
    & MMed-RAG & / & 21.18 & 32.35 & 71.04 & 31.22 & 38.95 & 90.10 & 93.74 & 74.26$_{\pm0.8}$ \\
    \cmidrule(lr){1-11}
    \multicolumn{2}{c}{} & \multicolumn{9}{c}{\textbf{Fixed-Template Evolving}} \\
    \cmidrule(lr){3-11}
    \multirow{4}{*}{Multi-agent}
    & MedAgent-Pro & / & \underline{22.04} & \underline{34.92} & 72.21 & 32.36 & 40.38 & \underline{91.65} & 94.13 & \underline{75.39}$_{\pm0.9}$ \\
    \cmidrule(lr){3-11}
    \multicolumn{2}{c}{} & \multicolumn{9}{c}{\textbf{Adaptive Evolving}} \\
    \cmidrule(lr){3-11}
    & M$^3$Prune & / & 21.92 & 34.49 & \underline{72.85} & \underline{34.84} & \underline{41.03} & 90.87 & \underline{94.14} & 75.35$_{\pm1.3}$ \\
    & ARG-Designer$^\dagger$ & / & 21.78 & 34.34 & 72.71 & 34.68 & 40.86 & 90.64 & 93.89 & 75.13$_{\pm0.9}$ \\
    \cmidrule(lr){3-11}
    & Ours & / & \textbf{24.44} & \textbf{35.89} & \textbf{74.84} & \textbf{38.02} & \textbf{43.30} & \textbf{93.20} & \textbf{97.07} & \textbf{77.86}$_{\pm1.2}$ \\
    \midrule
    \rowcolor{gray!20} \multicolumn{11}{c}{\textbf{Few-shot Training}} \\
    \midrule
    \multirow{6}{*}{Zero-shot}
    & \multirow{3}{*}{SP} & 10 & 18.95 & 30.45 & 69.20 & 28.13 & 36.68 & 88.45 & 84.72 & 69.95$_{\pm0.6}$ \\
    & & 20 & 18.36 & 30.76 & 69.72 & 28.54 & 36.85 & 89.25 & 84.42 & 70.17$_{\pm1.0}$ \\
    & & 40 & 18.97 & 30.36 & 69.21 & 28.13 & 36.67 & 89.13 & 84.84 & 70.21$_{\pm0.4}$ \\
    \cmidrule(lr){3-11}
    & \multirow{3}{*}{CoT} & 10 & 15.81 & 30.88 & 69.84 & 26.93 & 35.87 & 87.97 & 89.63 & 71.16$_{\pm1.4}$ \\
    & & 20 & 16.68 & 30.63 & 69.81 & 27.42 & 36.14 & 86.83 & 89.49 & 70.82$_{\pm0.8}$ \\
    & & 40 & 16.95 & 31.02 & 70.29 & 27.92 & 36.55 & 87.88 & 89.17 & 71.20$_{\pm0.3}$ \\
    \cmidrule(lr){1-11}
    \multirow{3}{*}{Multimodal RAG}
    & \multirow{3}{*}{MMed-RAG} & 10 & 20.45 & 32.49 & 70.78 & 30.67 & 38.60 & 89.32 & 90.46 & 72.79$_{\pm1.1}$ \\
    & & 20 & 21.46 & 33.09 & 71.31 & 30.98 & 39.21 & 89.22 & 88.57 & 72.33$_{\pm0.7}$ \\
    & & 40 & 22.10 & 32.89 & 71.41 & 30.28 & 39.17 & 88.15 & 90.90 & 72.74$_{\pm0.9}$ \\
    \cmidrule(lr){1-11}
    \multicolumn{2}{c}{} & \multicolumn{9}{c}{\textbf{Fixed-Template Evolving}} \\
    \cmidrule(lr){3-11}
    \multirow{15}{*}{Multi-agent}
    & \multirow{3}{*}{MedAgent-Pro} & 10 & 21.38 & 34.16 & 71.74 & 32.08 & 39.84 & 89.30 & 92.72 & 73.95$_{\pm0.8}$ \\
    & & 20 & 21.36 & 34.04 & 71.17 & 31.22 & 39.45 & 89.35 & 91.76 & 73.52$_{\pm1.4}$ \\
    & & 40 & 21.47 & 34.30 & 71.71 & 32.30 & 39.95 & 90.45 & 91.53 & 73.98$_{\pm0.6}$ \\
    \cmidrule(lr){3-11}
    \multicolumn{2}{c}{} & \multicolumn{9}{c}{\textbf{Adaptive Evolving}} \\
    \cmidrule(lr){3-11}
    & \multirow{3}{*}{M$^3$Prune} & 10 & 22.67 & 33.88 & 72.96 & 33.80 & 40.83 & 88.47 & 92.87 & 74.06$_{\pm1.2}$ \\
    & & 20 & 22.11 & 34.42 & 73.01 & 33.31 & 40.71 & 88.53 & 93.09 & 74.11$_{\pm0.5}$ \\
    & & 40 & 21.62 & 33.90 & 72.48 & 33.34 & 40.34 & 88.13 & 93.70 & 74.06$_{\pm1.3}$ \\
    \cmidrule(lr){3-11}
    & \multirow{3}{*}{ARG-Designer$^\dagger$} & 10 & 22.40 & 33.60 & 72.67 & 33.50 & 40.46 & 88.08 & 92.50 & 73.68$_{\pm1.0}$ \\
    & & 20 & 21.95 & 34.24 & 72.83 & 33.12 & 40.47 & 88.28 & 92.77 & 73.84$_{\pm0.5}$ \\
    & & 40 & 21.45 & 33.72 & 72.29 & 33.15 & 40.09 & 87.77 & 93.27 & 73.71$_{\pm0.6}$ \\
    \cmidrule(lr){3-11}
    & \multirow{3}{*}{Ours} & 10 & \textbf{23.05} & \textbf{35.18} & \textbf{73.96} & \textbf{36.83} & \textbf{42.26} & \textbf{93.00} & \textbf{96.69} & \textbf{77.32}$_{\pm1.0}$ \\
    & & 20 & \textbf{23.59} & \textbf{34.53} & \textbf{74.28} & \textbf{37.03} & \textbf{42.36} & \textbf{92.66} & \textbf{96.28} & \textbf{77.10}$_{\pm0.9}$ \\
    & & 40 & \textbf{23.07} & \textbf{35.07} & \textbf{73.71} & \textbf{37.55} & \textbf{42.35} & \textbf{93.21} & \textbf{96.23} & \textbf{77.26}$_{\pm0.8}$ \\
    \bottomrule
  \end{tabular}
  \caption{The performance of full-set and few-shot training is reported using the closed‑source \textbf{Qwen3.6-Plus (397B)} accessed via its API. Since the API does not support gradient backpropagation or parameter updates, model fine-tuning is infeasible. We therefore restrict our comparison to training-free baselines. The t-tests demonstrate the improvements are statistically significant with $p<0.05$ level. Bold black indicates the best results, while underlined is the second ranked results.}
  \label{tab:qwen3.6plus_performance}
  \vspace{-1em}
\end{table*}

\begin{table*}[!t]
  \centering
  \setlength{\tabcolsep}{14pt}
  \begin{tabular}{c|c|c|c|c|c|c}
    \toprule
    Dataset $\rightarrow$& \multicolumn{2}{c|}{MIMIC-CXR} & \multicolumn{2}{c|}{IU-Xray} & \multicolumn{2}{c}{OmniMedVQA} \\
    \cmidrule(lr){2-3} \cmidrule(lr){4-5} \cmidrule(lr){6-7}
    Models $\downarrow$ & \#Tokens & $\mathcal{F}$ & \#Tokens & $\mathcal{F}$ & \#Tokens & $\mathcal{F}$ \\
    \midrule
    CARE         & 11.11k & 40.08 & 6.46k & 89.90 & 12.79k & 92.80 \\
    AdaCoMed     & 10.24k & 41.46 & 8.04k & 90.10 & 13.19k & 92.45 \\
    MedAgent-Pro & 8.59k  & 41.42 & 6.31k & 88.35 & 13.03k & 91.01 \\
    MMedAgent-RL & 11.24k & 41.49 & 6.09k & 88.93 & 14.12k & 91.80 \\
    M$^3$Prune   & 8.23k  & 40.09 & 6.02k & 89.13 & 12.92k & 90.51 \\
    ARG-Designer$^\dagger$ & 8.71k & 40.38 & 6.44k & 88.54 & 13.57k & 90.11 \\
    \cmidrule(lr){1-7}
    \texttt{MedPrune}     & 6.54k  & 42.26 & 4.95k & 91.07 & 10.88k & 96.36 \\
    \bottomrule
  \end{tabular}
  \caption{Token consumption of per multi-modal medical question and dataset metrics ($\mathcal{F}$) on three datasets.}
  \label{tab:tokens}
\end{table*}

\subsection{Hyperparameters Analysis}
\label{hyper_para}
\begin{itemize}
    \item \textbf{Communication Rounds and Noise Levels:} Table ~\ref{tab_hyperparameter} reports performance under varying numbers of communication rounds ($T$) and noise levels ($\delta$).
    Experiments on Qwen3-VL (8B) and HuatuoGPT-Vision (7B) show that performance remains stable across these settings.
    Table \ref{training_settings} presents experiments on training samples to demonstrate stability.
    \item \textbf{Varying Number of Agents:} We adjust the agent count per dataset according to question complexity and task requirements. To explore how this configuration affects performance, we experiment on MIMIC-CXR with the Qwen3-VL (8B) and HuatuoGPT-Vision (7B) models, steadily increasing the number of departments and agents per department. As illustrated in Fig.~\ref{heatmap_agent_nums_two_models}, \texttt{MedPrune} remains robust under moderate variations in agent numbers, without noticeable performance loss. Thus, the model does not rely critically on an exact agent configuration.
    \item \textbf{Pruning Rates Analysis:} To study how pruning strength affects \texttt{MedPrune}, we evaluate node and edge pruning rates under varying sparsity configurations on MIMIC-CXR, using Qwen3-VL (8B) and HuatuoGPT-Vision (7B) as backbone models.
    We vary the pruning rates across a wide spectrum, ranging from extremely low to extremely high, to examine model behavior under both overly dense and overly sparse topological configurations.
    As illustrated in Fig.~\ref{heatmap_pruning_rate_two_models}, both overly conservative and overly aggressive pruning lead to suboptimal performance. Excessively low pruning rates retain redundant edges, introducing noise that degrades decision quality, while overly high pruning rates remove critical communication pathways. The optimal performance is achieved at moderate pruning levels, indicating that a balanced sparsity preserves informative interactions while effectively filtering noise.
\end{itemize}

\subsection{Detailed Reasoning Cost Results}
\label{detailed_cost}
In Table ~\ref{tab:tokens}, we provide the comprehensive token consumption per instance.
We assess the computational cost by measuring the average token usage per question.
For precise computational cost accounting, we explicitly clarify that the counted input tokens cover agent role prompts, multi-round dialogue historical messages, as well as repeated image and question text embeddings passed to each specialist agent. We further specify that output tokens are calculated as the total sum of all generation tokens produced by every agent throughout all collaborative communication rounds.

Table \ref{tab:training_cost} reports comprehensive training-side metrics, including GPU hours, peak VRAM, training epochs and training token consumption.
Most competitive multi-agent baselines require updating full graph topology, leading to substantial training costs.
In contrast, \texttt{MedPrune} freezes the entire vision-language backbone and only optimizes lightweight soft adjacency matrices via policy gradient, significantly reducing training resource consumption.
For fixed-template models that do not require training and only require calling APIs for simple multi-agent collaboration, our training cost is acceptable in real-world scenarios, while our inference cost yields far greater long-term benefits than these fixed-template models.

\begin{table*}[t]
\centering
\footnotesize
\begin{tabular}{l c c c c c}
\toprule
Method & \makecell[c]{Tune \\ Backbone} & \makecell[c]{Total GPU Hours \\ (H)} & \makecell[c]{Peak VRAM \\ (GB)} & \makecell[c]{Train Convergence \\ Epochs} & \makecell[c]{Train Tokens}\\
\midrule
\makecell[l]{CARE / AdaCoMed \\ MedAgent-Pro} & \xmark & 0 & $<$10 & 0 & 0 \\ \midrule
\makecell[l]{MMedAgent-RL / M$^3$Prune \\ ARG-Designer} & \makecell[c]{Topology \\ Module} & 17 / 5 / 7 & 49 / 32 / 29 & 8 / 2 / 5 & 18.6k / 14.3k / 16.5k \\ \midrule
\textbf{MedPrune (Ours)} & \makecell[c]{Topology \\ Module} & 1.5 & 18 & 2 & 10.2k \\
\bottomrule
\end{tabular}
\caption{Training average overhead of all multi-agent baselines on single GPU on three datasets.
``Tune Backbone'': \xmark: Core vision-language transformer backbone fully frozen with no trainable extra modules;
Topology Module: Only multi-agent communication graph parameters optimized, vision-language backbone frozen. The training process of multi-agent baselines is \textbf{one-time}, but the model used in the inference process is reused \textbf{multiple times}.}
\label{tab:training_cost}
\end{table*}

\subsection{Prompt Sensitivity}
\label{prompt_sen}
We evaluate the learned typology's sensitivity to prompt variations by freezing \texttt{MedPrune’s} pruning weights and applying four different prompt templates at inference stage:
\begin{itemize}
    \item Prompt‑Ours: the default clinical prompts from the main paper (baseline).
    \item Prompt‑Detailed: extended version augmented with comprehensive clinical diagnostic workflows, differential diagnosis criteria, and standardized imaging interpretation checklists beyond the original baseline prompts.
    \item Prompt‑Simplified: containing only core instructions.
    \item Prompt‑Reversed: extreme robustness test variant that fully swaps the role prompts of chief physicians and regular doctors.
\end{itemize}

Table ~\ref{tab:Prompt Robustness} shows only minor performance differences across all prompt variants, with the largest deviation below 1.1\%. This confirms that \texttt{MedPrune’s} learned topology is robust to prompt changes, validating both our default prompt design and the stability of the method without retraining.

\begin{table*}[!htbp]
\centering
\begin{tabular}{lcccc}
\toprule
\textbf{Prompt Variant} & \textbf{MIMIC-CXR} & \textbf{IU-Xray} & \textbf{OmniMedVQA} & \textbf{Average} \\
\midrule
Prompt-Ours (Main)    & \textbf{42.26} & \textbf{91.07} & \textbf{96.36} & \textbf{76.56} \\
Prompt-Detailed       & 42.01 & 90.87 & 95.91 & 76.26 \\
Prompt-Simplified     & 41.75 & 90.68 & 95.46 & 75.96 \\
Prompt-Reversed       & 41.32 & 90.49 & 94.81 & 75.54 \\
\bottomrule
\end{tabular}
\caption{Prompt sensitivity of \texttt{MedPrune}. The well-trained topological pruning weights are fixed, only the agent prompt templates are replaced during inference.}
\label{tab:Prompt Robustness}
\end{table*}

\begin{table*}[!htbp]
\centering
\begin{tabular}{cccc}
\toprule
\textbf{ Dataset $\rightarrow$} & \multirow{2}{*}{\textbf{{\makecell{MIMIC-CXR $\rightarrow$IU-Xray}}}} & \multirow{2}{*}{\textbf{{\makecell{PMC‑OA $\rightarrow$MIMIC-CXR}}}} & \multirow{2}{*}{\textbf{Average}} \\
\textbf{Method $\downarrow$} & & & \\
\midrule
AdaCoMed & 85.05 & 36.95 & 61.00 \\
MMedAgent-RL & 82.91 & 35.98 & 59.45 \\
M$^3$Prune & 85.44 & 36.99 & 61.22 \\
ARG-Designer$^\dagger$ & 84.46 & 37.02 & 60.74 \\
\textbf{Ours (MedPrune)} & \textbf{87.96} & \textbf{38.96} & \textbf{63.46} \\
\bottomrule
\end{tabular}
\caption{Cross-Dataset Transfer results under full-set training based on Qwen3‑VL (8B) backbone. MIMIC-CXR→IU-Xray reports accuracy (\%); PMC-OA→MIMIC-CXR reports the average of BLEU, ROUGE-L, BERT-F1 and METEOR (Avg.).}
\label{tab:cross_dataset_transfer}
\end{table*}

\subsection{Cross‑Dataset Transfer}
\label{cross_data}
We further evaluate \texttt{MedPrune’s} transferability across datasets under full‑data training. All methods share the Qwen3‑VL (8B) backbone. We compare MedPrune with AdaCoMed, MMedAgent‑RL, M$^3$Prune and ARG-Designer$^\dagger$ in two scenarios: (1) training on MIMIC‑CXR and testing on unseen IU‑Xray; (2) training on PMC‑OA \cite{DBLP:conf/miccai/LinZZWZWX23} and testing on unseen MIMIC‑CXR.

Table ~\ref{tab:cross_dataset_transfer} shows that all methods experience moderate performance drops due to dataset shift, yet MedPrune consistently outperforms the baselines. This confirms that our heterogeneous node‑edge pruning strategy generalizes well to unseen datasets.

\subsection{Case Study}
\begin{figure*}[!t]
\centering
\includegraphics[width=15cm,height=8cm]{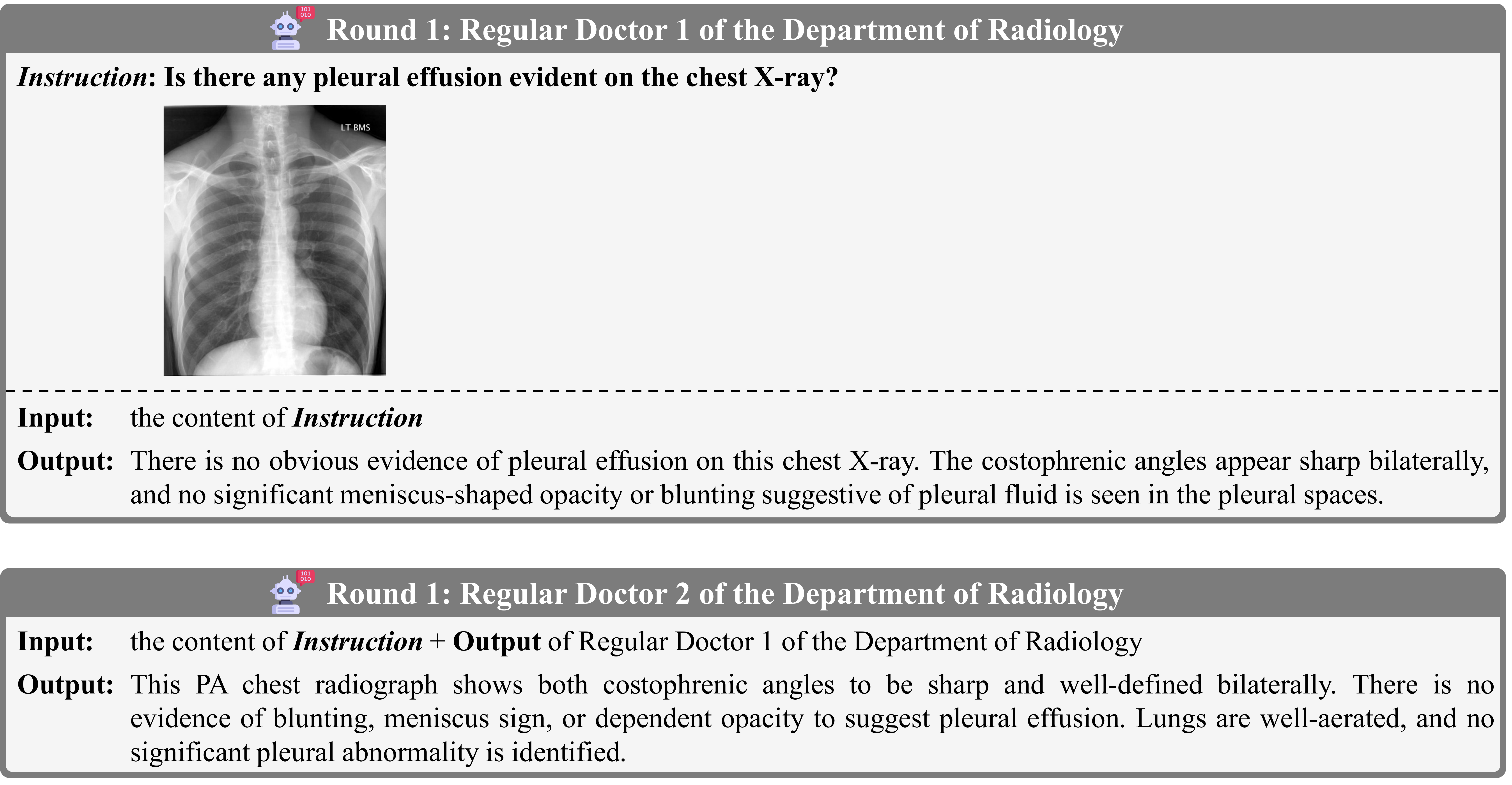}
\vspace{3mm}
\includegraphics[width=15cm,height=8cm]{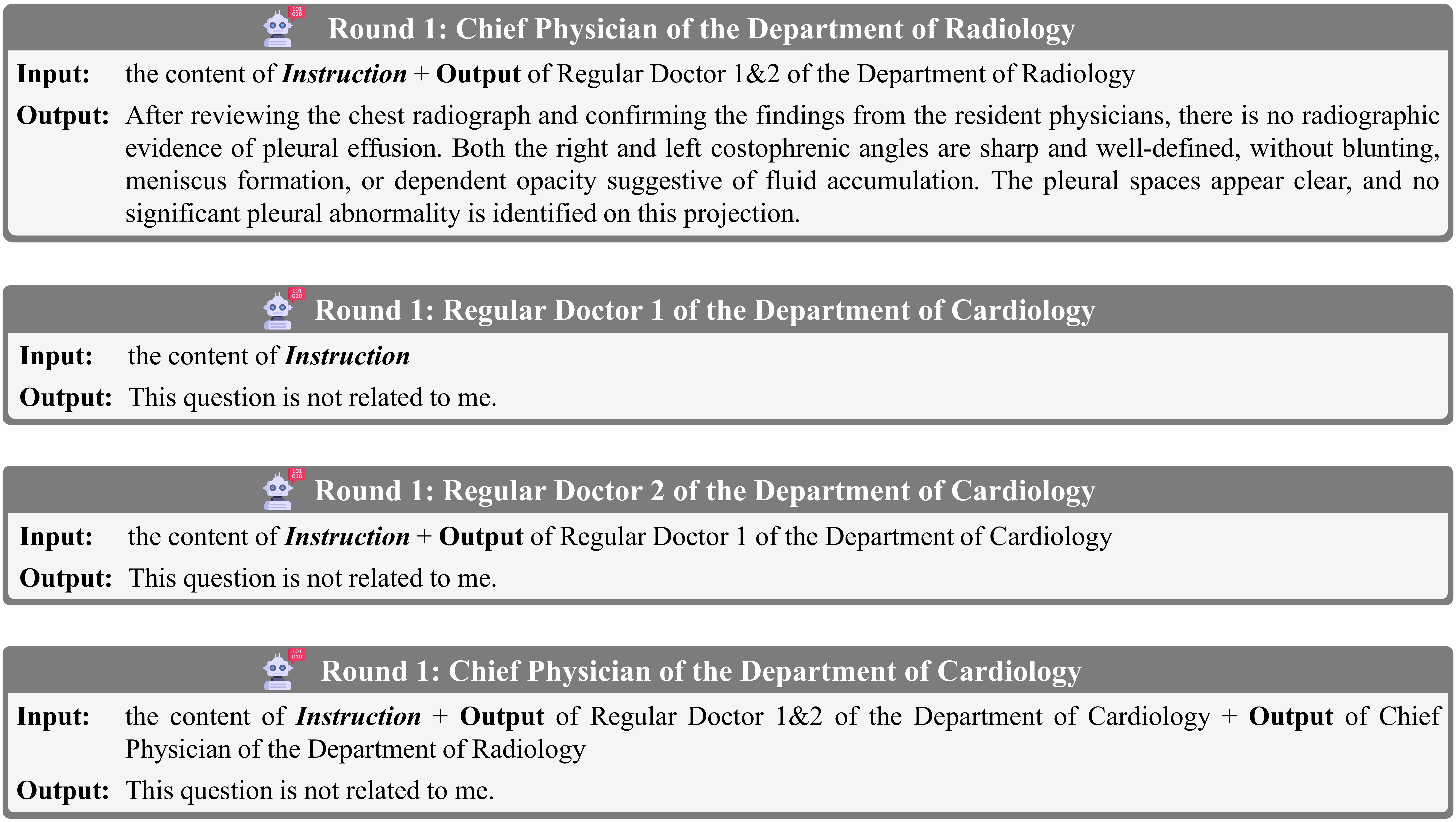}
\vspace{3mm}
\includegraphics[width=15cm,height=8cm]{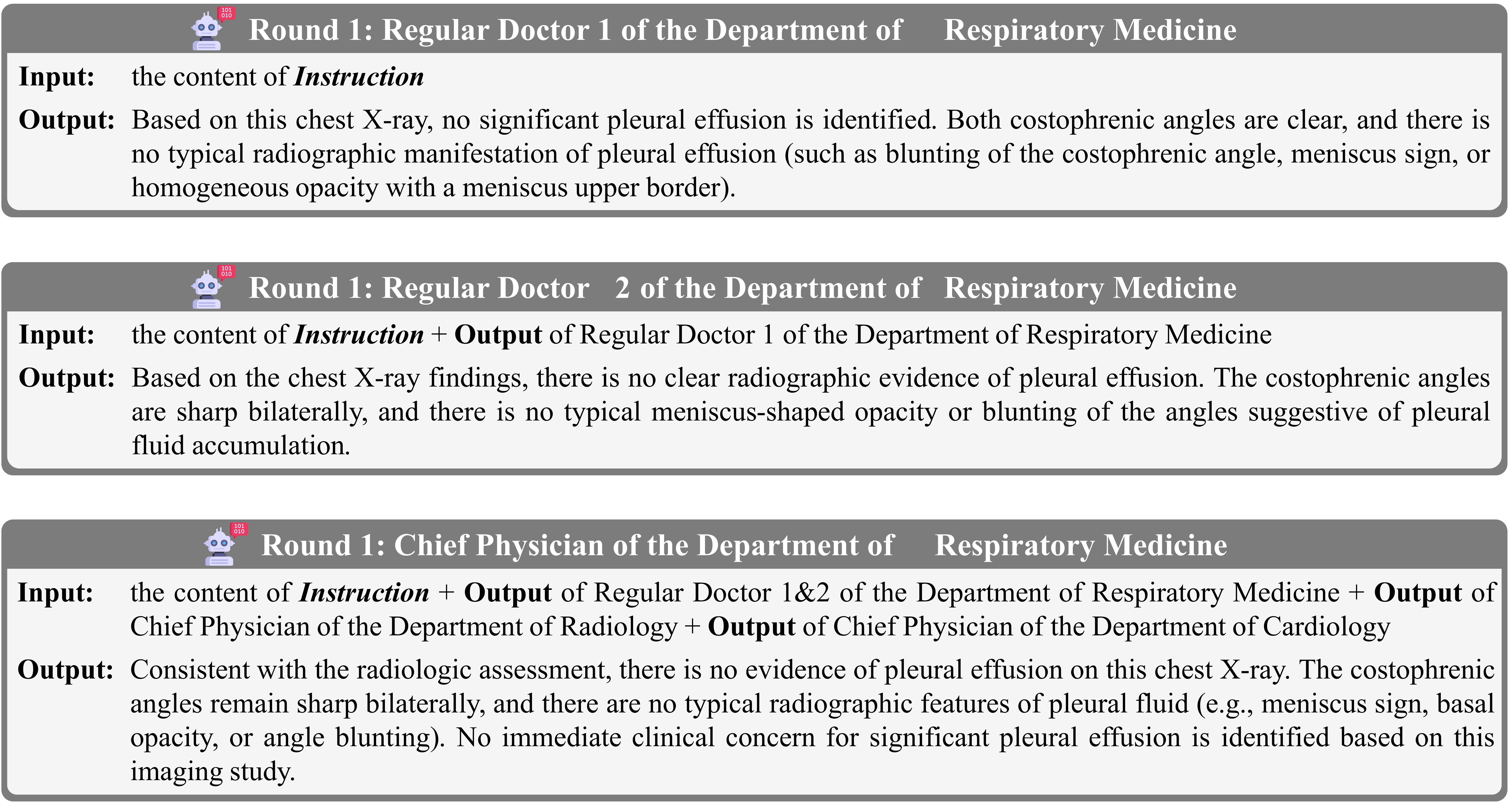}
\vspace{-1em}
\caption{Case study of Round 1 Discussion.}
\label{case_study1}
\end{figure*}

\begin{figure*}[!t]
\centering
\includegraphics[width=15cm,height=7cm]{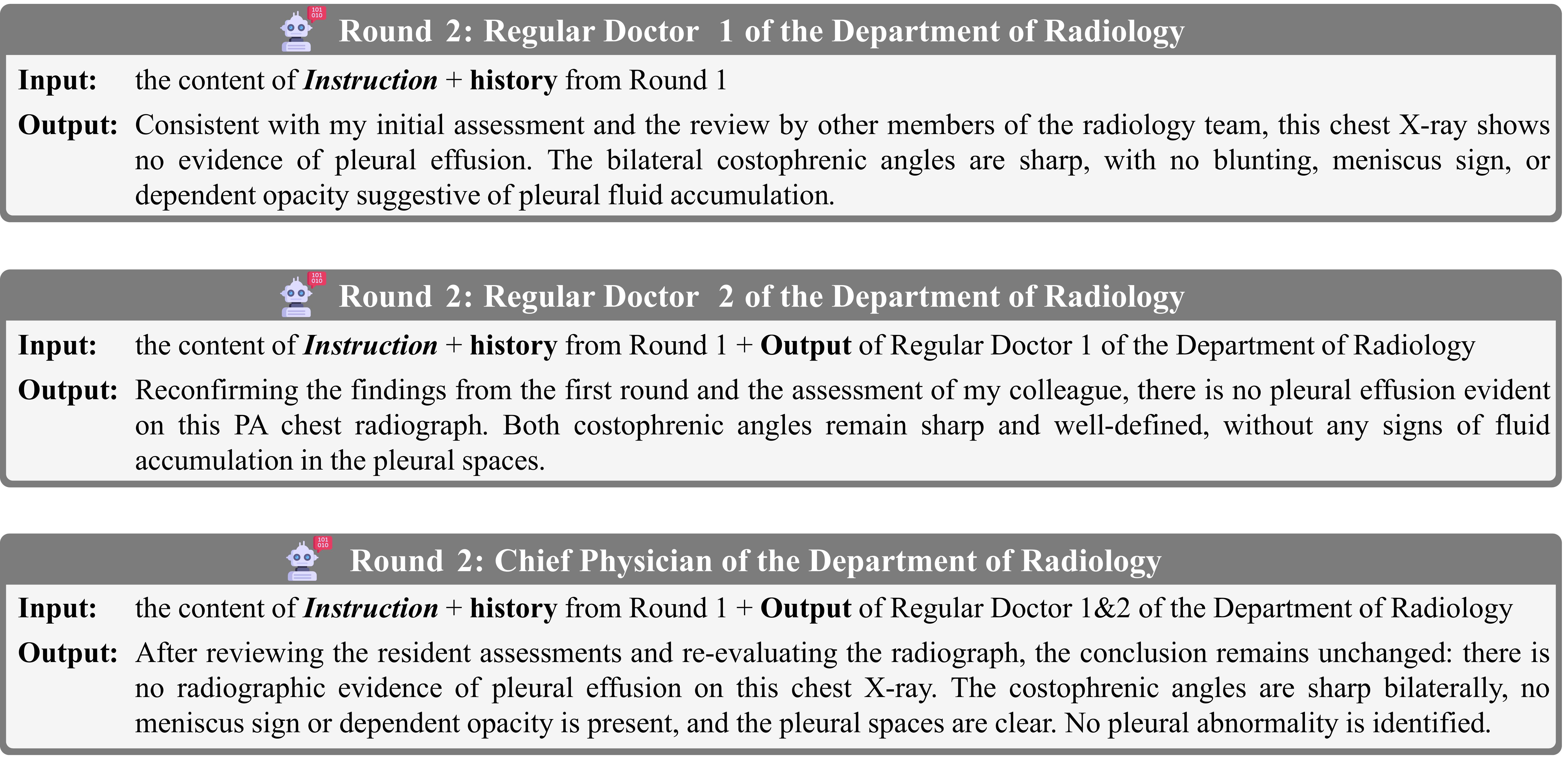}
\vspace{3mm}
\includegraphics[width=15cm,height=6cm]{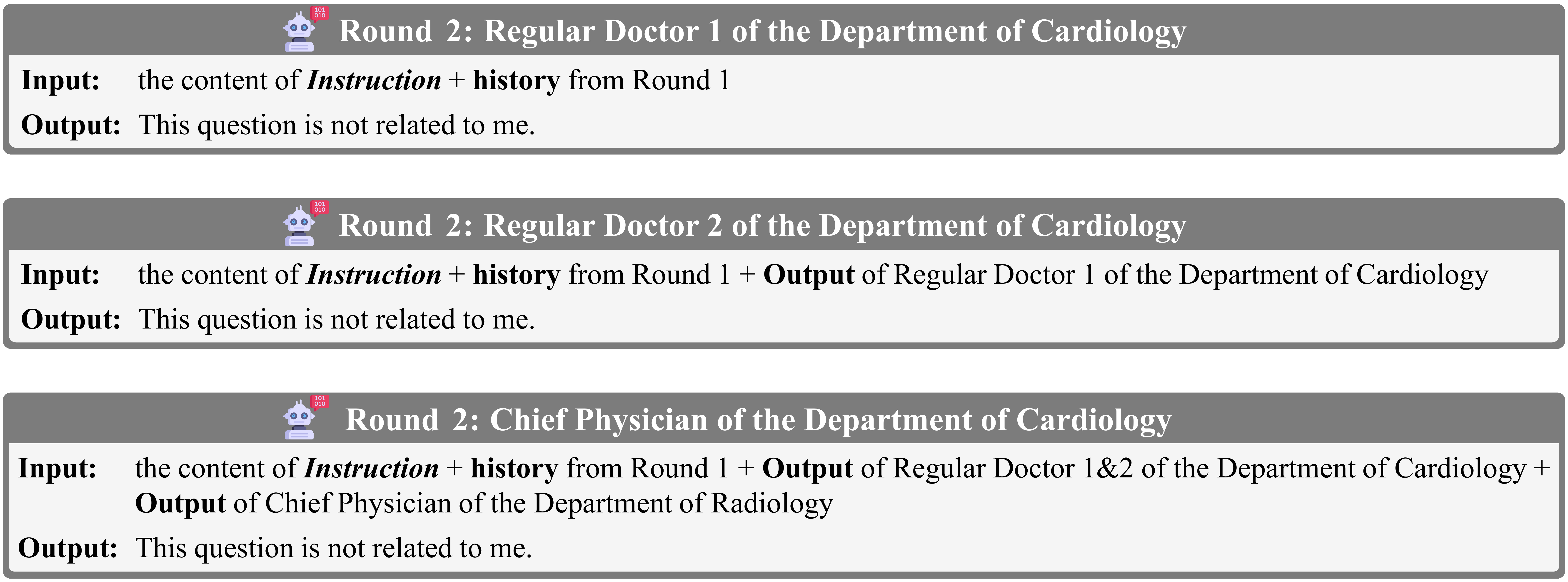}
\vspace{3mm}
\includegraphics[width=15cm,height=8cm]{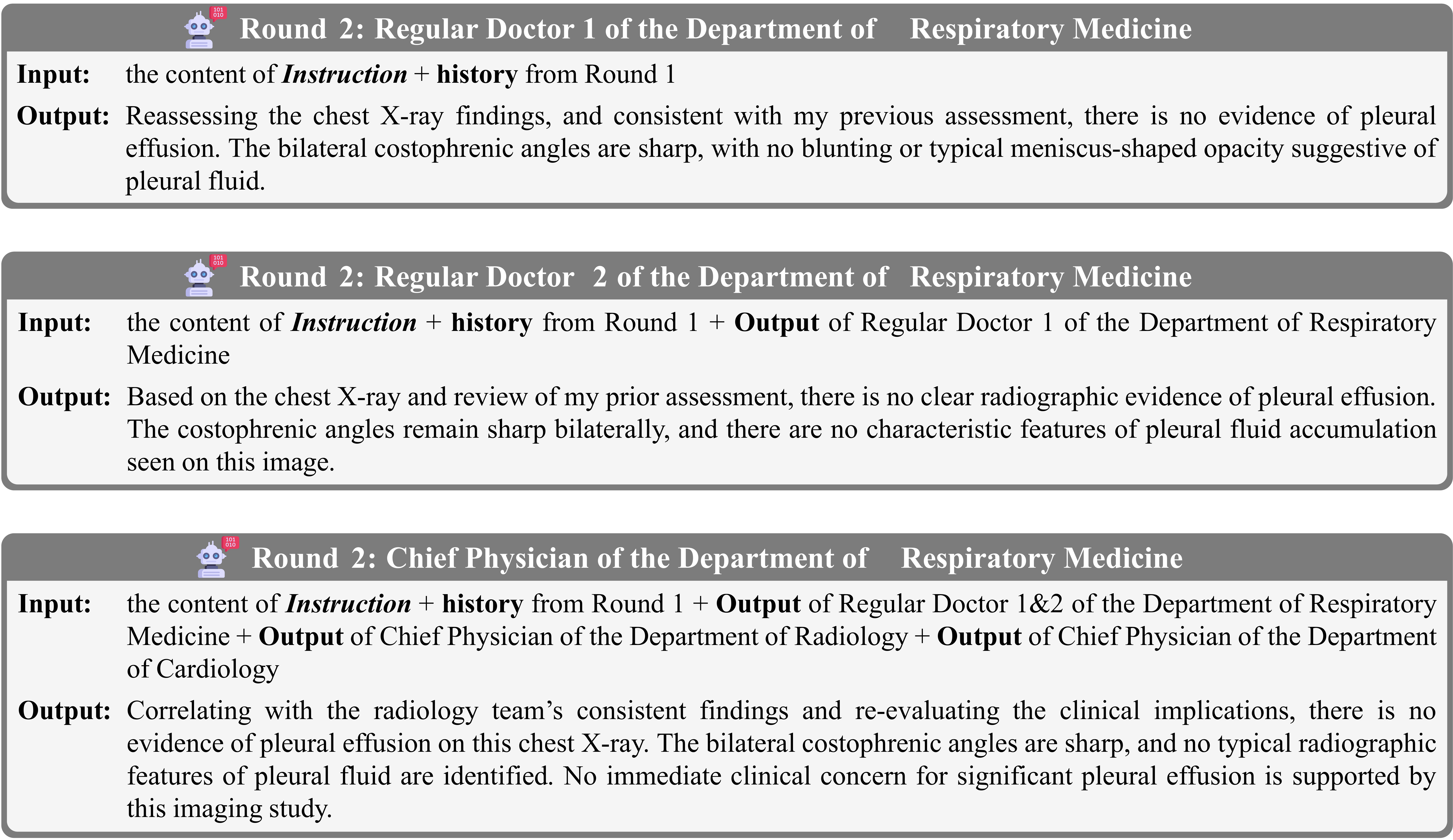}
\vspace{3mm}
\includegraphics[width=15cm,height=2cm]{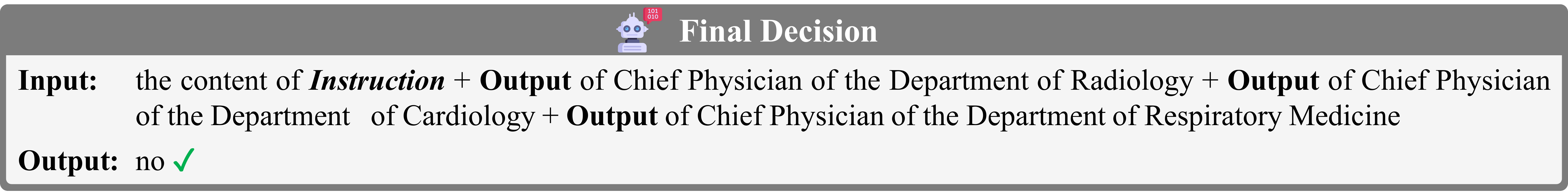}
\caption{Case study of Round 2 Discussion and Final Decision.}
\label{case_study2}
\end{figure*}

Fig.~\ref{case_study1} and ~\ref{case_study2} illustrate a case from the IU-Xray dataset, demonstrating the intra- and inter-departmental collaboration mechanism of \texttt{MedPrune}.

Specifically, in the first discussion round, agents from Radiology and Respiratory Medicine jointly examine the chest X‑ray. They conclude that there is no evidence of pleural effusion supported by several observable signs: (1) the bilateral costophrenic angles remain sharp and clear, (2) there is no blunting, no crescent-shaped shadows, and no lower-lung opacification.
Meanwhile, agents from Cardiology deem the question outside their specialty. They therefore abstain from making any judgment.
In the second round, agents from Radiology and Respiratory Medicine re‑examine the image. They integrate prior intra‑departmental opinions, the conclusions from the previous round, and inputs from other departments. They consistently reaffirm the absence of pleural effusion, again citing the clear costophrenic angles and the lack of characteristic signs. Their conclusion remains unchanged from the first round. Agents in the Cardiology department continue to abstain from participation.
After multiple rounds of discussion, the Radiology and Respiratory Medicine agents reach full consensus. They determine that the chest X‑ray shows no pleural effusion. The final answer is \textbf{no}. The Cardiology department does not participate in this question.

\end{document}